\documentclass{article} 
\usepackage{Template-2026/fancyhdr}

\usepackage[preprint]{Template-2026/colm2026_conference}

\usepackage{amsmath,amsfonts,bm}

\def\eqref#1{equation~\ref{#1}}

\def\1{\bm{1}}

\DeclareMathAlphabet{\mathsfit}{\encodingdefault}{\sfdefault}{m}{sl}
\SetMathAlphabet{\mathsfit}{bold}{\encodingdefault}{\sfdefault}{bx}{n}

\usepackage{microtype}
\usepackage{hyperref}
\usepackage{url}
\usepackage{booktabs}
\usepackage{graphicx}
\usepackage{wrapfig}
\usepackage{fvextra}
\usepackage[most]{tcolorbox}
\usepackage{placeins}
\usepackage{lineno}
\newtcblisting{PromptBlock}[1][]{
  enhanced,
  breakable,
  colback=blue!3,
  colframe=blue!45!black,
  colbacktitle=blue!55!black,
  coltitle=white,
  fonttitle=\bfseries,
  boxrule=0.6pt,
  arc=2pt,
  left=6pt,
  right=6pt,
  top=6pt,
  bottom=6pt,
  listing only,
  title={#1},
  listing options={
    basicstyle=\small\ttfamily,
    breaklines=true,
    breakatwhitespace=false,
    columns=fullflexible
  }
}

\usepackage{xcolor}
\definecolor{GapHighRed}{HTML}{B35757}   
\definecolor{GapLowBlue}{HTML}{2E5FA7}    
\definecolor{FigureLavender}{RGB}{238,232,247}

\definecolor{darkblue}{rgb}{0, 0, 0.5}
\definecolor{BestBlue}{rgb}{0.12, 0.47, 0.71}
\definecolor{WorstRed}{rgb}{0.84, 0.15, 0.16}
\hypersetup{colorlinks=true, citecolor=darkblue, linkcolor=darkblue, urlcolor=darkblue}

\title{Beyond Perception Errors: Semantic Fixation in Large Vision-Language Models}

\author{Md Tanvirul Alam \\
Rochester Institute of Technology\\
Rochester, NY, USA\\
\texttt{ma8235@rit.edu}
}

\begin{document}

\ifcolmsubmission
\linenumbers
\fi

\maketitle

\begin{abstract}
Large vision-language models (VLMs) often rely on familiar semantic priors, but existing evaluations do not cleanly separate perception failures from rule-mapping failures. We study this behavior as \emph{semantic fixation}: preserving a default interpretation even when the prompt specifies an alternative, equally valid mapping. To isolate this effect, we introduce \textit{VLM-Fix}, a controlled benchmark over four abstract strategy games that evaluates identical terminal board states under paired standard and inverse rule formulations. Across 14 open and closed VLMs, accuracy consistently favors standard rules, revealing a robust semantic-fixation gap. Prompt interventions support this mechanism: neutral alias prompts substantially narrow the inverse-rule gap, while semantically loaded aliases reopen it. Post-training is strongly rule-aligned: training on one rule improves same-rule transfer but hurts opposite-rule transfer, while joint-rule training improves broader transfer. To test external validity beyond synthetic games, we evaluate analogous defamiliarization interventions on \textit{VLMBias} and observe the same qualitative pattern. Finally, late-layer activation steering partially recovers degraded performance, indicating that semantic-fixation errors are at least partly editable in late representations. Project page, code, and dataset available at \url{https://maveryn.github.io/vlm-fix/}.
\end{abstract}

\section{Introduction}
Large vision-language models (VLMs) have recently achieved strong performance across instruction following, visual reasoning, and open-ended image understanding, and benchmark evaluations similarly report high aggregate results on broad multimodal tasks~\citep{hurst2024gpt,yang2023dawn,liu2023visual,deitke2025molmo,wang2025internvl3,bai2025qwen3}. However, these gains do not necessarily translate to robust reasoning in controlled settings: a growing body of work shows that VLMs often rely heavily on learned language priors, which can drive systematic bias and hallucination \citep{agrawal2018don,hall2023visogender,lee2025vlind,zhou2023rome,sharma2024vision,fu2025hidden,vo2025vision}. Across such evaluations, model predictions often track familiar semantic patterns rather than task-specified evidence \citep{ruggeri2023multidimvlmbias,huang2025visbias,vo2025vision,zhou2023rome}.

These findings motivate a complementary way to study prior dependence. Prior work often probes this behavior through biased prompting or counterfactual visual manipulations, but those settings can still leave ambiguity about whether failures stem from perceptual changes or from rigid semantic expectations. Here we target the latter directly, asking whether VLMs can revise decisions when task goals redefine the same perceptual state. We call this failure mode \textit{semantic fixation}, adapting the cognitive psychology notion of fixation (the Einstellung effect), where people persist with familiar solution patterns despite equally valid alternatives \citep{luchins1942mechanization,bilalic2008einstellung}. Related prior-driven behavior has also been observed in modern language models, including anchoring effects \citep{alavi2023large,huang2025anchoring,kim2025limitations}. In this work, we examine semantic fixation in VLMs: models may default to familiar meanings even when prompts specify alternative mappings. Because the same perceptual state can require different decisions under different task goals, robust reasoning depends on semantic remapping, not only visual recognition~\citep{jang2022bc,shi2025hi}.

\begin{figure*}[t]
\centering
\resizebox{0.8\textwidth}{!}{%
\setlength{\tabcolsep}{0pt}
\small
\begin{tabular}{@{}p{0.55\linewidth}p{0.45\linewidth}@{}}
\begin{minipage}[t]{\linewidth}
\centering
\textbf{Prompt Variants}\\[0.8mm]
\fcolorbox{black!25}{FigureLavender}{%
\begin{minipage}[t]{0.96\linewidth}
\footnotesize
\textbf{Base.} You are given a $3\times3$ grid for a two-player game (X and O). If a player has 3 in a row (horizontal, vertical, or diagonal), that player \textcolor{blue}{wins}/\textcolor{red}{loses} , and the other player \textcolor{blue}{loses} / \textcolor{red}{wins}. The game has ended. Who is the winner?
\end{minipage}}\\[0.45mm]
\fcolorbox{black!25}{FigureLavender}{%
\begin{minipage}[t]{0.96\linewidth}
\footnotesize
\textbf{Alias.} You are given a $3\times3$ grid for a two-player game (X and O). Outcome tags are POM and TOV. If a player has 3 in a row, that player is \textcolor{blue}{POM} / \textcolor{red}{TOV}, and the other player is \textcolor{blue}{TOV} / \textcolor{red}{POM}. The game has ended. Who is the TOV?
\end{minipage}}\\[0.45mm]
\fcolorbox{black!25}{FigureLavender}{%
\begin{minipage}[t]{0.96\linewidth}
\footnotesize
\textbf{SemAlias.} You are given a $3\times3$ grid for a two-player game (X and O). Outcome tags are POM and TOV, \textcolor{green}{POM means favorable and TOV means unfavorable}. (continues as above)
\end{minipage}}
\end{minipage}
&
\begin{minipage}[t]{\linewidth}
\centering
\textbf{Board Example}\\[0.8mm]
\includegraphics[width=0.2\linewidth]{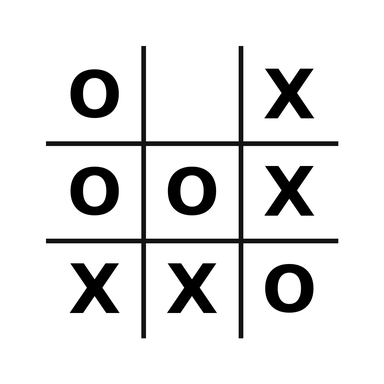}\\[1.5mm]
\textbf{Result}\\[0.8mm]
\includegraphics[width=0.85\linewidth]{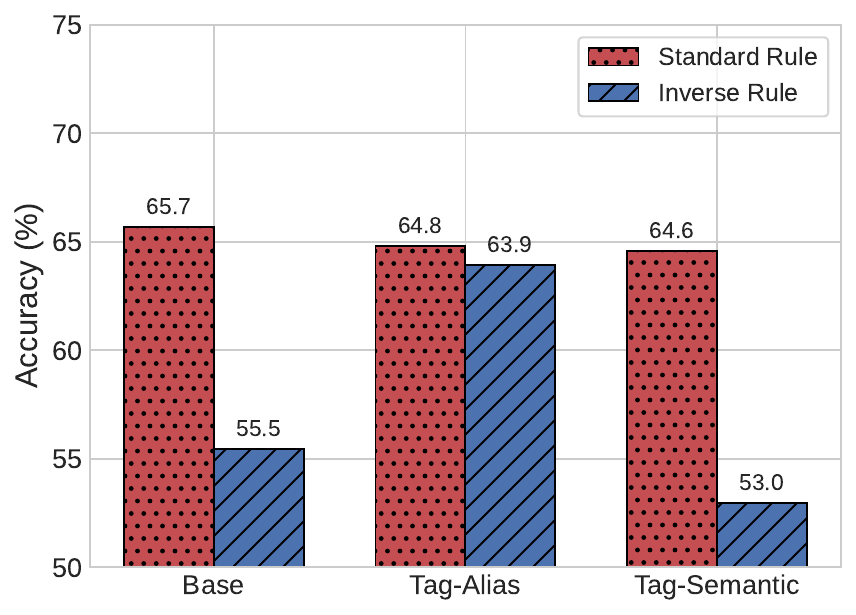}
\end{minipage}
\end{tabular}
}
\caption{Tic-Tac-Toe example from VLM-Fix. Left: base, tag-alias, and tag-semantic prompt variants. Right-top: terminal board state. Right-bottom: accuracy trend across prompt variants. Blue marks standard-rule assignments, red marks inverse (mis\`ere) assignments, and green marks the added semantic definition in the tag-semantic prompt.}
\label{fig:intro_prompt_variants}
\end{figure*}

To investigate this hypothesis, we build \textit{VLM-Fix}, a synthetic benchmark spanning four abstract strategy games: Tic-Tac-Toe, Connect Four, Reversi, and Dots and Boxes. For each game, we evaluate identical terminal board states under both standard and inverse (mis\`ere-style) rules, so only the winner/loser semantics change. This design isolates interpretation from perception and yields a direct measure of semantic fixation. Across both open and closed VLMs, accuracy is consistently higher under standard rules, revealing a robust semantic-fixation gap. Prompt interventions further support this mechanism: replacing winner/loser terms with neutral tags substantially narrows the inverse-rule gap, while reintroducing semantic valence with the same tags restores it (Figure~\ref{fig:intro_prompt_variants}).

Beyond these inference-time prompt effects, we examine how post-training reshapes semantic fixation. Post-training with supervised fine-tuning (SFT) and reinforcement learning with verifiable rewards (RLVR) is strongly rule-aligned: training on one rule improves same-rule performance but hurts cross-rule transfer, whereas joint training improves generalization across both rules. Finally, late-layer activation steering recovers degraded performance, indicating that much of the error stems from late semantic readouts rather than failures in visual recognition. We also observe a similar trend on the \textit{VLMBias} benchmark~\citep{vo2025vision}, specifically on its counterfactual counting task, supporting external validity beyond synthetic games.

Our main findings are:
\vspace{-0.5em}
\begin{enumerate}
\item On \textit{VLM-Fix}, inverse rules reduce accuracy despite identical visual states, indicating semantic fixation under rule remapping.
\item Semantic framing strongly modulates this gap: neutral aliases reduce it, while semantically loaded aliases restore it.
\item Post-training is rule-dependent: SFT/RLVR improve same-rule performance but induce negative transfer to the opposite rule, while joint rule training improves generalization across both rules.
\item Late-layer activation steering partially recovers degraded inverse-rule performance, indicating that some of the error is editable at late semantic readout stages.
\item We observe the same qualitative pattern on \textit{VLMBias}~\citep{vo2025vision}, supporting external validity beyond the synthetic game setting.
\end{enumerate}

\section{Related Work}

Prior work shows that large vision-language models (VLMs) exhibit systematic bias and reliability limitations beyond aggregate benchmark scores. Social-bias evaluations report stereotypical and representational skew in multimodal outputs \citep{ruggeri2023multidimvlmbias,huang2025visbias}, while broader robustness studies find brittle behavior under distribution shift and weak grounding under challenging visual conditions \citep{yang2023dawn,liu2023mmbench}. Counterfactual and minimally perturbed benchmarks further expose this fragility: C-VQA reports large drops on counterfactual questions \citep{zhang2023tvoff}, VisMin shows weaknesses on counting and spatial edits \citep{awal2024vismin}, and SpatialEval finds reduced visual reliance when redundant text is present \citep{wang2024spatialeval}. Hallucination-focused analyses (e.g., POPE, HallusionBench) and representation-level studies also indicate persistent misalignment between visual evidence and model outputs \citep{li2023pope,guan2024hallusionbench,fu2025hidden}.

Fixation-style reasoning has been studied more directly in language models. Prior work links LLM behavior to the cognitive notion of fixation (Einstellung), showing persistent reliance on familiar but suboptimal mappings \citep{luchins1942mechanization,bilalic2008einstellung,alavi2023large}. Related findings on anchoring and inflexible reasoning similarly suggest that LLM predictions can remain tied to initial or familiar semantic frames \citep{huang2025anchoring,kim2025limitations}. Broader cognitive-bias evaluations report human-like intuitive biases in modern LLMs \citep{hagendorff2023humanlike}. However, these studies are predominantly text-only and do not isolate fixation in multimodal settings where perceptual evidence is held constant and only semantic mapping changes.

Bias-mitigation work in both LLMs and VLMs has explored decoding, prompting, and counterfactual training interventions. In language models, self-debiasing and causal-intervention methods reduce harmful continuations and biased reasoning \citep{schick2021self,wu2024decot,sun2024causal,xia2024aligning}. In vision-language models, counterfactual prompt learning and counterfactual data interventions aim to reduce spurious correlations between semantics and visual categories \citep{he2022cpl,howard2024socialcounterfactuals,vo2025vision}. Our study complements these lines by providing a controlled rule-switch setting that cleanly separates perception from semantic interpretation, and by showing that neutral alias prompts reduce the inverse-rule gap while semantically loaded aliases restore it.

\section{VLM-Fix: Dataset and Evaluation Setup}
\label{sec:setup}

We design VLM-Fix, a controlled synthetic benchmark for evaluating model behavior under matched visual states but different semantic interpretations. The benchmark comprises four abstract strategy games: Tic-Tac-Toe ($3\times3$), Reversi ($5\times5$), Connect Four ($4\times4$), and Dots and Boxes ($6\times6$). For each game, we generate 300 unique terminal base states, excluding draws and balancing the canonical winner evenly across the two players. Focusing on terminal states fixes the visual evidence at the end of play and avoids confounds from intermediate game trajectories; these same states are reused across all image and prompt settings. Full details of base-state generation are provided in Appendix~\ref{app:dataset-details}.

\paragraph{Rule conditions.}
Each example is evaluated under one of two rule conditions: \textbf{Standard}, corresponding to the canonical objective of the game, and \textbf{Inverse}, which reverses the usual winning condition. For each condition, we ask two complementary query types: the identity of the \emph{winner} and the identity of the \emph{loser}. This matched design controls for board complexity, allowing performance differences to be attributed to semantic interpretation rather than board-state difficulty. The inverse rules we consider are valid game variants~\citep{wikipedia_ttt_variants,solitairelab_reversi}, although they are likely to be much less represented in web-scale pretraining data than their canonical counterparts. Table~\ref{tab:vlmfix_rules} summarizes the rule definitions used in VLM-Fix.

\begin{table}[t]
\centering
\small
\caption{Standard and inverse rule conditions in VLM-Fix.}
\label{tab:vlmfix_rules}
\resizebox{0.98\linewidth}{!}{%
\begin{tabular}{p{0.15\linewidth}p{0.42\linewidth}p{0.42\linewidth}}
\toprule
\textbf{Game} & \textbf{Standard rule} & \textbf{Inverse rule} \\
\midrule
Tic-Tac-Toe & A player who gets 3-in-a-row wins. & A player who gets 3-in-a-row loses. \\
Connect Four & A player who gets 4-in-a-row wins. & A player who gets 4-in-a-row loses. \\
Reversi & The player with more pieces wins. & The player with fewer pieces wins. \\
Dots and Boxes & The player who claims more boxes wins. & The player who claims fewer boxes wins. \\
\bottomrule
\end{tabular}
}
\end{table}

\paragraph{Rendering variants.}
We use three visual rendering styles. \textbf{Base} uses the canonical board appearance and default player symbols. \textbf{Checkerboard} changes the board texture while keeping the canonical symbols unchanged. \textbf{Glyph} preserves the board layout but replaces canonical player symbols with randomly sampled alphabetic glyphs. For each $(\text{game}, \text{state})$ pair, glyph symbols are deterministically sampled from \texttt{A..Z}, excluding $\{X, O, A, B\}$ to avoid collisions with canonical game labels.

\begin{figure}[t]
\centering
\setlength{\fboxsep}{2pt}
\small
\fcolorbox{black!25}{FigureLavender}{%
\begin{minipage}[t]{0.97\linewidth}
\vspace{0.35mm}
\begin{minipage}[t]{0.16\linewidth}
\vspace{0pt}
\centering
\includegraphics[width=0.54\linewidth]{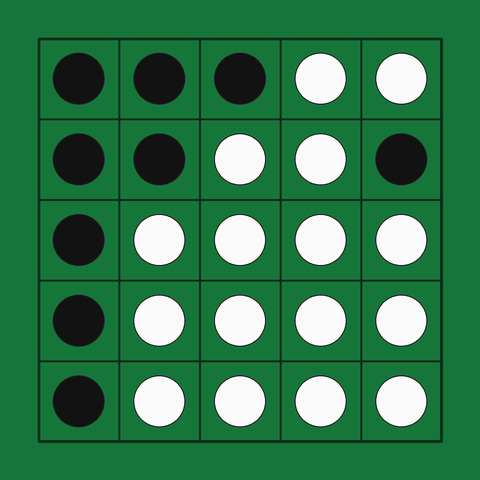}\\[0.7mm]
\includegraphics[width=0.54\linewidth]{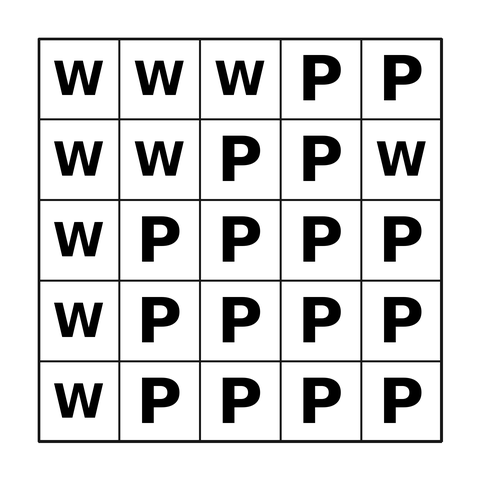}\\[0.7mm]
\includegraphics[width=0.54\linewidth]{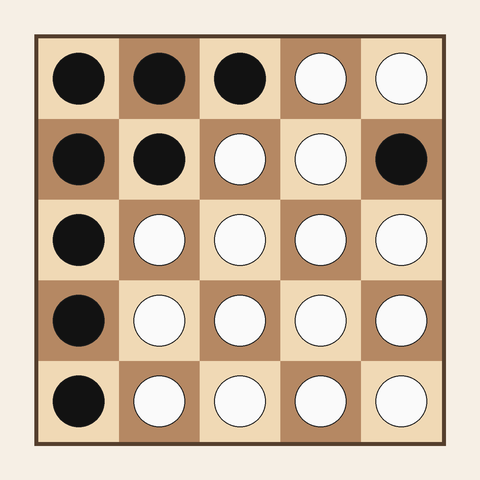}
\end{minipage}\hspace{0.002\linewidth}%
\color{black!45}\vrule width 0.6pt\color{black}%
\hspace{0.004\linewidth}%
\begin{minipage}[t]{0.80\linewidth}
\vspace{0pt}
\raggedright
\scriptsize
\textbf{Base:} You are given a 5x5 grid for a two-player game. Players are Black and White. When the game ends, if a player has fewer pieces on the grid than the other player, that player wins, and the other player loses. The game has ended. Who is the winner?\\[0.8mm]
\textcolor{black!55}{\rule{\linewidth}{0.3pt}}\\[-0.2mm]
\textbf{Alias:} You are given a 5x5 grid for a two-player game. Players are Black and White. Outcome tags: KAP and POM. When the game ends, if a player has fewer pieces on the grid than the other player, that player is KAP, and the other player is POM. The game has ended. Who is the KAP?\\[0.8mm]
\textcolor{black!55}{\rule{\linewidth}{0.3pt}}\\[-0.2mm]
\textbf{SemAlias:} You are given a 5x5 grid for a two-player game. Players are Black and White. Outcome tags: KAP and POM. KAP means favorable outcome; POM means unfavorable outcome. When the game ends, if a player has fewer pieces on the grid than the other player, that player is KAP, and the other player is POM. The game has ended. Who is the KAP?
\end{minipage}
\vspace{0.35mm}
\end{minipage}}
\caption{Compact Reversi example from VLM-Fix. Left: the three rendering variants, shown in order as the base, glyph, and checkerboard renderings used in the benchmark. Right: the three prompt variants that preserve the same underlying inverse-rule state while changing only semantic framing.}
\label{fig:vlm_fix_reversi_compact}
\end{figure}

\paragraph{Prompt variants.}
We use three prompt families to vary the semantic framing of the same visual input. Across all variants, we avoid naming the game verbatim in the prompt (e.g., ``Tic-Tac-Toe'' or ``Reversi'') to reduce lexical cues toward canonical rule interpretations. \textbf{Base} states the game rule directly and asks for the winner or loser using canonical outcome terms. \textbf{Alias} replaces canonical terms such as \emph{winner} and \emph{loser} with arbitrary tags. \textbf{SemAlias} retains these arbitrary tags but explicitly defines their meaning (e.g., \emph{favorable} versus \emph{unfavorable}). This design helps disentangle the effect of lexical familiarity from that of explicit semantic grounding. Figure~\ref{fig:vlm_fix_reversi_compact} shows a representative Reversi example across the three rendering variants and three prompt variants; corresponding examples for the other games are shown in Appendix Figure~\ref{fig:vlm_fix_examples}.

\paragraph{Dataset size.}
In the main experiments, we consider five configurations: \texttt{Base}, \texttt{Glyph}, \texttt{Checkerboard}, \texttt{Alias}, and \texttt{SemAlias}. \texttt{Base}/\texttt{Glyph}/\texttt{Checkerboard} share the same prompt template but differ in rendering, whereas \texttt{Base}/\texttt{Alias}/\texttt{SemAlias} share the canonical rendering but differ in prompt semantics. For each base state, we evaluate both rule conditions (\textbf{Standard} and \textbf{Inverse}), both query targets (\textbf{winner} and \textbf{loser}), and both multimodal orderings (\textbf{image-first} and \textbf{text-first}), yielding $300 \times 2 \times 2 \times 2 = 2400$ examples per game for each image--prompt configuration. Across the five main configurations, this results in 12{,}000 evaluated examples per game. We further evaluate each example under both \textbf{Direct} answering and \textbf{Chain-of-Thought (CoT)} prompting, yielding 24{,}000 evaluated examples per game. We additionally include descriptive-rule and text-only variants as control settings to isolate instruction-following effects from visual grounding: in the descriptive-rule variant, prompts ask directly about board properties (e.g., who forms the winning pattern or who has more/fewer pieces) rather than asking who is the winner or loser under a named rule, while in the text-only variant, the board is provided as an ASCII rendering without an image. Example prompts for both controls are shown in the Appendix.

\section{Experiment \& Results}

\subsection{Models}
We evaluate 14 VLMs spanning five model families and multiple parameter scales: GPT-4.1 and GPT-5.2 \citep{singh2025openai}; Sonnet-4.0 and Sonnet-4.5 \citep{anthropic2026models}; Qwen2.5-VL-3B, Qwen2.5-VL-7B, Qwen3-VL-4B, Qwen3-VL-8B, and Qwen3-VL-32B \citep{DBLP:journals/corr/abs-2502-13923,bai2025qwen3}; InternVL3.5-4B, InternVL3.5-8B, and InternVL3.5-14B \citep{wang2025internvl3}; and Molmo2-4B and Molmo2-8B \citep{clark2026molmo2}. We use greedy decoding (temperature 0) with a maximum generation budget of 1,024 tokens. To keep API evaluation cost manageable, the four closed-source models are evaluated on an aligned reduced subset of 300 states per game, whereas the open-weight models are evaluated on the full benchmark expansions.

\subsection{Results}
\label{sec:vlm_fix_results}
\paragraph{Baseline Performance.}
Table~\ref{tab:vlm_fix_direct_core} reports baseline results under \texttt{Base}, where prompts use canonical winner/loser semantics. Averaged across games and models, standard-rule accuracy is 67.1\% versus 52.5\% under inverse rules, a 14.6-point gap. Because standard and inverse evaluations reuse identical terminal boards, this drop cannot be attributed to changes in visual evidence; instead, it indicates difficulty with semantic rule remapping. The gap appears in all four games and is largest for Dots and Boxes (73.8\% vs. 50.0\%, a 23.8-point difference). At the model level, 13 of 14 models show lower inverse-rule accuracy than standard-rule accuracy; the only exception is Sonnet-4.5. Together, these baseline results provide the main empirical signal of semantic fixation: performance is strong when task semantics align with familiar priors, but degrades when the same perceptual state requires an alternative interpretation.

\begin{table*}[t]
\centering
\caption{VLM-Fix results under the baseline setting (\texttt{Base}; canonical rendering; direct answering). Each cell reports accuracy (\%) under the standard or inverse rule, aggregated over winner/loser question targets and image-first/text-first orders.}
\label{tab:vlm_fix_direct_core}
\resizebox{\textwidth}{!}{%
\begin{tabular}{l|cc|cc|cc|cc|cc}
\toprule
 & \multicolumn{2}{c|}{Tic-Tac-Toe} & \multicolumn{2}{c|}{Reversi} & \multicolumn{2}{c|}{Connect Four} & \multicolumn{2}{c|}{Dots and Boxes} & \multicolumn{2}{c}{Average} \\
Model & Std & Inv & Std & Inv & Std & Inv & Std & Inv & Std & Inv \\
\midrule
GPT-4.1 & 86.0 & 76.0 & 85.7 & 75.0 & 69.0 & 41.3 & 88.3 & 60.3 & 82.2 & 63.2 \\
GPT-5.2 & 93.7 & 86.0 & 95.7 & 91.0 & 69.0 & 53.0 & 93.7 & 75.7 & 88.0 & 76.4 \\
Sonnet-4.0 & 51.0 & 51.7 & 76.0 & 50.3 & 63.3 & 65.7 & 95.0 & 92.3 & 71.3 & 65.0 \\
Sonnet-4.5 & 50.0 & 56.0 & 85.7 & 67.0 & 53.0 & 86.7 & 78.3 & 77.0 & 66.8 & 71.7 \\
\hline
Qwen2.5-VL-3B & 54.2 & 49.9 & 57.2 & 50.0 & 51.7 & 50.6 & 51.7 & 44.2 & 53.7 & 48.7 \\
Qwen2.5-VL-7B & 66.5 & 44.7 & 61.3 & 41.5 & 58.9 & 48.2 & 65.8 & 32.6 & 63.1 & 41.8 \\
InternVL3.5-4B & 60.7 & 45.9 & 63.7 & 34.2 & 52.2 & 49.8 & 70.1 & 30.5 & 61.7 & 40.1 \\
InternVL3.5-8B & 73.8 & 57.6 & 63.2 & 45.2 & 54.9 & 49.0 & 72.2 & 55.6 & 66.0 & 51.9 \\
InternVL3.5-14B & 66.4 & 58.9 & 52.9 & 55.4 & 62.9 & 49.2 & 53.2 & 42.9 & 58.9 & 51.6 \\
Qwen3-VL-4B & 57.4 & 48.0 & 64.3 & 52.1 & 50.3 & 47.8 & 65.0 & 38.8 & 59.3 & 46.7 \\
Qwen3-VL-8B & 57.6 & 47.7 & 69.8 & 42.4 & 51.4 & 51.6 & 76.2 & 36.7 & 63.8 & 44.6 \\
Qwen3-VL-32B & 70.0 & 66.4 & 83.4 & 56.9 & 54.6 & 51.3 & 73.7 & 55.9 & 70.4 & 57.6 \\
Molmo2-4B & 61.3 & 42.8 & 65.1 & 22.2 & 55.2 & 48.7 & 71.7 & 22.7 & 63.3 & 34.1 \\
Molmo2-8B & 71.1 & 44.9 & 82.2 & 41.4 & 55.2 & 47.4 & 77.6 & 34.5 & 71.5 & 42.1 \\
\midrule
\textbf{Average} & 65.7 & 55.5 & 71.9 & 51.8 & 57.3 & 52.9 & 73.8 & 50.0 & 67.1 & 52.5 \\
\bottomrule
\end{tabular}%
}
\end{table*}

\paragraph{Input Interventions.}

Table~\ref{tab:vlm_fix_direct_interventions} summarizes how the main input interventions affect the average standard/inverse gap across the 14 models. Visual perturbations yield only modest changes: \texttt{Glyph} slightly improves inverse accuracy (54.95), while \texttt{Checkerboard} slightly lowers overall accuracy. In contrast, semantic framing interventions are much stronger: \texttt{Alias} raises inverse accuracy to 63.08 and reduces the standard--inverse gap to 2.29 points, the smallest among all settings. When semantic valence is reintroduced with \texttt{SemAlias}, inverse accuracy drops back to 53.51 and the gap reopens. This pattern aligns with our semantic-fixation account: neutral aliases reduce prior-driven bias, whereas semantically loaded tags restore it. Full game-wise breakdowns are reported in the Appendix.

\begin{table*}[t]
\centering
\caption{Average effect of input interventions on VLM-Fix direct results, aggregated over the 14 models. Columns report standard and inverse accuracy (\%) for each game and the macro-average across games.}
\label{tab:vlm_fix_direct_interventions}
\resizebox{0.75\textwidth}{!}{%
\begin{tabular}{l|cc|cc|cc|cc|cc}
\toprule
Config & \multicolumn{2}{c|}{Tic-Tac-Toe} & \multicolumn{2}{c|}{Reversi} & \multicolumn{2}{c|}{Connect Four} & \multicolumn{2}{c|}{Dots and Boxes} & \multicolumn{2}{c}{Average} \\
 & Std & Inv & Std & Inv & Std & Inv & Std & Inv & Std & Inv \\
\midrule
Base & \textbf{65.7} & 55.5 & 71.9 & 51.8 & 57.3 & 52.9 & 73.8 & 50.0 & 67.1 & 52.5 \\
Glyph & 65.2 & 54.6 & \textbf{74.1} & 53.3 & \textbf{61.8} & 56.9 & 70.3 & 54.9 & \textbf{67.8} & 54.9 \\
Checkerboard & 63.7 & 53.9 & 67.1 & 52.6 & 58.4 & 52.9 & \textbf{74.5} & 52.3 & 65.9 & 52.9 \\
Alias & 64.8 & \underline{\textbf{63.9}} & 68.3 & \underline{\textbf{66.6}} & 57.5 & \underline{\textbf{57.9}} & 70.9 & \underline{\textbf{63.9}} & 65.4 & \underline{\textbf{63.1}} \\
SemAlias & 64.6 & 53.0 & 71.6 & 54.7 & 58.3 & 53.5 & 70.3 & 52.9 & 66.2 & 53.5 \\
\bottomrule
\end{tabular}%
}
\end{table*}

\paragraph{Control Settings.}

Table~\ref{tab:vlm_fix_boundary_conditions} reports results for additional control settings. Descriptive prompting is substantially more balanced than the baseline (71.0/68.8), indicating that the standard--inverse gap in the base setup is not simply due to poor instruction following. Text-only evaluation still shows a clear inverse drop (69.2/58.6), indicating that the effect is largely text-driven and persists even without image input.

CoT improves overall accuracy relative to direct prompting, but the same qualitative pattern remains. Under \texttt{Base}, a sizable standard--inverse gap persists (85.0 vs.\ 74.1). \texttt{Alias} largely closes this gap (84.8/84.1), whereas \texttt{SemAlias} partially reopens it (85.5/77.3), consistent with the direct-answering results.

\begin{table*}[t]
\centering
\caption{Boundary-condition results on VLM-Fix. Columns report standard and inverse accuracy (\%) for each game and the macro-average across games.}
\label{tab:vlm_fix_boundary_conditions}
\resizebox{0.75\textwidth}{!}{%
\begin{tabular}{l|cc|cc|cc|cc|cc}
\toprule
Setting & \multicolumn{2}{c|}{Tic-Tac-Toe} & \multicolumn{2}{c|}{Reversi} & \multicolumn{2}{c|}{Connect Four} & \multicolumn{2}{c|}{Dots and Boxes} & \multicolumn{2}{c}{Average} \\
 & Std & Inv & Std & Inv & Std & Inv & Std & Inv & Std & Inv \\
\midrule
Direct (Base) & 65.7 & 55.5 & 71.9 & 51.8 & 57.3 & 52.9 & 73.8 & 50.0 & 67.1 & 52.5 \\
CoT (Base) & \textbf{85.6} & 76.4 & 91.9 & 74.1 & 83.3 & 79.2 & 79.2 & 66.5 & 85.0 & 74.1 \\
CoT (Alias) & 84.2 & \underline{\textbf{84.8}} & 91.3 & \underline{\textbf{90.6}} & \textbf{84.4} & \underline{\textbf{83.9}} & 79.4 & 77.3 & 84.8 & \underline{\textbf{84.1}} \\
CoT (SemAlias) & 84.0 & 80.8 & \textbf{92.3} & 77.0 & \textbf{84.4} & 80.3 & 81.3 & 71.0 & \textbf{85.5} & 77.3 \\
Descriptive & 65.5 & 64.4 & 78.0 & 74.1 & 59.1 & 56.0 & \textbf{81.3} & \underline{\textbf{80.7}} & 71.0 & 68.8 \\
Text-only & 69.8 & 58.5 & 70.9 & 59.7 & 64.9 & 57.1 & 71.2 & 59.2 & 69.2 & 58.6 \\
\bottomrule
\end{tabular}%
}
\end{table*}

\paragraph{Input Order.}
We also compare image-first and text-first orderings for the 10 open-weight models; full marginals are reported in Appendix Tables~\ref{tab:vlm_fix_direct_order_marginals_open} and~\ref{tab:vlm_fix_cot_order_marginals_open}. Under direct prompting with \texttt{Base}, text-first amplifies the standard--inverse gap from 13.2 points (60.2/47.0) to 21.4 points (66.2/44.8). Under CoT, this order effect largely disappears, with similar standard--inverse gaps for image-first and text-first prompting.


\section{External Validity on \textit{VLMBias}}

VLM-Fix is designed to isolate semantic remapping cleanly rather than to maximize naturalism, which lets us derive a broader intervention hypothesis: if prior-driven failures are partly sustained by familiar image--prompt associations, then defamiliarizing either side of the pair should reduce error even in tasks without an explicit standard/inverse rule switch. We therefore evaluate on \textit{VLMBias}~\citep{vo2025vision}, a counterfactual-image benchmark designed to expose prior-driven errors, not as a second direct measurement of semantic fixation itself, but as an external transfer test of whether VLM-Fix-motivated defamiliarization interventions carry over to a more natural prior-sensitive setting. We focus on four counting subsets (\textbf{Animals}, \textbf{Logos}, \textbf{Flags}, and \textbf{Game Boards}).

\noindent\textbf{Image and Prompt Interventions.} We consider three interventions relative to \texttt{Base}: (1) \texttt{Flip}, which vertically flips the image while keeping the prompt unchanged; (2) \texttt{Alias}, which replaces the target object term (e.g., ``animal'') with a generic \texttt{ITEM} token (one ITEM = one counted object) while keeping the image unchanged; and (3) \texttt{Flip+Alias}, which combines both. \texttt{Flip} creates visually defamiliarized inputs, especially for photo-realistic \textbf{Animals} and \textbf{Logos}, while preserving task-relevant pixels and full scene complexity. Unlike the setup in~\citep{vo2025vision}, which removes backgrounds (and can reduce perceptual difficulty), \texttt{Flip} primarily alters orientation cues. \texttt{Alias} is motivated by VLM-Fix findings that reducing semantically loaded priors reduces bias; because counting datasets typically lack explicit rule semantics, it tests whether lexical defamiliarization alone can reduce prior-driven errors. Figure~\ref{fig:biased_examples} shows a representative Animals example; examples for the other subsets appear in the Appendix.

\begin{figure}[t]
\centering
\setlength{\fboxsep}{2pt}
\small
\fcolorbox{black!25}{FigureLavender}{%
\begin{minipage}[t]{0.97\linewidth}
\vspace{1.0mm}
\begin{minipage}[t]{0.18\linewidth}
\vspace{0pt}
\centering
\includegraphics[width=0.94\linewidth]{images/vlms-biased/03_Animals_animal_020_alpaca_notitle_Q1_px384__original.png}
\end{minipage}\hspace{0.006\linewidth}%
\color{black!45}\vrule width 0.6pt\color{black}%
\hspace{0.006\linewidth}%
\begin{minipage}[t]{0.18\linewidth}
\vspace{0pt}
\centering
\includegraphics[width=0.94\linewidth]{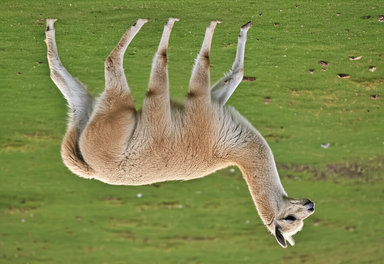}
\end{minipage}\hspace{0.008\linewidth}%
\color{black!45}\vrule width 0.6pt\color{black}%
\hspace{0.010\linewidth}%
\begin{minipage}[t]{0.56\linewidth}
\vspace{0pt}
\raggedright
\scriptsize
\textbf{Base:} Count the legs of this animal. Answer with a number in curly brackets, e.g., \{9\}.\\[0.2mm]
\textcolor{black!55}{\rule{\linewidth}{0.35pt}}\\[0.2mm]
\textbf{Alias:} Count the visible legs in this image, but report them as ITEMs (one ITEM = one visible leg). How many ITEMs are there? Answer with a number in curly brackets, e.g., \{9\}.
\end{minipage}
\vspace{1.0mm}
\end{minipage}}
\caption{Representative Animals example from \textit{VLMBias}. Left: the Base and Flip images. Right: the corresponding Base and Alias prompts.}
\label{fig:biased_examples}
\end{figure}

\noindent\textbf{Results.} Table~\ref{tab:vlms_are_biased_acc_bias} summarizes aggregate \textit{VLMBias} performance across the four subsets, averaged over the 14 evaluated models. Relative to \texttt{Base}, both single perturbations help: \texttt{Flip} improves overall accuracy from 11.6\% to 13.3\% while reducing bias from 76.7\% to 70.9\%, and \texttt{Alias} further improves these to 15.0\% and 69.5\%, respectively. The combined setting \texttt{Flip+Alias} is strongest overall, reaching 20.7\% accuracy and 58.9\% bias, with the largest task-level gain in Animals (3.6\% $\rightarrow$ 22.2\%). Crucially, this pattern holds without exception: in the pooled summary, \texttt{Flip+Alias} improves accuracy and reduces bias for all 14 models. Model-wise and subset-wise results are in Appendix~\ref{app:vlmbias-results}.

\begin{table*}[t]
\centering
\caption{Aggregate \textit{VLMBias} results across the four evaluation subsets, averaged over the 14 models using the canonical row-level summary.}
\label{tab:vlms_are_biased_acc_bias}
\resizebox{0.75\textwidth}{!}{%
\begin{tabular}{l|cc|cc|cc|cc|cc}
\toprule
Config & \multicolumn{2}{c|}{Animals} & \multicolumn{2}{c|}{Logos} & \multicolumn{2}{c|}{Flags} & \multicolumn{2}{c|}{Game Boards} & \multicolumn{2}{c}{Overall} \\
 & Acc & Bias & Acc & Bias & Acc & Bias & Acc & Bias & Acc & Bias \\
\midrule
Base & 3.6 & 93.2 & 13.9 & 71.4 & 25.9 & 55.1 & 11.4 & 66.6 & 11.6 & 76.7 \\
Flip & 5.7 & 88.9 & 16.6 & 60.2 & 24.2 & 55.4 & 13.7 & 61.1 & 13.3 & 70.9 \\
Alias & 8.5 & 83.8 & 14.9 & 65.6 & 30.6 & 51.8 & 14.1 & 58.0 & 15.0 & 69.5 \\
Flip+Alias & 22.2 & 67.5 & 15.7 & 56.0 & 29.0 & 49.8 & 16.2 & 51.1 & 20.7 & 58.9 \\
\bottomrule
\end{tabular}%
}
\end{table*}

\section{Training Interventions and Transfer}
\label{sec:vlm-fix-posttrain}

We explore two post-training strategies, supervised fine-tuning (SFT) and reinforcement learning with verifiable rewards (RLVR) using GRPO \citep{shao2024deepseekmath,guo2025deepseek}, across two transfer settings: rule transfer within VLM-Fix and synthetic leg-count transfer to \textit{VLMBias}. These experiments test whether targeted post-training can improve robustness under counterfactual task formulations, and whether such gains transfer across rules, games, and benchmarks.

\subsection{VLM-Fix Rule-Transfer Splits}

We evaluate SFT and RLVR on three VLM-Fix transfer splits, D1--D3, designed to probe cross-rule and cross-game generalization. D1 trains on standard-rule supervision and its canonical test file contains inverse-rule examples, D2 reverses this direction, and D3 trains on Tic-Tac-Toe and Reversi with both rules before testing on Connect Four and Dots and Boxes with both rules. For D1 and D2, we also evaluate the complementary held-out same-rule split, so comparisons include both same-rule and cross-rule transfer under the original image and base prompt. Full split construction, training hyperparameters, and evaluation details are provided in Appendix~\ref{sec:appendix_vlmfix_posttrain_details}.

\begin{figure}[t]
\centering
\begin{minipage}{0.90\columnwidth}
\centering
\includegraphics[width=0.48\linewidth]{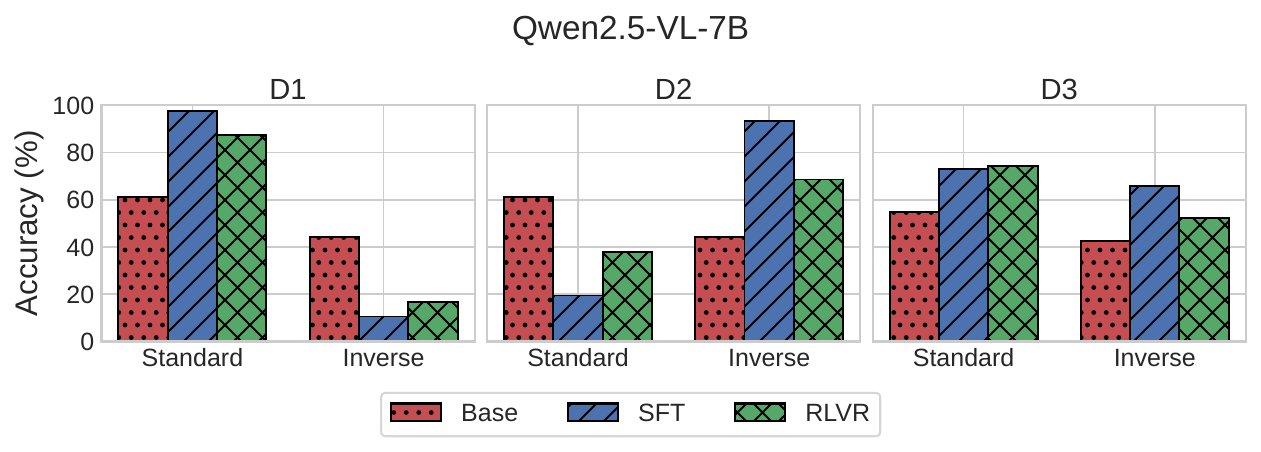}\hfill
\includegraphics[width=0.48\linewidth]{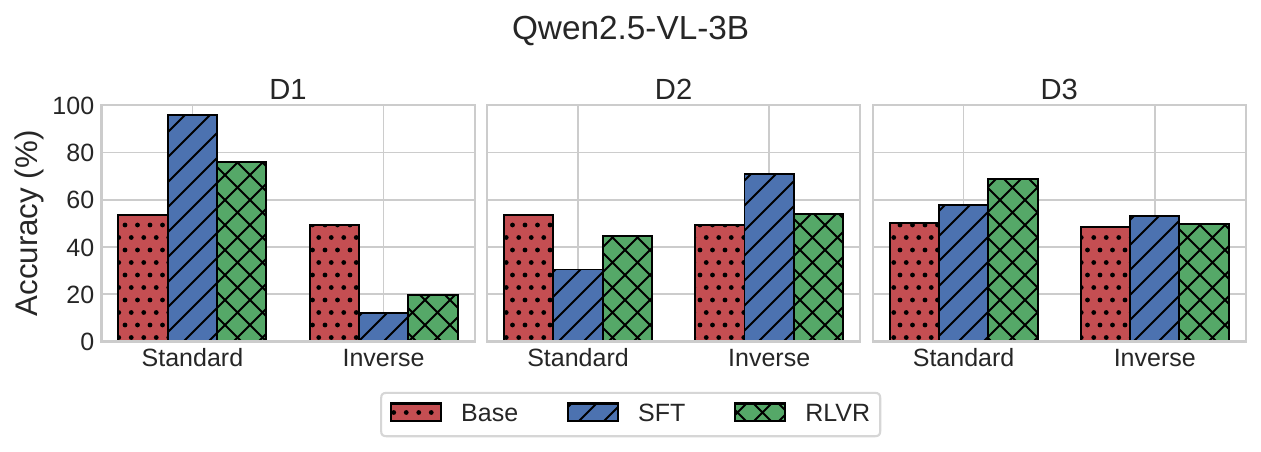}
\caption{VLM-Fix splits (D1--D3) for Qwen2.5-VL-7B (left) and Qwen2.5-VL-3B (right): D1/D2 report same-rule and opposite-rule evaluation, and D3 reports held-out standard/inverse evaluation on Connect Four and Dots and Boxes.}
\label{fig:vlm_fix_posttrain_d123_qwen}
\end{minipage}
\end{figure}

\paragraph{Results.}
Figure~\ref{fig:vlm_fix_posttrain_d123_qwen} shows a consistent same-rule versus cross-rule pattern. On D1 and D2, post-training improves held-out accuracy when evaluation matches the training rule (with SFT typically strongest), but performance drops below the base model when the rule mapping is flipped; RLVR is less brittle than SFT under this shift. On D3, where training covers both rules on source games, both methods improve transfer to held-out Connect Four and Dots and Boxes, with RLVR stronger on standard-rule evaluation and SFT stronger on inverse-rule evaluation (especially for Qwen2.5-VL-7B). Additional SFT D1--D3 transfer results for Molmo2-4B and InternVL3.5-4B are reported in Appendix Figure~\ref{fig:appendix_posttrain_additional_d123_abs}. Overall, post-training is strongly rule-conditional: same-rule transfer is strong, cross-rule transfer is negative, and cross-game transfer improves when both semantic mappings are seen during training.

\begin{figure}[t]
\centering
\begin{minipage}{0.90\columnwidth}
\centering
\includegraphics[width=0.48\linewidth]{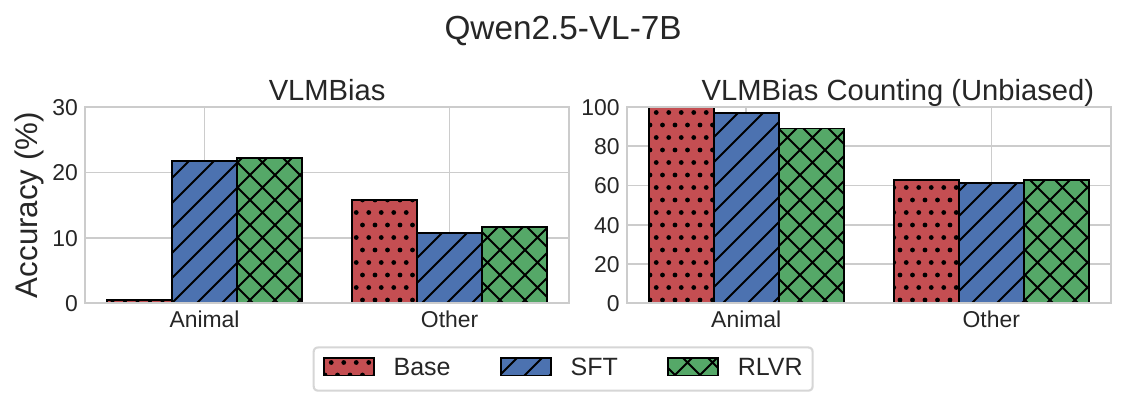}\hfill
\includegraphics[width=0.48\linewidth]{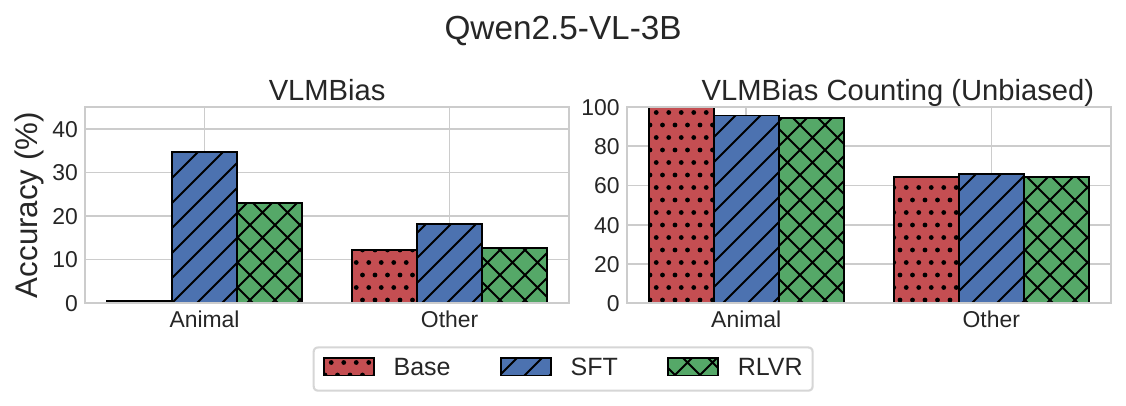}
\caption{Synthetic leg-count transfer for Qwen2.5-VL-7B (left) and Qwen2.5-VL-3B (right), comparing Base, SFT, and RLVR on \textit{VLMBias} and \textit{VLMBias Counting (Unbiased)} (Animals vs Other).}
\label{fig:vlm_bias_posttrain_d123_qwen}
\end{minipage}
\end{figure}

\subsection{Synthetic Leg-Count Transfer to VLMBias}
\label{sec:vlm-fix-synth-posttrain-main}

We next test transfer from a synthetic leg-counting dataset of procedurally rendered bird and quadruped glyphs. We post-train the same Qwen2.5-VL-3B and Qwen2.5-VL-7B backbones with SFT and RLVR, then evaluate on \textit{VLMBias} (Animals vs Other) and \textit{VLMBias Counting (Unbiased)} (Animals vs Other). Example synthetic glyph categories are shown in Appendix Figure~\ref{fig:appendix_synthleg_gallery}. Full dataset construction and training details are provided in Appendix~\ref{sec:appendix_synthlegs_posttrain_details}.

\paragraph{Results.}
Figure~\ref{fig:vlm_bias_posttrain_d123_qwen} shows that transfer from synthetic leg-count supervision is concentrated on the challenging \textit{VLMBias} Animals slice. Both Qwen models start at 0\% there, then rise to 34.6\%/23.1\% (SFT/RLVR) for Qwen2.5-VL-3B and 21.8\%/22.2\% for Qwen2.5-VL-7B. Gains on the \textit{VLMBias} Other slice are smaller and less consistent. By contrast, \textit{VLMBias Counting (Unbiased)} is already easy for the base models (Animals near ceiling, Other around the mid-60s), so post-training yields only minor changes and can slightly reduce accuracy. Additional synthetic-leg-count transfer results for Molmo2-4B and InternVL3.5-4B are reported in Appendix Figure~\ref{fig:appendix_posttrain_additional_vlmsbias_transfer}. Overall, synthetic leg-count training can debias models on the targeted animal leg-count task, but this benefit does not generalize to other objects.

\section{Activation Steering Analysis}

\subsection{VLM-Fix steering}
\paragraph{Setup.}
We study whether late-layer activation steering can edit rule-sensitive behavior on VLM-Fix without retraining. In the main analysis, donor and target examples come from the same game under canonical rendering, base prompts, and direct answering. We patch one of the final 12 decoder layers at the query token during prompt processing and steer the target activation toward a matched donor representation. Matching is determined by a lightweight router that predicts the relevant rule/answer bucket before patching. Full protocol details, split construction, and routing definitions are provided in Appendix~\ref{app:steering_details_vlmfix}.

\begin{figure*}[t]
\centering
\includegraphics[width=0.79\textwidth]{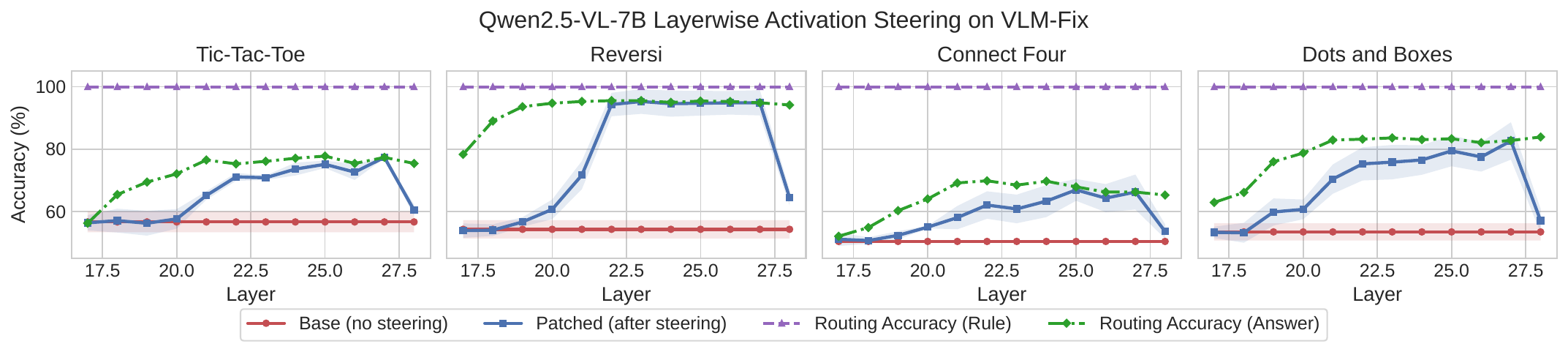}
\caption{Layerwise activation-steering results on VLM-Fix for Qwen2.5-VL-7B across Tic-Tac-Toe, Reversi, Connect Four, and Dots and Boxes (left to right).}
\label{fig:mech_vlmfix_qwen7b_4games}
\end{figure*}

\paragraph{Results.}
Figure~\ref{fig:mech_vlmfix_qwen7b_4games} shows that late-layer steering improves patched accuracy when routing is reliable, with the strongest gains in Reversi and Dots and Boxes. Gains are concentrated in later layers, where rule and answer routing are more stable, and are limited in harder settings such as Connect Four. Additional VLM-Fix steering results for Molmo2 and InternVL models are reported in Appendix~\ref{app:activation_steering_details}. Overall, the results indicate that rule-dependent representations are editable, but only when donor routing is accurate.

\subsection{VLMBias Animals steering}
\paragraph{Setup.}
We run a parallel donor-based steering analysis on the Animals subset of VLMBias in a controlled 2-leg versus 3-leg setting. Because base models are near 0\% on the 3-leg slice, they do not provide reliable donor traces; we therefore use the SFT models trained on synthetic leg-count as the donor and steer the corresponding base model across the final 12 decoder layers using held-out train/test splits. As in the VLM-Fix analysis, steering is routed through a simple classifier before the patched generation step. Full dataset construction, split details, and steering protocol are given in Appendix~\ref{app:steering_details_vlmbias}.

\begin{figure*}[t]
\centering
\includegraphics[width=0.79\textwidth]{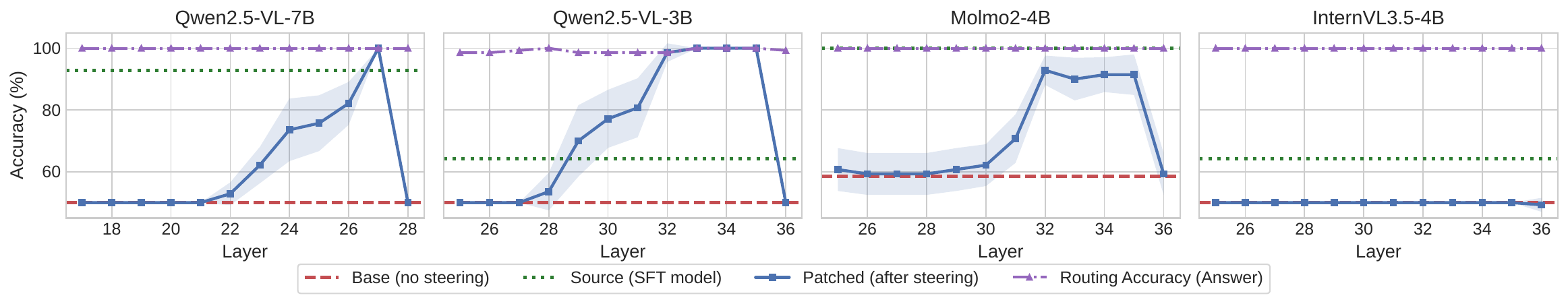}
\caption{Layerwise SFT$\rightarrow$Base activation-steering results on \textit{VLMBias} Animals for Qwen2.5-VL-7B, Qwen2.5-VL-3B, Molmo2-4B, and InternVL3.5-4B (left to right).}
\label{fig:mech_vlmbias_qwen_synth}
\end{figure*}

\paragraph{Results.}
Figure~\ref{fig:mech_vlmbias_qwen_synth} shows that donor-based steering can improve patched accuracy, with the clearest late-layer gains in the Qwen models. Additional VLMBias Animals steering results for Molmo2-4B and InternVL3.5-4B are reported in Appendix~\ref{app:activation_steering_details}. Overall, the results indicate that the relevant counting representation is partly editable, but its accessibility varies across architectures and depends on having reliable donor traces (here provided by SFT).

\section{Discussion and Limitations}

In this work, we study semantic fixation in a controlled setting where perception is held fixed and only rule semantics change. Across games and models in VLM-Fix, identical boards yield a consistent standard--inverse gap, and prompt interventions strengthen this account: neutral aliasing narrows the gap, while semantically loaded aliases reopen it. Post-training results further show strong rule alignment, with robust same-rule gains and broader transfer primarily when both mappings are included during training. Consistent trends on \textit{VLMBias} suggest that defamiliarizing image--prompt associations can also help in a related prior-sensitive benchmark. At the same time, these conclusions are bounded by the evaluation settings, since VLM-Fix is synthetic by design and \textit{VLMBias} provides external support in a related but not identical task family. Activation-steering results indicate that the effect is editable in late layers, but developing a full mechanistic account of how semantic representations are formed and used during inference remains an important direction for future work.


\clearpage

\bibliography{main}
\bibliographystyle{Template-2026/colm2026_conference}

\clearpage

\appendix
\section{Dataset Details}
\label{app:dataset-details}

\subsection{Base-State Construction}
We construct 300 unique terminal states per game, exclude draws, and balance canonical winners (150/150).

\paragraph{Tic-Tac-Toe ($3\times3$).}
We sample legal alternating-play terminal boards with an exclusive single-line winner (exactly one winning line). To control terminal-pattern diversity, we enforce win-pattern quotas:
\begin{itemize}
  \item 100 horizontal-win states,
  \item 100 vertical-win states,
  \item 50 main-diagonal-win states,
  \item 50 anti-diagonal-win states.
\end{itemize}

\paragraph{Reversi ($5\times5$).}
We generate terminal states via random legal self-play from the standard center initialization, with pass handling applied whenever a player has no legal move and rollouts continued until a terminal board is reached. Draw states are removed.

\paragraph{Connect Four ($4\times4$).}
We generate terminal states via random legal drop dynamics (gravity-constrained column drops) and stop each trajectory at the first winning terminal. Boards in which both players have winning lines are removed. A retained terminal may contain multiple winning lines, but only for the same winner. Across the 300 retained states, winning-line orientation counts are 109 vertical, 105 horizontal, 44 main-diagonal, and 42 anti-diagonal.

\paragraph{Dots and Boxes ($6\times6$).}
Terminal boards are represented as fully assigned claimed-cell outcomes. These are synthetic terminal outcomes (not full edge-by-edge legal game trajectories). We enforce a non-zero margin constraint by sampling winner margins from
\(\{2,4,6,8,10,12\}\).

\subsection{Prompt-Output Formatting}
\begin{figure*}[t]
\centering
\setlength{\fboxsep}{2pt}
\setlength{\tabcolsep}{0pt}
\tiny
\newcommand{\vlmFixColOneBg}{purple!8}
\newcommand{\vlmFixColTwoBg}{green!8}
\newcommand{\vlmFixColThreeBg}{orange!10}
\newcommand{\vlmFixColFourBg}{blue!6}
\newlength{\vlmFixPanelHeight}
\setlength{\vlmFixPanelHeight}{13.4cm}
\begin{tabular}{@{}cccc@{}}
\fcolorbox{black!25}{\vlmFixColOneBg}{%
\begin{minipage}[t][\vlmFixPanelHeight][t]{0.236\linewidth}
\vspace{0pt}
\centering
\textbf{Tic-Tac-Toe}\\[0.6mm]
\textit{Base}\\[-0.2mm]
\includegraphics[height=1.35cm]{images/vlm-fix/S001__s001__standard__direct__winner.png}\\[0.4mm]
\textit{Glyph}\\[-0.2mm]
\includegraphics[height=1.35cm]{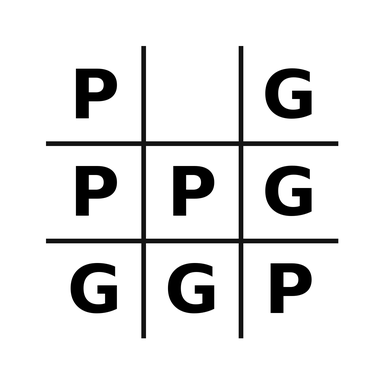}\\[0.4mm]
\textit{Checkerboard}\\[-0.2mm]
\includegraphics[height=1.35cm]{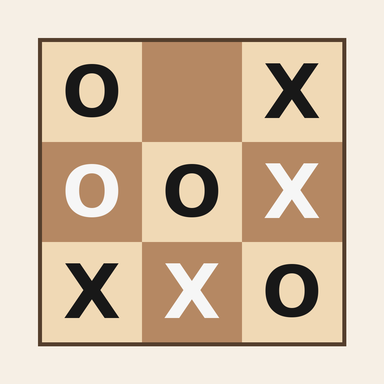}\\[-0.3mm]
\textcolor{black!55}{\rule{\linewidth}{0.35pt}}\\[0.5mm]
\raggedright
\textbf{Base:} You are given a 3x3 grid for a two-player game. Players are X and O. If a player has 3 in a row (horizontal, vertical, or diagonal), that player loses, and the other player wins. The game has ended. Who is the winner?\\[3.0mm]
\textcolor{black!55}{\rule{\linewidth}{0.35pt}}\\[0.6mm]
\textbf{Alias:} You are given a 3x3 grid for a two-player game. Players are X and O. Outcome tags: POM and TOV. If a player has 3 in a row (horizontal, vertical, or diagonal), that player is TOV, and the other player is POM. The game has ended. Who is the POM?\\[3.0mm]
\textcolor{black!55}{\rule{\linewidth}{0.35pt}}\\[0.6mm]
\textbf{SemAlias:} You are given a 3x3 grid for a two-player game. Players are X and O. Outcome tags: POM and TOV. POM means favorable outcome; TOV means unfavorable outcome. If a player has 3 in a row (horizontal, vertical, or diagonal), that player is TOV, and the other player is POM. The game has ended. Who is the POM?
\end{minipage}} &
\fcolorbox{black!25}{\vlmFixColTwoBg}{%
\begin{minipage}[t][\vlmFixPanelHeight][t]{0.236\linewidth}
\vspace{0pt}
\centering
\textbf{Reversi}\\[0.6mm]
\textit{Base}\\[-0.2mm]
\includegraphics[height=1.35cm]{images/vlm-fix/S049__s001__standard__direct__winner.png}\\[0.4mm]
\textit{Glyph}\\[-0.2mm]
\includegraphics[height=1.35cm]{images/vlm-fix/S065__s001__standard__direct__winner.png}\\[0.4mm]
\textit{Checkerboard}\\[-0.2mm]
\includegraphics[height=1.35cm]{images/vlm-fix/S057__s001__standard__direct__winner.png}\\[-0.3mm]
\textcolor{black!55}{\rule{\linewidth}{0.35pt}}\\[0.5mm]
\raggedright
\textbf{Base:} You are given a 5x5 grid for a two-player game. Players are Black and White. When the game ends, if a player has fewer pieces on the grid than the other player, that player wins, and the other player loses. The game has ended. Who is the winner?\\[0.5mm]
\textcolor{black!55}{\rule{\linewidth}{0.35pt}}\\[0.6mm]
\textbf{Alias:} You are given a 5x5 grid for a two-player game. Players are Black and White. Outcome tags: KAP and POM. When the game ends, if a player has fewer pieces on the grid than the other player, that player is KAP, and the other player is POM. The game has ended. Who is the KAP?\\[0.5mm]
\textcolor{black!55}{\rule{\linewidth}{0.35pt}}\\[0.6mm]
\textbf{SemAlias:} You are given a 5x5 grid for a two-player game. Players are Black and White. Outcome tags: KAP and POM. KAP means favorable outcome; POM means unfavorable outcome. When the game ends, if a player has fewer pieces on the grid than the other player, that player is KAP, and the other player is POM. The game has ended. Who is the KAP?
\end{minipage}} &
\fcolorbox{black!25}{\vlmFixColThreeBg}{%
\begin{minipage}[t][\vlmFixPanelHeight][t]{0.236\linewidth}
\vspace{0pt}
\centering
\textbf{Connect Four}\\[0.6mm]
\textit{Base}\\[-0.2mm]
\includegraphics[height=1.35cm]{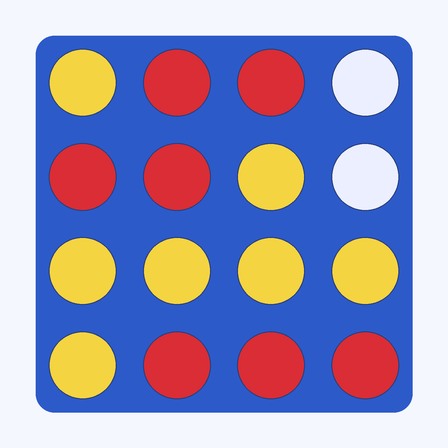}\\[0.4mm]
\textit{Glyph}\\[-0.2mm]
\includegraphics[height=1.35cm]{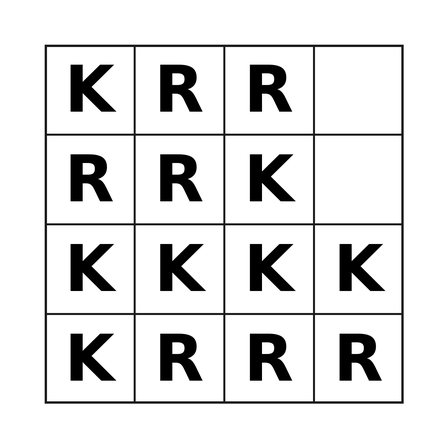}\\[0.4mm]
\textit{Checkerboard}\\[-0.2mm]
\includegraphics[height=1.35cm]{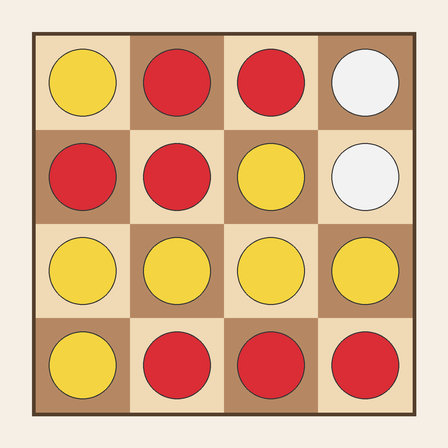}\\[-0.3mm]
\textcolor{black!55}{\rule{\linewidth}{0.35pt}}\\[0.5mm]
\raggedright
\textbf{Base:} You are given a 4x4 vertical grid for a two-player game. Players are Red and Yellow. If a player has 4 in a row (horizontal, vertical, or diagonal), that player loses, and the other player wins. The game has ended. Who is the winner?\\[0.5mm]
\textcolor{black!55}{\rule{\linewidth}{0.35pt}}\\[0.6mm]
\textbf{Alias:} You are given a 4x4 vertical grid for a two-player game. Players are Red and Yellow. Outcome tags: TOV and POM. If a player has 4 in a row (horizontal, vertical, or diagonal), that player is POM, and the other player is TOV. The game has ended. Who is the TOV?\\[0.5mm]
\textcolor{black!55}{\rule{\linewidth}{0.35pt}}\\[0.6mm]
\textbf{SemAlias:} You are given a 4x4 vertical grid for a two-player game. Players are Red and Yellow. Outcome tags: TOV and POM. TOV means favorable outcome; POM means unfavorable outcome. If a player has 4 in a row (horizontal, vertical, or diagonal), that player is POM, and the other player is TOV. The game has ended. Who is the TOV?
\end{minipage}} &
\fcolorbox{black!25}{\vlmFixColFourBg}{%
\begin{minipage}[t][\vlmFixPanelHeight][t]{0.236\linewidth}
\vspace{0pt}
\centering
\textbf{Dots and Boxes}\\[0.6mm]
\textit{Base}\\[-0.2mm]
\includegraphics[height=1.35cm]{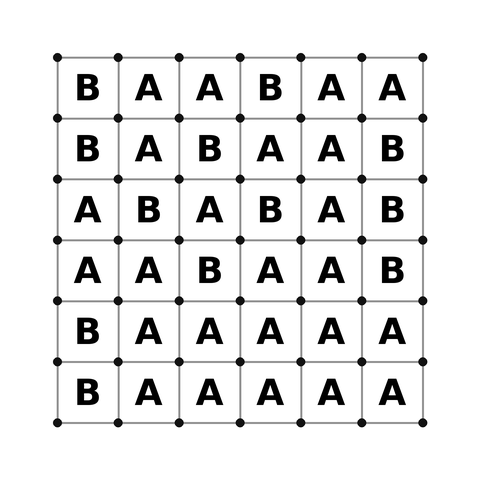}\\[0.4mm]
\textit{Glyph}\\[-0.2mm]
\includegraphics[height=1.35cm]{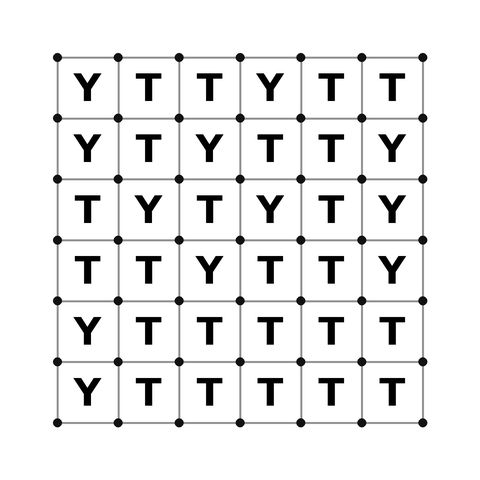}\\[0.4mm]
\textit{Checkerboard}\\[-0.2mm]
\includegraphics[height=1.35cm]{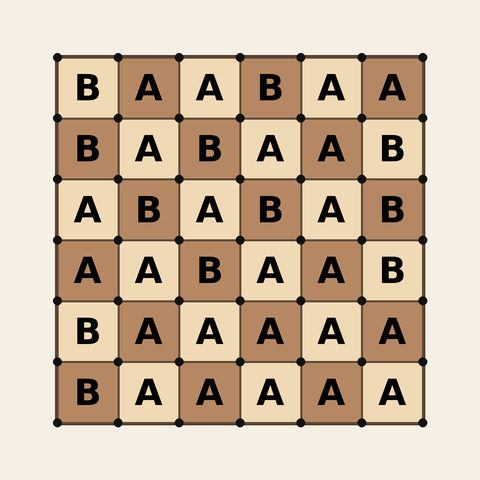}\\[-0.3mm]
\textcolor{black!55}{\rule{\linewidth}{0.35pt}}\\[0.5mm]
\raggedright
\textbf{Base:} You are given a 6x6 dot grid for a two-player game. Players are A and B. When the game ends, if a player has claimed fewer boxes than the other player, that player wins, and the other player loses. The game has ended. Who is the winner?\\[0.5mm]
\textcolor{black!55}{\rule{\linewidth}{0.35pt}}\\[0.6mm]
\textbf{Alias:} You are given a 6x6 dot grid for a two-player game. Players are A and B. Outcome tags: RIL and NEX. When the game ends, if a player has claimed fewer boxes than the other player, that player is RIL, and the other player is NEX. The game has ended. Who is the RIL?\\[0.5mm]
\textcolor{black!55}{\rule{\linewidth}{0.35pt}}\\[0.6mm]
\textbf{SemAlias:} You are given a 6x6 dot grid for a two-player game. Players are A and B. Outcome tags: RIL and NEX. RIL means favorable outcome; NEX means unfavorable outcome. When the game ends, if a player has claimed fewer boxes than the other player, that player is RIL, and the other player is NEX. The game has ended. Who is the RIL?
\end{minipage}}
\end{tabular}
\caption{Representative VLM-Fix inputs across four games. In each column, the top three images show rendering variants in the order \textit{Base} $\rightarrow$ \textit{Glyph} $\rightarrow$ \textit{Checkerboard}. Below the images, we show the three prompt variants used in this work: \textit{Base}, \textit{Alias}, and \textit{SemAlias}.}
\label{fig:vlm_fix_examples}
\end{figure*}

For all prompt families, we enforce explicit answer-format instructions by response variant:
Representative VLM-Fix rendering and prompt examples are shown in Figure~\ref{fig:vlm_fix_examples}.
\begin{itemize}
  \item \textbf{Direct:} ``Answer with only \verb|<label1>| or \verb|<label2>|. Do not add any other text.''
  \item \textbf{CoT:} ``Reason step by step. After that, give the answer inside \verb|\boxed{ }|.''
\end{itemize}

\begin{figure*}[t]
\centering
\setlength{\fboxsep}{1pt}
\setlength{\tabcolsep}{0pt}
\tiny
\newlength{\biasedHeaderRow}
\newlength{\biasedImageRow}
\newlength{\biasedPanelHeight}
\setlength{\biasedHeaderRow}{0.42cm}
\setlength{\biasedImageRow}{2.45cm}
\setlength{\biasedPanelHeight}{8.55cm}
\begin{tabular}{@{}cccc@{}}
\fcolorbox{black!25}{purple!8}{%
\begin{minipage}[t][\biasedPanelHeight][t]{0.254\linewidth}
\vspace{0pt}
\centering
\begin{minipage}[t][\biasedHeaderRow][c]{\linewidth}
\centering
\textbf{Animals}
\end{minipage}\\[-0.1mm]
\begin{minipage}[t][\biasedImageRow][c]{\linewidth}
\centering
\includegraphics[height=2.45cm]{images/vlms-biased/03_Animals_animal_020_alpaca_notitle_Q1_px384__original.png}
\end{minipage}\\[0.8mm]
\begin{minipage}[t][\biasedImageRow][c]{\linewidth}
\centering
\includegraphics[height=2.45cm]{images/vlms-biased/03_Animals_animal_020_alpaca_notitle_Q1_px384__flipped.png}
\end{minipage}\\[0.8mm]
\textcolor{black!55}{\rule{\linewidth}{0.35pt}}\\[0.6mm]
\raggedright\sloppy
\begin{minipage}[t]{\linewidth}
\textbf{Base:} Count the legs of this animal. Answer with a number in curly brackets, e.g., \{9\}.
\end{minipage}\\[0.5mm]
\textcolor{black!55}{\rule{\linewidth}{0.35pt}}\\[0.7mm]
\begin{minipage}[t]{\linewidth}
\textbf{Alias:} Count the visible legs in this image, but report them as ITEMs (one ITEM = one visible leg). How many ITEMs are there? Answer with a number in curly brackets, e.g., {9}.
\end{minipage}
\end{minipage}} &
\fcolorbox{black!25}{green!8}{%
\begin{minipage}[t][\biasedPanelHeight][t]{0.239\linewidth}
\vspace{0pt}
\centering
\begin{minipage}[t][\biasedHeaderRow][c]{\linewidth}
\centering
\textbf{Logos}
\end{minipage}\\[-0.1mm]
\begin{minipage}[t][\biasedImageRow][c]{\linewidth}
\centering
\includegraphics[height=2.45cm]{images/vlms-biased/01_Logos_car_059_notitle_Q1_px384__original.png}
\end{minipage}\\[0.8mm]
\begin{minipage}[t][\biasedImageRow][c]{\linewidth}
\centering
\includegraphics[height=2.45cm]{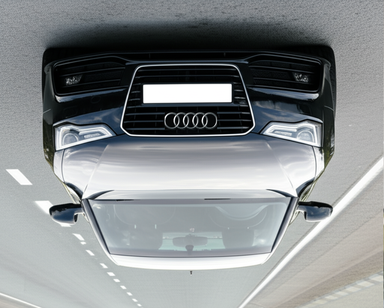}
\end{minipage}\\[0.8mm]
\textcolor{black!55}{\rule{\linewidth}{0.35pt}}\\[0.6mm]
\raggedright
\begin{minipage}[t]{\linewidth}
\textbf{Base:} Count the overlapping circles in the logo of this car. Answer with a number in curly brackets, e.g., \{9\}.
\end{minipage}\\[0.5mm]
\textcolor{black!55}{\rule{\linewidth}{0.35pt}}\\[0.7mm]
\begin{minipage}[t]{\linewidth}
\textbf{Alias:} Count the overlapping circles in the logo of this car, but report them as ITEMs (one ITEM = one overlapping circle). How many ITEMs are there? Answer with a number in curly brackets, e.g., \{9\}.
\end{minipage}
\end{minipage}} &
\fcolorbox{black!25}{orange!10}{%
\begin{minipage}[t][\biasedPanelHeight][t]{0.237\linewidth}
\vspace{0pt}
\centering
\begin{minipage}[t][\biasedHeaderRow][c]{\linewidth}
\centering
\textbf{Flags}
\end{minipage}\\[-0.1mm]
\begin{minipage}[t][\biasedImageRow][c]{\linewidth}
\centering
\includegraphics[height=2.45cm]{images/vlms-biased/01_Flags_flag_stars_020_notitle_Q1_px384__original.png}
\end{minipage}\\[0.8mm]
\begin{minipage}[t][\biasedImageRow][c]{\linewidth}
\centering
\includegraphics[height=2.45cm]{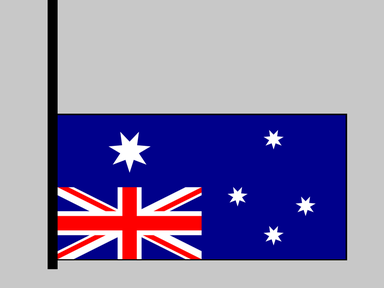}
\end{minipage}\\[0.8mm]
\textcolor{black!55}{\rule{\linewidth}{0.35pt}}\\[0.6mm]
\raggedright
\begin{minipage}[t]{\linewidth}
\textbf{Base:} Count the stars in this flag. Answer with a number in curly brackets, e.g., \{9\}.
\end{minipage}\\[0.5mm]
\textcolor{black!55}{\rule{\linewidth}{0.35pt}}\\[0.7mm]
\begin{minipage}[t]{\linewidth}
\textbf{Alias:} Count the stars in this image, but report them as ITEMs (one ITEM = one star). How many ITEMs are there? Answer with a number in curly brackets, e.g., \{9\}.
\end{minipage}
\end{minipage}} &
\fcolorbox{black!25}{blue!6}{%
\begin{minipage}[t][\biasedPanelHeight][t]{0.237\linewidth}
\vspace{0pt}
\centering
\begin{minipage}[t][\biasedHeaderRow][c]{\linewidth}
\centering
\textbf{Game Boards}
\end{minipage}\\[-0.1mm]
\begin{minipage}[t][\biasedImageRow][c]{\linewidth}
\centering
\includegraphics[height=2.45cm]{images/vlms-biased/05_Game_Boards_sudoku_grid_01_row_add_first_row_notitle_px384_Q1__original.png}
\end{minipage}\\[0.8mm]
\begin{minipage}[t][\biasedImageRow][c]{\linewidth}
\centering
\includegraphics[height=2.45cm]{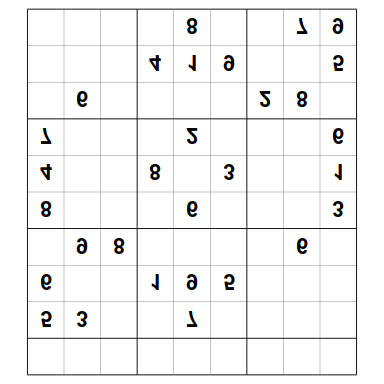}
\end{minipage}\\[0.8mm]
\textcolor{black!55}{\rule{\linewidth}{0.35pt}}\\[0.6mm]
\raggedright
\begin{minipage}[t]{\linewidth}
\textbf{Base:} Count the rows on this puzzle. Answer with a number in curly brackets, e.g., \{9\}.
\end{minipage}\\[0.5mm]
\textcolor{black!55}{\rule{\linewidth}{0.35pt}}\\[0.7mm]
\begin{minipage}[t]{\linewidth}
\textbf{Alias:} Count the rows on this image, but report them as ITEMs (one ITEM = one row). How many ITEMs are there? Answer with a number in curly brackets, e.g., \{9\}.
\end{minipage}
\end{minipage}}
\end{tabular}
\caption{Representative examples from the four \textit{VLMBias} subsets. Top row: Base images. Second row: Flip images. Third row: Base prompts. Fourth row: Alias prompts.}
\label{fig:biased_examples_appendix}
\end{figure*}

Representative VLMBias examples across Animals, Logos, Flags, and Game Boards are shown in Figure~\ref{fig:biased_examples_appendix}.

\section{Additional Results}

\subsection{Task-wise accuracy on VLM-Fix}

Across the four game-specific direct evaluations (Tables~\ref{tab:vlm_fix_direct_tictactoe}, \ref{tab:vlm_fix_direct_reversi}, \ref{tab:vlm_fix_direct_connect4}, and \ref{tab:vlm_fix_direct_dots_boxes}), Reversi and Dots and Boxes show the highest standard accuracy (74.07\% and 74.49\%), while Connect Four is consistently the hardest (57.26\%--61.79\%). Inverse-rule accuracy is lower than standard under Base, Glyph, Checkerboard, and SemAlias, but Alias consistently closes this gap across all games (Tic-Tac-Toe: 64.80/63.93, Reversi: 68.32/66.58, Connect Four: 57.48/57.90, Dots and Boxes: 70.88/63.89). Overall, neutral aliasing is the strongest intervention for mitigating rule-inversion sensitivity in VLM-Fix.

\begin{table*}[t]
\centering
\caption{VLM-Fix direct results on Tic-Tac-Toe. Cells report \textbf{standard/inverse accuracy (\%)}; subscripts in non-baseline columns give Holm-adjusted paired McNemar $p$-values versus Base for the same rule. Within each row, \textcolor{GapHighRed}{red} and \textcolor{GapLowBlue}{blue} mark the largest and smallest $(\mathrm{Standard}-\mathrm{Inverse})$ gaps.}
\label{tab:vlm_fix_direct_tictactoe}
\resizebox{\textwidth}{!}{%
\begin{tabular}{lccccc}
\toprule
Model & Base & Glyph & Checkerboard & Alias & SemAlias \\
 & Std/Inv (Acc) & Std/Inv (Acc) & Std/Inv (Acc) & Std/Inv (Acc) & Std/Inv (Acc) \\
\midrule
GPT-4.1 & 86.0/76.0 & 84.0$_{\scriptscriptstyle(\!\sim\!1.00)}$/60.3$_{\scriptscriptstyle(\!<\!0.01)}$ & 77.3$_{\scriptscriptstyle(\!\sim\!0.01)}$/61.3$_{\scriptscriptstyle(\!<\!0.01)}$ & \textcolor{GapLowBlue}{88.0$_{\scriptscriptstyle(\!\sim\!0.42)}$/85.3$_{\scriptscriptstyle(\!<\!0.01)}$} & \textcolor{GapHighRed}{87.0$_{\scriptscriptstyle(\!\sim\!1.00)}$/21.3$_{\scriptscriptstyle(\!<\!0.01)}$} \\
GPT-5.2 & 93.7/86.0 & 82.0$_{\scriptscriptstyle(\!<\!0.01)}$/70.7$_{\scriptscriptstyle(\!<\!0.01)}$ & \textcolor{GapHighRed}{89.0$_{\scriptscriptstyle(\!\sim\!0.20)}$/70.0$_{\scriptscriptstyle(\!<\!0.01)}$} & \textcolor{GapLowBlue}{94.0$_{\scriptscriptstyle(\!\sim\!1.00)}$/88.3$_{\scriptscriptstyle(\!\sim\!1.00)}$} & 95.3$_{\scriptscriptstyle(\!\sim\!1.00)}$/86.0$_{\scriptscriptstyle(\!\sim\!1.00)}$ \\
Sonnet-4.0 & \textcolor{GapLowBlue}{51.0/51.7} & \textcolor{GapHighRed}{62.3$_{\scriptscriptstyle(\!<\!0.01)}$/52.3$_{\scriptscriptstyle(\!\sim\!1.00)}$} & 54.3$_{\scriptscriptstyle(\!\sim\!0.10)}$/50.0$_{\scriptscriptstyle(\!\sim\!0.75)}$ & 57.7$_{\scriptscriptstyle(\!<\!0.01)}$/57.7$_{\scriptscriptstyle(\!\sim\!0.66)}$ & 59.3$_{\scriptscriptstyle(\!<\!0.01)}$/50.0$_{\scriptscriptstyle(\!\sim\!1.00)}$ \\
Sonnet-4.5 & \textcolor{GapLowBlue}{50.0/56.0} & \textcolor{GapHighRed}{61.0$_{\scriptscriptstyle(\!\sim\!0.01)}$/58.3$_{\scriptscriptstyle(\!\sim\!1.00)}$} & 50.0$_{\scriptscriptstyle(\!\sim\!1.00)}$/52.7$_{\scriptscriptstyle(\!\sim\!0.26)}$ & 54.3$_{\scriptscriptstyle(\!<\!0.01)}$/52.0$_{\scriptscriptstyle(\!\sim\!0.18)}$ & 52.3$_{\scriptscriptstyle(\!\sim\!0.06)}$/52.7$_{\scriptscriptstyle(\!\sim\!0.25)}$ \\
Qwen2.5-VL-3B & \textcolor{GapHighRed}{54.2/49.9} & 51.7$_{\scriptscriptstyle(\!\sim\!0.60)}$/48.9$_{\scriptscriptstyle(\!\sim\!1.00)}$ & 53.6$_{\scriptscriptstyle(\!\sim\!1.00)}$/50.0$_{\scriptscriptstyle(\!\sim\!1.00)}$ & \textcolor{GapLowBlue}{52.0$_{\scriptscriptstyle(\!\sim\!0.03)}$/51.9$_{\scriptscriptstyle(\!\sim\!0.37)}$} & 52.3$_{\scriptscriptstyle(\!\sim\!0.08)}$/51.2$_{\scriptscriptstyle(\!\sim\!0.73)}$ \\
Qwen2.5-VL-7B & \textcolor{GapHighRed}{66.5/44.7} & 59.5$_{\scriptscriptstyle(\!<\!0.01)}$/49.1$_{\scriptscriptstyle(\!\sim\!0.16)}$ & 65.2$_{\scriptscriptstyle(\!\sim\!1.00)}$/43.9$_{\scriptscriptstyle(\!\sim\!1.00)}$ & \textcolor{GapLowBlue}{65.8$_{\scriptscriptstyle(\!\sim\!1.00)}$/63.7$_{\scriptscriptstyle(\!<\!0.01)}$} & 64.2$_{\scriptscriptstyle(\!\sim\!0.23)}$/51.2$_{\scriptscriptstyle(\!<\!0.01)}$ \\
InternVL3.5-4B & \textcolor{GapHighRed}{60.7/45.9} & 60.2$_{\scriptscriptstyle(\!\sim\!1.00)}$/51.7$_{\scriptscriptstyle(\!<\!0.01)}$ & 54.4$_{\scriptscriptstyle(\!<\!0.01)}$/48.8$_{\scriptscriptstyle(\!\sim\!0.03)}$ & \textcolor{GapLowBlue}{60.1$_{\scriptscriptstyle(\!\sim\!1.00)}$/60.6$_{\scriptscriptstyle(\!<\!0.01)}$} & 57.2$_{\scriptscriptstyle(\!\sim\!0.02)}$/55.7$_{\scriptscriptstyle(\!<\!0.01)}$ \\
InternVL3.5-8B & 73.8/57.6 & 69.1$_{\scriptscriptstyle(\!<\!0.01)}$/57.4$_{\scriptscriptstyle(\!\sim\!1.00)}$ & 71.2$_{\scriptscriptstyle(\!\sim\!0.02)}$/57.8$_{\scriptscriptstyle(\!\sim\!1.00)}$ & \textcolor{GapLowBlue}{69.4$_{\scriptscriptstyle(\!<\!0.01)}$/66.3$_{\scriptscriptstyle(\!<\!0.01)}$} & \textcolor{GapHighRed}{71.2$_{\scriptscriptstyle(\!<\!0.01)}$/49.8$_{\scriptscriptstyle(\!<\!0.01)}$} \\
InternVL3.5-14B & 66.4/58.9 & 68.0$_{\scriptscriptstyle(\!\sim\!1.00)}$/58.9$_{\scriptscriptstyle(\!\sim\!1.00)}$ & \textcolor{GapHighRed}{65.2$_{\scriptscriptstyle(\!\sim\!1.00)}$/55.4$_{\scriptscriptstyle(\!\sim\!0.02)}$} & \textcolor{GapLowBlue}{67.2$_{\scriptscriptstyle(\!\sim\!1.00)}$/67.8$_{\scriptscriptstyle(\!<\!0.01)}$} & 68.3$_{\scriptscriptstyle(\!\sim\!0.19)}$/59.6$_{\scriptscriptstyle(\!\sim\!1.00)}$ \\
Qwen3-VL-4B & \textcolor{GapHighRed}{57.4/48.0} & 55.5$_{\scriptscriptstyle(\!\sim\!1.00)}$/48.3$_{\scriptscriptstyle(\!\sim\!1.00)}$ & 58.2$_{\scriptscriptstyle(\!\sim\!1.00)}$/51.4$_{\scriptscriptstyle(\!\sim\!0.05)}$ & \textcolor{GapLowBlue}{55.8$_{\scriptscriptstyle(\!\sim\!1.00)}$/57.6$_{\scriptscriptstyle(\!<\!0.01)}$} & 54.4$_{\scriptscriptstyle(\!\sim\!0.17)}$/52.8$_{\scriptscriptstyle(\!<\!0.01)}$ \\
Qwen3-VL-8B & 57.6/47.7 & \textcolor{GapHighRed}{65.8$_{\scriptscriptstyle(\!<\!0.01)}$/49.7$_{\scriptscriptstyle(\!\sim\!0.68)}$} & 55.8$_{\scriptscriptstyle(\!\sim\!0.09)}$/50.0$_{\scriptscriptstyle(\!\sim\!0.14)}$ & \textcolor{GapLowBlue}{54.7$_{\scriptscriptstyle(\!<\!0.01)}$/54.8$_{\scriptscriptstyle(\!<\!0.01)}$} & 54.6$_{\scriptscriptstyle(\!<\!0.01)}$/45.8$_{\scriptscriptstyle(\!\sim\!0.68)}$ \\
Qwen3-VL-32B & 70.0/66.4 & \textcolor{GapHighRed}{65.9$_{\scriptscriptstyle(\!<\!0.01)}$/61.3$_{\scriptscriptstyle(\!<\!0.01)}$} & 66.1$_{\scriptscriptstyle(\!<\!0.01)}$/64.9$_{\scriptscriptstyle(\!\sim\!0.28)}$ & \textcolor{GapLowBlue}{68.2$_{\scriptscriptstyle(\!\sim\!0.03)}$/70.0$_{\scriptscriptstyle(\!<\!0.01)}$} & 66.9$_{\scriptscriptstyle(\!<\!0.01)}$/62.3$_{\scriptscriptstyle(\!\sim\!0.02)}$ \\
Molmo2-4B & \textcolor{GapHighRed}{61.3/42.8} & 59.2$_{\scriptscriptstyle(\!\sim\!0.56)}$/48.2$_{\scriptscriptstyle(\!<\!0.01)}$ & 62.0$_{\scriptscriptstyle(\!\sim\!1.00)}$/52.3$_{\scriptscriptstyle(\!<\!0.01)}$ & \textcolor{GapLowBlue}{57.6$_{\scriptscriptstyle(\!<\!0.01)}$/56.7$_{\scriptscriptstyle(\!<\!0.01)}$} & 57.9$_{\scriptscriptstyle(\!<\!0.01)}$/51.6$_{\scriptscriptstyle(\!<\!0.01)}$ \\
Molmo2-8B & \textcolor{GapHighRed}{71.1/44.9} & 68.8$_{\scriptscriptstyle(\!\sim\!0.24)}$/49.8$_{\scriptscriptstyle(\!<\!0.01)}$ & 68.8$_{\scriptscriptstyle(\!\sim\!0.14)}$/46.6$_{\scriptscriptstyle(\!\sim\!0.45)}$ & \textcolor{GapLowBlue}{62.4$_{\scriptscriptstyle(\!<\!0.01)}$/62.3$_{\scriptscriptstyle(\!<\!0.01)}$} & 63.2$_{\scriptscriptstyle(\!<\!0.01)}$/51.7$_{\scriptscriptstyle(\!<\!0.01)}$ \\
\midrule
\textbf{Average} & 65.7/55.5 & 65.2/54.6 & 63.7/53.9 & 64.8/63.9 & 64.6/53.0 \\
\bottomrule
\end{tabular}%
}
\end{table*}

\begin{table*}[t]
\centering
\caption{VLM-Fix direct results on Reversi (same format as Table~\ref{tab:vlm_fix_direct_tictactoe}).}
\label{tab:vlm_fix_direct_reversi}
\resizebox{\textwidth}{!}{%
\begin{tabular}{lccccc}
\toprule
Model & Base & Glyph & Checkerboard & Alias & SemAlias \\
 & Std/Inv (Acc) & Std/Inv (Acc) & Std/Inv (Acc) & Std/Inv (Acc) & Std/Inv (Acc) \\
\midrule
GPT-4.1 & \textcolor{GapHighRed}{85.7/75.0} & 87.3$_{\scriptscriptstyle(\!\sim\!1.00)}$/77.7$_{\scriptscriptstyle(\!\sim\!0.72)}$ & 86.3$_{\scriptscriptstyle(\!\sim\!1.00)}$/79.3$_{\scriptscriptstyle(\!\sim\!0.71)}$ & 87.3$_{\scriptscriptstyle(\!\sim\!1.00)}$/80.0$_{\scriptscriptstyle(\!\sim\!0.04)}$ & \textcolor{GapLowBlue}{86.3$_{\scriptscriptstyle(\!\sim\!1.00)}$/81.3$_{\scriptscriptstyle(\!<\!0.01)}$} \\
GPT-5.2 & 95.7/91.0 & \textcolor{GapHighRed}{92.0$_{\scriptscriptstyle(\!\sim\!0.26)}$/84.3$_{\scriptscriptstyle(\!\sim\!0.03)}$} & 90.3$_{\scriptscriptstyle(\!\sim\!0.03)}$/86.7$_{\scriptscriptstyle(\!\sim\!0.18)}$ & \textcolor{GapLowBlue}{96.3$_{\scriptscriptstyle(\!\sim\!1.00)}$/93.0$_{\scriptscriptstyle(\!\sim\!1.00)}$} & 95.3$_{\scriptscriptstyle(\!\sim\!1.00)}$/88.7$_{\scriptscriptstyle(\!\sim\!1.00)}$ \\
Sonnet-4.0 & \textcolor{GapHighRed}{76.0/50.3} & 80.7$_{\scriptscriptstyle(\!\sim\!0.32)}$/55.3$_{\scriptscriptstyle(\!\sim\!0.64)}$ & 69.3$_{\scriptscriptstyle(\!\sim\!0.28)}$/51.3$_{\scriptscriptstyle(\!\sim\!0.91)}$ & 84.7$_{\scriptscriptstyle(\!\sim\!0.02)}$/71.3$_{\scriptscriptstyle(\!<\!0.01)}$ & \textcolor{GapLowBlue}{84.7$_{\scriptscriptstyle(\!\sim\!0.02)}$/78.0$_{\scriptscriptstyle(\!<\!0.01)}$} \\
Sonnet-4.5 & \textcolor{GapHighRed}{85.7/67.0} & \textcolor{GapLowBlue}{86.7$_{\scriptscriptstyle(\!\sim\!1.00)}$/81.3$_{\scriptscriptstyle(\!<\!0.01)}$} & 79.0$_{\scriptscriptstyle(\!\sim\!0.44)}$/69.7$_{\scriptscriptstyle(\!\sim\!0.40)}$ & 89.0$_{\scriptscriptstyle(\!\sim\!1.00)}$/82.3$_{\scriptscriptstyle(\!<\!0.01)}$ & 86.7$_{\scriptscriptstyle(\!\sim\!1.00)}$/75.7$_{\scriptscriptstyle(\!<\!0.01)}$ \\
Qwen2.5-VL-3B & 57.2/50.0 & \textcolor{GapHighRed}{57.7$_{\scriptscriptstyle(\!\sim\!1.00)}$/43.1$_{\scriptscriptstyle(\!<\!0.01)}$} & 56.4$_{\scriptscriptstyle(\!\sim\!1.00)}$/50.0$_{\scriptscriptstyle(\!\sim\!1.00)}$ & 58.0$_{\scriptscriptstyle(\!\sim\!1.00)}$/55.8$_{\scriptscriptstyle(\!<\!0.01)}$ & \textcolor{GapLowBlue}{54.5$_{\scriptscriptstyle(\!\sim\!0.07)}$/55.8$_{\scriptscriptstyle(\!<\!0.01)}$} \\
Qwen2.5-VL-7B & 61.3/41.5 & \textcolor{GapHighRed}{70.1$_{\scriptscriptstyle(\!<\!0.01)}$/40.4$_{\scriptscriptstyle(\!\sim\!1.00)}$} & 58.6$_{\scriptscriptstyle(\!\sim\!0.07)}$/44.6$_{\scriptscriptstyle(\!\sim\!0.05)}$ & \textcolor{GapLowBlue}{60.2$_{\scriptscriptstyle(\!\sim\!0.66)}$/63.1$_{\scriptscriptstyle(\!<\!0.01)}$} & 59.8$_{\scriptscriptstyle(\!\sim\!0.59)}$/41.2$_{\scriptscriptstyle(\!\sim\!1.00)}$ \\
InternVL3.5-4B & \textcolor{GapHighRed}{63.7/34.2} & 66.5$_{\scriptscriptstyle(\!\sim\!0.19)}$/40.5$_{\scriptscriptstyle(\!<\!0.01)}$ & 49.9$_{\scriptscriptstyle(\!<\!0.01)}$/42.7$_{\scriptscriptstyle(\!<\!0.01)}$ & \textcolor{GapLowBlue}{61.2$_{\scriptscriptstyle(\!\sim\!0.09)}$/56.4$_{\scriptscriptstyle(\!<\!0.01)}$} & 62.0$_{\scriptscriptstyle(\!\sim\!0.19)}$/40.0$_{\scriptscriptstyle(\!<\!0.01)}$ \\
InternVL3.5-8B & 63.2/45.2 & 72.9$_{\scriptscriptstyle(\!<\!0.01)}$/46.2$_{\scriptscriptstyle(\!\sim\!0.98)}$ & 54.5$_{\scriptscriptstyle(\!<\!0.01)}$/40.2$_{\scriptscriptstyle(\!<\!0.01)}$ & \textcolor{GapLowBlue}{55.2$_{\scriptscriptstyle(\!<\!0.01)}$/51.7$_{\scriptscriptstyle(\!<\!0.01)}$} & \textcolor{GapHighRed}{71.3$_{\scriptscriptstyle(\!<\!0.01)}$/42.5$_{\scriptscriptstyle(\!\sim\!0.09)}$} \\
InternVL3.5-14B & 52.9/55.4 & \textcolor{GapHighRed}{67.4$_{\scriptscriptstyle(\!<\!0.01)}$/56.3$_{\scriptscriptstyle(\!\sim\!1.00)}$} & \textcolor{GapLowBlue}{51.4$_{\scriptscriptstyle(\!<\!0.01)}$/56.0$_{\scriptscriptstyle(\!\sim\!1.00)}$} & 51.1$_{\scriptscriptstyle(\!<\!0.01)}$/55.3$_{\scriptscriptstyle(\!\sim\!1.00)}$ & 49.8$_{\scriptscriptstyle(\!<\!0.01)}$/54.3$_{\scriptscriptstyle(\!\sim\!1.00)}$ \\
Qwen3-VL-4B & 64.3/52.1 & 61.9$_{\scriptscriptstyle(\!\sim\!0.29)}$/51.7$_{\scriptscriptstyle(\!\sim\!1.00)}$ & 57.9$_{\scriptscriptstyle(\!<\!0.01)}$/47.2$_{\scriptscriptstyle(\!<\!0.01)}$ & \textcolor{GapLowBlue}{55.0$_{\scriptscriptstyle(\!<\!0.01)}$/54.9$_{\scriptscriptstyle(\!\sim\!0.38)}$} & \textcolor{GapHighRed}{58.1$_{\scriptscriptstyle(\!<\!0.01)}$/42.1$_{\scriptscriptstyle(\!<\!0.01)}$} \\
Qwen3-VL-8B & 69.8/42.4 & 71.1$_{\scriptscriptstyle(\!\sim\!1.00)}$/43.3$_{\scriptscriptstyle(\!\sim\!1.00)}$ & 59.7$_{\scriptscriptstyle(\!<\!0.01)}$/42.9$_{\scriptscriptstyle(\!\sim\!1.00)}$ & \textcolor{GapLowBlue}{58.9$_{\scriptscriptstyle(\!<\!0.01)}$/58.2$_{\scriptscriptstyle(\!<\!0.01)}$} & \textcolor{GapHighRed}{69.7$_{\scriptscriptstyle(\!\sim\!1.00)}$/41.3$_{\scriptscriptstyle(\!\sim\!1.00)}$} \\
Qwen3-VL-32B & 83.4/56.9 & 71.7$_{\scriptscriptstyle(\!<\!0.01)}$/59.5$_{\scriptscriptstyle(\!\sim\!0.24)}$ & 74.8$_{\scriptscriptstyle(\!<\!0.01)}$/59.0$_{\scriptscriptstyle(\!\sim\!0.24)}$ & \textcolor{GapLowBlue}{76.8$_{\scriptscriptstyle(\!<\!0.01)}$/90.9$_{\scriptscriptstyle(\!<\!0.01)}$} & \textcolor{GapHighRed}{79.2$_{\scriptscriptstyle(\!<\!0.01)}$/45.1$_{\scriptscriptstyle(\!<\!0.01)}$} \\
Molmo2-4B & 65.1/22.2 & \textcolor{GapHighRed}{72.5$_{\scriptscriptstyle(\!<\!0.01)}$/25.1$_{\scriptscriptstyle(\!\sim\!0.16)}$} & 68.7$_{\scriptscriptstyle(\!<\!0.01)}$/25.0$_{\scriptscriptstyle(\!<\!0.01)}$ & \textcolor{GapLowBlue}{48.5$_{\scriptscriptstyle(\!<\!0.01)}$/58.7$_{\scriptscriptstyle(\!<\!0.01)}$} & 61.7$_{\scriptscriptstyle(\!\sim\!0.01)}$/38.0$_{\scriptscriptstyle(\!<\!0.01)}$ \\
Molmo2-8B & 82.2/41.4 & 78.5$_{\scriptscriptstyle(\!\sim\!0.10)}$/41.2$_{\scriptscriptstyle(\!\sim\!1.00)}$ & 83.2$_{\scriptscriptstyle(\!\sim\!0.84)}$/42.2$_{\scriptscriptstyle(\!\sim\!1.00)}$ & \textcolor{GapLowBlue}{74.3$_{\scriptscriptstyle(\!<\!0.01)}$/60.5$_{\scriptscriptstyle(\!<\!0.01)}$} & \textcolor{GapHighRed}{83.3$_{\scriptscriptstyle(\!\sim\!0.84)}$/41.3$_{\scriptscriptstyle(\!\sim\!1.00)}$} \\
\midrule
\textbf{Average} & 71.9/51.8 & 74.1/53.3 & 67.1/52.6 & 68.3/66.6 & 71.6/54.7 \\
\bottomrule
\end{tabular}%
}
\end{table*}

\begin{table*}[t]
\centering
\caption{VLM-Fix direct results on Connect Four (same format as Table~\ref{tab:vlm_fix_direct_tictactoe}).}
\label{tab:vlm_fix_direct_connect4}
\resizebox{\textwidth}{!}{%
\begin{tabular}{lccccc}
\toprule
Model & Base & Glyph & Checkerboard & Alias & SemAlias \\
 & Std/Inv (Acc) & Std/Inv (Acc) & Std/Inv (Acc) & Std/Inv (Acc) & Std/Inv (Acc) \\
\midrule
GPT-4.1 & 69.0/41.3 & 74.0$_{\scriptscriptstyle(\!\sim\!1.00)}$/48.3$_{\scriptscriptstyle(\!\sim\!0.33)}$ & 73.3$_{\scriptscriptstyle(\!\sim\!1.00)}$/52.0$_{\scriptscriptstyle(\!\sim\!0.05)}$ & \textcolor{GapLowBlue}{69.7$_{\scriptscriptstyle(\!\sim\!1.00)}$/62.7$_{\scriptscriptstyle(\!<\!0.01)}$} & \textcolor{GapHighRed}{69.0$_{\scriptscriptstyle(\!\sim\!1.00)}$/36.3$_{\scriptscriptstyle(\!\sim\!0.35)}$} \\
GPT-5.2 & 69.0/53.0 & \textcolor{GapLowBlue}{71.0$_{\scriptscriptstyle(\!\sim\!1.00)}$/65.3$_{\scriptscriptstyle(\!<\!0.01)}$} & 69.0$_{\scriptscriptstyle(\!\sim\!1.00)}$/53.0$_{\scriptscriptstyle(\!\sim\!1.00)}$ & 68.0$_{\scriptscriptstyle(\!\sim\!1.00)}$/56.0$_{\scriptscriptstyle(\!\sim\!1.00)}$ & \textcolor{GapHighRed}{71.7$_{\scriptscriptstyle(\!\sim\!1.00)}$/51.7$_{\scriptscriptstyle(\!\sim\!1.00)}$} \\
Sonnet-4.0 & \textcolor{GapHighRed}{63.3/65.7} & 56.0$_{\scriptscriptstyle(\!\sim\!0.34)}$/69.7$_{\scriptscriptstyle(\!\sim\!1.00)}$ & 60.3$_{\scriptscriptstyle(\!\sim\!1.00)}$/68.0$_{\scriptscriptstyle(\!\sim\!1.00)}$ & 65.3$_{\scriptscriptstyle(\!\sim\!1.00)}$/78.3$_{\scriptscriptstyle(\!<\!0.01)}$ & \textcolor{GapLowBlue}{68.3$_{\scriptscriptstyle(\!\sim\!0.63)}$/82.7$_{\scriptscriptstyle(\!<\!0.01)}$} \\
Sonnet-4.5 & \textcolor{GapLowBlue}{53.0/86.7} & 74.0$_{\scriptscriptstyle(\!<\!0.01)}$/87.7$_{\scriptscriptstyle(\!\sim\!1.00)}$ & \textcolor{GapHighRed}{67.0$_{\scriptscriptstyle(\!<\!0.01)}$/72.7$_{\scriptscriptstyle(\!<\!0.01)}$} & 60.7$_{\scriptscriptstyle(\!<\!0.01)}$/77.3$_{\scriptscriptstyle(\!<\!0.01)}$ & 62.0$_{\scriptscriptstyle(\!<\!0.01)}$/92.7$_{\scriptscriptstyle(\!\sim\!0.04)}$ \\
Qwen2.5-VL-3B & 51.7/50.6 & 52.2$_{\scriptscriptstyle(\!\sim\!1.00)}$/49.7$_{\scriptscriptstyle(\!\sim\!1.00)}$ & \textcolor{GapLowBlue}{49.4$_{\scriptscriptstyle(\!\sim\!0.13)}$/50.2$_{\scriptscriptstyle(\!\sim\!1.00)}$} & 50.6$_{\scriptscriptstyle(\!\sim\!1.00)}$/49.6$_{\scriptscriptstyle(\!\sim\!1.00)}$ & \textcolor{GapHighRed}{50.3$_{\scriptscriptstyle(\!\sim\!1.00)}$/47.7$_{\scriptscriptstyle(\!\sim\!0.62)}$} \\
Qwen2.5-VL-7B & 58.9/48.2 & \textcolor{GapHighRed}{62.7$_{\scriptscriptstyle(\!\sim\!0.23)}$/50.7$_{\scriptscriptstyle(\!\sim\!0.88)}$} & 57.0$_{\scriptscriptstyle(\!\sim\!0.96)}$/48.5$_{\scriptscriptstyle(\!\sim\!1.00)}$ & \textcolor{GapLowBlue}{60.7$_{\scriptscriptstyle(\!\sim\!1.00)}$/57.8$_{\scriptscriptstyle(\!<\!0.01)}$} & 57.5$_{\scriptscriptstyle(\!\sim\!1.00)}$/47.9$_{\scriptscriptstyle(\!\sim\!1.00)}$ \\
InternVL3.5-4B & 52.2/49.8 & \textcolor{GapHighRed}{60.0$_{\scriptscriptstyle(\!<\!0.01)}$/50.4$_{\scriptscriptstyle(\!\sim\!1.00)}$} & 53.4$_{\scriptscriptstyle(\!\sim\!0.95)}$/49.8$_{\scriptscriptstyle(\!\sim\!1.00)}$ & \textcolor{GapLowBlue}{50.9$_{\scriptscriptstyle(\!\sim\!1.00)}$/50.6$_{\scriptscriptstyle(\!\sim\!1.00)}$} & 52.2$_{\scriptscriptstyle(\!\sim\!1.00)}$/49.1$_{\scriptscriptstyle(\!\sim\!1.00)}$ \\
InternVL3.5-8B & 54.9/49.0 & \textcolor{GapHighRed}{68.2$_{\scriptscriptstyle(\!<\!0.01)}$/55.9$_{\scriptscriptstyle(\!<\!0.01)}$} & 56.2$_{\scriptscriptstyle(\!\sim\!0.77)}$/50.2$_{\scriptscriptstyle(\!\sim\!1.00)}$ & \textcolor{GapLowBlue}{54.4$_{\scriptscriptstyle(\!\sim\!1.00)}$/54.1$_{\scriptscriptstyle(\!<\!0.01)}$} & 55.4$_{\scriptscriptstyle(\!\sim\!1.00)}$/48.7$_{\scriptscriptstyle(\!\sim\!1.00)}$ \\
InternVL3.5-14B & 62.9/49.2 & 68.8$_{\scriptscriptstyle(\!<\!0.01)}$/59.8$_{\scriptscriptstyle(\!<\!0.01)}$ & \textcolor{GapHighRed}{63.2$_{\scriptscriptstyle(\!\sim\!1.00)}$/47.5$_{\scriptscriptstyle(\!\sim\!0.07)}$} & \textcolor{GapLowBlue}{58.2$_{\scriptscriptstyle(\!<\!0.01)}$/59.6$_{\scriptscriptstyle(\!<\!0.01)}$} & 60.8$_{\scriptscriptstyle(\!\sim\!0.34)}$/46.5$_{\scriptscriptstyle(\!\sim\!0.21)}$ \\
Qwen3-VL-4B & 50.3/47.8 & 52.6$_{\scriptscriptstyle(\!\sim\!1.00)}$/50.2$_{\scriptscriptstyle(\!\sim\!0.62)}$ & \textcolor{GapHighRed}{52.2$_{\scriptscriptstyle(\!\sim\!1.00)}$/48.0$_{\scriptscriptstyle(\!\sim\!1.00)}$} & \textcolor{GapLowBlue}{51.0$_{\scriptscriptstyle(\!\sim\!1.00)}$/50.5$_{\scriptscriptstyle(\!\sim\!0.33)}$} & 51.3$_{\scriptscriptstyle(\!\sim\!1.00)}$/48.4$_{\scriptscriptstyle(\!\sim\!1.00)}$ \\
Qwen3-VL-8B & 51.4/51.6 & 52.7$_{\scriptscriptstyle(\!\sim\!1.00)}$/50.4$_{\scriptscriptstyle(\!\sim\!0.93)}$ & 51.6$_{\scriptscriptstyle(\!\sim\!1.00)}$/51.0$_{\scriptscriptstyle(\!\sim\!0.84)}$ & \textcolor{GapLowBlue}{48.0$_{\scriptscriptstyle(\!\sim\!0.37)}$/48.8$_{\scriptscriptstyle(\!\sim\!0.34)}$} & \textcolor{GapHighRed}{50.2$_{\scriptscriptstyle(\!\sim\!1.00)}$/45.5$_{\scriptscriptstyle(\!<\!0.01)}$} \\
Qwen3-VL-32B & 54.6/51.3 & 56.9$_{\scriptscriptstyle(\!\sim\!0.35)}$/56.6$_{\scriptscriptstyle(\!\sim\!0.01)}$ & 53.7$_{\scriptscriptstyle(\!\sim\!0.87)}$/52.2$_{\scriptscriptstyle(\!\sim\!0.88)}$ & \textcolor{GapLowBlue}{57.0$_{\scriptscriptstyle(\!\sim\!0.08)}$/56.7$_{\scriptscriptstyle(\!<\!0.01)}$} & \textcolor{GapHighRed}{57.8$_{\scriptscriptstyle(\!\sim\!0.04)}$/49.0$_{\scriptscriptstyle(\!\sim\!0.88)}$} \\
Molmo2-4B & 55.2/48.7 & 54.2$_{\scriptscriptstyle(\!\sim\!1.00)}$/50.2$_{\scriptscriptstyle(\!\sim\!1.00)}$ & \textcolor{GapHighRed}{56.1$_{\scriptscriptstyle(\!\sim\!1.00)}$/48.5$_{\scriptscriptstyle(\!\sim\!1.00)}$} & \textcolor{GapLowBlue}{51.7$_{\scriptscriptstyle(\!\sim\!0.10)}$/51.7$_{\scriptscriptstyle(\!\sim\!0.99)}$} & 51.8$_{\scriptscriptstyle(\!\sim\!0.16)}$/48.2$_{\scriptscriptstyle(\!\sim\!1.00)}$ \\
Molmo2-8B & 55.2/47.4 & \textcolor{GapHighRed}{61.7$_{\scriptscriptstyle(\!<\!0.01)}$/52.1$_{\scriptscriptstyle(\!\sim\!0.02)}$} & 55.4$_{\scriptscriptstyle(\!\sim\!1.00)}$/48.8$_{\scriptscriptstyle(\!\sim\!0.51)}$ & \textcolor{GapLowBlue}{58.7$_{\scriptscriptstyle(\!\sim\!0.03)}$/57.1$_{\scriptscriptstyle(\!<\!0.01)}$} & 57.5$_{\scriptscriptstyle(\!\sim\!0.16)}$/54.0$_{\scriptscriptstyle(\!<\!0.01)}$ \\
\midrule
\textbf{Average} & 57.3/52.9 & 61.8/56.9 & 58.4/52.9 & 57.5/57.9 & 58.3/53.5 \\
\bottomrule
\end{tabular}%
}
\end{table*}

\begin{table*}[t]
\centering
\caption{VLM-Fix direct results on Dots and Boxes (same format as Table~\ref{tab:vlm_fix_direct_tictactoe}).}
\label{tab:vlm_fix_direct_dots_boxes}
\resizebox{\textwidth}{!}{%
\begin{tabular}{lccccc}
\toprule
Model & Base & Glyph & Checkerboard & Alias & SemAlias \\
 & Std/Inv (Acc) & Std/Inv (Acc) & Std/Inv (Acc) & Std/Inv (Acc) & Std/Inv (Acc) \\
\midrule
GPT-4.1 & \textcolor{GapHighRed}{88.3/60.3} & 80.3$_{\scriptscriptstyle(\!\sim\!0.02)}$/60.7$_{\scriptscriptstyle(\!\sim\!1.00)}$ & 90.7$_{\scriptscriptstyle(\!\sim\!1.00)}$/71.3$_{\scriptscriptstyle(\!<\!0.01)}$ & \textcolor{GapLowBlue}{89.7$_{\scriptscriptstyle(\!\sim\!1.00)}$/81.0$_{\scriptscriptstyle(\!<\!0.01)}$} & 85.7$_{\scriptscriptstyle(\!\sim\!0.91)}$/72.0$_{\scriptscriptstyle(\!<\!0.01)}$ \\
GPT-5.2 & 93.7/75.7 & 91.7$_{\scriptscriptstyle(\!\sim\!1.00)}$/74.3$_{\scriptscriptstyle(\!\sim\!1.00)}$ & \textcolor{GapLowBlue}{90.7$_{\scriptscriptstyle(\!\sim\!1.00)}$/74.3$_{\scriptscriptstyle(\!\sim\!1.00)}$} & 94.0$_{\scriptscriptstyle(\!\sim\!1.00)}$/76.3$_{\scriptscriptstyle(\!\sim\!1.00)}$ & \textcolor{GapHighRed}{92.0$_{\scriptscriptstyle(\!\sim\!1.00)}$/73.0$_{\scriptscriptstyle(\!\sim\!1.00)}$} \\
Sonnet-4.0 & \textcolor{GapHighRed}{95.0/92.3} & 97.3$_{\scriptscriptstyle(\!\sim\!1.00)}$/96.0$_{\scriptscriptstyle(\!\sim\!0.21)}$ & 98.3$_{\scriptscriptstyle(\!\sim\!0.10)}$/95.7$_{\scriptscriptstyle(\!\sim\!0.30)}$ & \textcolor{GapLowBlue}{94.3$_{\scriptscriptstyle(\!\sim\!1.00)}$/96.0$_{\scriptscriptstyle(\!\sim\!0.21)}$} & 96.0$_{\scriptscriptstyle(\!\sim\!1.00)}$/95.0$_{\scriptscriptstyle(\!\sim\!0.30)}$ \\
Sonnet-4.5 & \textcolor{GapLowBlue}{78.3/77.0} & 97.0$_{\scriptscriptstyle(\!<\!0.01)}$/94.7$_{\scriptscriptstyle(\!<\!0.01)}$ & \textcolor{GapHighRed}{83.0$_{\scriptscriptstyle(\!\sim\!0.78)}$/67.0$_{\scriptscriptstyle(\!<\!0.01)}$} & 80.0$_{\scriptscriptstyle(\!\sim\!1.00)}$/64.0$_{\scriptscriptstyle(\!<\!0.01)}$ & 77.0$_{\scriptscriptstyle(\!\sim\!1.00)}$/68.0$_{\scriptscriptstyle(\!<\!0.01)}$ \\
Qwen2.5-VL-3B & 51.7/44.2 & 51.7$_{\scriptscriptstyle(\!\sim\!1.00)}$/47.8$_{\scriptscriptstyle(\!\sim\!0.37)}$ & \textcolor{GapHighRed}{57.2$_{\scriptscriptstyle(\!<\!0.01)}$/41.5$_{\scriptscriptstyle(\!\sim\!0.37)}$} & \textcolor{GapLowBlue}{51.2$_{\scriptscriptstyle(\!\sim\!1.00)}$/51.4$_{\scriptscriptstyle(\!<\!0.01)}$} & 51.8$_{\scriptscriptstyle(\!\sim\!1.00)}$/46.0$_{\scriptscriptstyle(\!\sim\!0.58)}$ \\
Qwen2.5-VL-7B & \textcolor{GapHighRed}{65.8/32.6} & 64.2$_{\scriptscriptstyle(\!\sim\!0.52)}$/41.5$_{\scriptscriptstyle(\!<\!0.01)}$ & 63.1$_{\scriptscriptstyle(\!<\!0.01)}$/38.1$_{\scriptscriptstyle(\!<\!0.01)}$ & \textcolor{GapLowBlue}{54.9$_{\scriptscriptstyle(\!<\!0.01)}$/52.4$_{\scriptscriptstyle(\!<\!0.01)}$} & 53.2$_{\scriptscriptstyle(\!<\!0.01)}$/38.4$_{\scriptscriptstyle(\!<\!0.01)}$ \\
InternVL3.5-4B & \textcolor{GapHighRed}{70.1/30.5} & \textcolor{GapLowBlue}{58.8$_{\scriptscriptstyle(\!<\!0.01)}$/43.3$_{\scriptscriptstyle(\!<\!0.01)}$} & 69.5$_{\scriptscriptstyle(\!\sim\!1.00)}$/39.8$_{\scriptscriptstyle(\!<\!0.01)}$ & 71.8$_{\scriptscriptstyle(\!\sim\!0.72)}$/50.4$_{\scriptscriptstyle(\!<\!0.01)}$ & 70.4$_{\scriptscriptstyle(\!\sim\!1.00)}$/32.4$_{\scriptscriptstyle(\!\sim\!0.25)}$ \\
InternVL3.5-8B & 72.2/55.6 & \textcolor{GapLowBlue}{56.9$_{\scriptscriptstyle(\!<\!0.01)}$/50.6$_{\scriptscriptstyle(\!\sim\!0.03)}$} & \textcolor{GapHighRed}{75.9$_{\scriptscriptstyle(\!<\!0.01)}$/56.4$_{\scriptscriptstyle(\!\sim\!0.79)}$} & 70.2$_{\scriptscriptstyle(\!\sim\!0.14)}$/53.1$_{\scriptscriptstyle(\!\sim\!0.38)}$ & 63.3$_{\scriptscriptstyle(\!<\!0.01)}$/50.3$_{\scriptscriptstyle(\!<\!0.01)}$ \\
InternVL3.5-14B & 53.2/42.9 & 59.1$_{\scriptscriptstyle(\!<\!0.01)}$/48.4$_{\scriptscriptstyle(\!<\!0.01)}$ & \textcolor{GapHighRed}{56.6$_{\scriptscriptstyle(\!<\!0.01)}$/42.3$_{\scriptscriptstyle(\!\sim\!0.95)}$} & \textcolor{GapLowBlue}{54.8$_{\scriptscriptstyle(\!\sim\!0.02)}$/69.7$_{\scriptscriptstyle(\!<\!0.01)}$} & 53.9$_{\scriptscriptstyle(\!\sim\!0.42)}$/49.8$_{\scriptscriptstyle(\!<\!0.01)}$ \\
Qwen3-VL-4B & \textcolor{GapHighRed}{65.0/38.8} & \textcolor{GapLowBlue}{53.5$_{\scriptscriptstyle(\!<\!0.01)}$/47.2$_{\scriptscriptstyle(\!<\!0.01)}$} & 60.2$_{\scriptscriptstyle(\!<\!0.01)}$/44.0$_{\scriptscriptstyle(\!<\!0.01)}$ & 59.5$_{\scriptscriptstyle(\!<\!0.01)}$/48.5$_{\scriptscriptstyle(\!<\!0.01)}$ & 57.5$_{\scriptscriptstyle(\!<\!0.01)}$/39.2$_{\scriptscriptstyle(\!\sim\!1.00)}$ \\
Qwen3-VL-8B & \textcolor{GapHighRed}{76.2/36.7} & 68.2$_{\scriptscriptstyle(\!<\!0.01)}$/36.8$_{\scriptscriptstyle(\!\sim\!1.00)}$ & 74.9$_{\scriptscriptstyle(\!\sim\!0.86)}$/39.0$_{\scriptscriptstyle(\!\sim\!0.02)}$ & \textcolor{GapLowBlue}{75.8$_{\scriptscriptstyle(\!\sim\!1.00)}$/58.7$_{\scriptscriptstyle(\!<\!0.01)}$} & 73.2$_{\scriptscriptstyle(\!\sim\!0.20)}$/39.3$_{\scriptscriptstyle(\!<\!0.01)}$ \\
Qwen3-VL-32B & 73.7/55.9 & 70.7$_{\scriptscriptstyle(\!\sim\!0.04)}$/53.5$_{\scriptscriptstyle(\!\sim\!0.38)}$ & \textcolor{GapHighRed}{76.2$_{\scriptscriptstyle(\!<\!0.01)}$/55.9$_{\scriptscriptstyle(\!\sim\!1.00)}$} & \textcolor{GapLowBlue}{72.6$_{\scriptscriptstyle(\!\sim\!0.07)}$/76.4$_{\scriptscriptstyle(\!<\!0.01)}$} & 71.2$_{\scriptscriptstyle(\!<\!0.01)}$/57.6$_{\scriptscriptstyle(\!\sim\!0.94)}$ \\
Molmo2-4B & \textcolor{GapHighRed}{71.7/22.7} & 62.5$_{\scriptscriptstyle(\!<\!0.01)}$/32.6$_{\scriptscriptstyle(\!<\!0.01)}$ & 72.8$_{\scriptscriptstyle(\!\sim\!0.72)}$/25.4$_{\scriptscriptstyle(\!\sim\!0.03)}$ & \textcolor{GapLowBlue}{49.3$_{\scriptscriptstyle(\!<\!0.01)}$/59.3$_{\scriptscriptstyle(\!<\!0.01)}$} & 64.0$_{\scriptscriptstyle(\!<\!0.01)}$/37.2$_{\scriptscriptstyle(\!<\!0.01)}$ \\
Molmo2-8B & \textcolor{GapHighRed}{77.6/34.5} & 71.8$_{\scriptscriptstyle(\!<\!0.01)}$/41.5$_{\scriptscriptstyle(\!<\!0.01)}$ & 73.7$_{\scriptscriptstyle(\!<\!0.01)}$/41.3$_{\scriptscriptstyle(\!<\!0.01)}$ & \textcolor{GapLowBlue}{74.1$_{\scriptscriptstyle(\!<\!0.01)}$/57.2$_{\scriptscriptstyle(\!<\!0.01)}$} & 75.0$_{\scriptscriptstyle(\!\sim\!0.01)}$/42.9$_{\scriptscriptstyle(\!<\!0.01)}$ \\
\midrule
\textbf{Average} & 73.8/50.0 & 70.3/54.9 & 74.5/52.3 & 70.9/63.9 & 70.3/52.9 \\
\bottomrule
\end{tabular}%
}
\end{table*}

\subsection{Results with CoT Prompting}

Tables~\ref{tab:vlm_fix_cot_tictactoe}, \ref{tab:vlm_fix_cot_reversi}, \ref{tab:vlm_fix_cot_connect4}, and \ref{tab:vlm_fix_cot_dots_boxes} report the CoT results with the same intervention columns used in the direct tables. API models are included in the Base, Alias, and SemAlias CoT conditions; Glyph and Checkerboard CoT entries are only available for the open-weight models. Across games, Alias remains the most reliable way to improve inverse-rule accuracy, while Base and SemAlias retain larger rule gaps on several tasks.

\begin{table*}[t]
\centering
\caption{VLM-Fix CoT results on Tic-Tac-Toe. API rows are included for the Base, Alias, and SemAlias CoT conditions; Glyph and Checkerboard CoT entries are unavailable for API models. Cells report \textbf{standard/inverse accuracy (\%)}; subscripts in non-baseline columns give Holm-adjusted paired McNemar $p$-values versus Base for the same rule. Within each row, \textcolor{GapHighRed}{red} and \textcolor{GapLowBlue}{blue} mark the largest and smallest $(\mathrm{Standard}-\mathrm{Inverse})$ gaps.}
\label{tab:vlm_fix_cot_tictactoe}
\resizebox{\textwidth}{!}{%
\begin{tabular}{lccccc}
\toprule
Model & Base & Glyph & Checkerboard & Alias & SemAlias \\
 & Std/Inv (Acc) & Std/Inv (Acc) & Std/Inv (Acc) & Std/Inv (Acc) & Std/Inv (Acc) \\
\midrule
GPT-4.1 & \textcolor{GapHighRed}{97.0/94.0} & -- & -- & 95.7$_{\scriptscriptstyle(\!\sim\!0.69)}$/96.7$_{\scriptscriptstyle(\!\sim\!0.37)}$ & \textcolor{GapLowBlue}{95.0$_{\scriptscriptstyle(\!\sim\!0.58)}$/97.0$_{\scriptscriptstyle(\!\sim\!0.37)}$} \\
GPT-5.2 & \textcolor{GapHighRed}{100.0/99.7} & -- & -- & \textcolor{GapLowBlue}{100.0$_{\scriptscriptstyle(\!\sim\!1.00)}$/100.0$_{\scriptscriptstyle(\!\sim\!1.00)}$} & 100.0$_{\scriptscriptstyle(\!\sim\!1.00)}$/100.0$_{\scriptscriptstyle(\!\sim\!1.00)}$ \\
Sonnet-4.0 & \textcolor{GapLowBlue}{95.0/96.7} & -- & -- & 97.7$_{\scriptscriptstyle(\!\sim\!0.03)}$/97.0$_{\scriptscriptstyle(\!\sim\!1.00)}$ & \textcolor{GapHighRed}{97.7$_{\scriptscriptstyle(\!\sim\!0.08)}$/96.7$_{\scriptscriptstyle(\!\sim\!1.00)}$} \\
Sonnet-4.5 & 96.7/95.3 & -- & -- & \textcolor{GapLowBlue}{95.3$_{\scriptscriptstyle(\!\sim\!0.58)}$/96.7$_{\scriptscriptstyle(\!\sim\!1.00)}$} & \textcolor{GapHighRed}{99.0$_{\scriptscriptstyle(\!\sim\!0.26)}$/96.0$_{\scriptscriptstyle(\!\sim\!1.00)}$} \\
Qwen2.5-VL-3B & 59.8/47.6 & 57.2$_{\scriptscriptstyle(\!\sim\!1.00)}$/49.7$_{\scriptscriptstyle(\!\sim\!1.00)}$ & \textcolor{GapHighRed}{59.2$_{\scriptscriptstyle(\!\sim\!1.00)}$/46.8$_{\scriptscriptstyle(\!\sim\!1.00)}$} & \textcolor{GapLowBlue}{59.3$_{\scriptscriptstyle(\!\sim\!1.00)}$/57.5$_{\scriptscriptstyle(\!<\!0.01)}$} & 58.8$_{\scriptscriptstyle(\!\sim\!1.00)}$/54.3$_{\scriptscriptstyle(\!<\!0.01)}$ \\
Qwen2.5-VL-7B & \textcolor{GapHighRed}{71.3/32.8} & 70.0$_{\scriptscriptstyle(\!\sim\!1.00)}$/45.2$_{\scriptscriptstyle(\!<\!0.01)}$ & 70.9$_{\scriptscriptstyle(\!\sim\!1.00)}$/33.6$_{\scriptscriptstyle(\!\sim\!1.00)}$ & \textcolor{GapLowBlue}{69.0$_{\scriptscriptstyle(\!\sim\!0.83)}$/69.3$_{\scriptscriptstyle(\!<\!0.01)}$} & 68.8$_{\scriptscriptstyle(\!\sim\!0.80)}$/64.6$_{\scriptscriptstyle(\!<\!0.01)}$ \\
InternVL3.5-4B & \textcolor{GapHighRed}{85.0/58.4} & 88.4$_{\scriptscriptstyle(\!\sim\!0.02)}$/83.6$_{\scriptscriptstyle(\!<\!0.01)}$ & 83.1$_{\scriptscriptstyle(\!\sim\!0.30)}$/62.6$_{\scriptscriptstyle(\!\sim\!0.02)}$ & \textcolor{GapLowBlue}{81.2$_{\scriptscriptstyle(\!<\!0.01)}$/82.1$_{\scriptscriptstyle(\!<\!0.01)}$} & 78.7$_{\scriptscriptstyle(\!<\!0.01)}$/76.7$_{\scriptscriptstyle(\!<\!0.01)}$ \\
InternVL3.5-8B & 90.9/85.2 & 93.0$_{\scriptscriptstyle(\!\sim\!0.24)}$/90.0$_{\scriptscriptstyle(\!<\!0.01)}$ & \textcolor{GapHighRed}{89.8$_{\scriptscriptstyle(\!\sim\!0.69)}$/82.3$_{\scriptscriptstyle(\!\sim\!0.12)}$} & \textcolor{GapLowBlue}{89.2$_{\scriptscriptstyle(\!\sim\!0.29)}$/89.7$_{\scriptscriptstyle(\!<\!0.01)}$} & 85.8$_{\scriptscriptstyle(\!<\!0.01)}$/83.4$_{\scriptscriptstyle(\!\sim\!0.40)}$ \\
InternVL3.5-14B & 86.7/82.9 & 87.4$_{\scriptscriptstyle(\!\sim\!1.00)}$/83.8$_{\scriptscriptstyle(\!\sim\!1.00)}$ & 84.7$_{\scriptscriptstyle(\!\sim\!0.39)}$/84.3$_{\scriptscriptstyle(\!\sim\!1.00)}$ & \textcolor{GapLowBlue}{79.8$_{\scriptscriptstyle(\!<\!0.01)}$/83.3$_{\scriptscriptstyle(\!\sim\!1.00)}$} & \textcolor{GapHighRed}{77.3$_{\scriptscriptstyle(\!<\!0.01)}$/69.0$_{\scriptscriptstyle(\!<\!0.01)}$} \\
Qwen3-VL-4B & \textcolor{GapLowBlue}{78.9/79.7} & 73.4$_{\scriptscriptstyle(\!<\!0.01)}$/73.8$_{\scriptscriptstyle(\!<\!0.01)}$ & \textcolor{GapHighRed}{75.8$_{\scriptscriptstyle(\!\sim\!0.21)}$/74.8$_{\scriptscriptstyle(\!\sim\!0.01)}$} & 79.7$_{\scriptscriptstyle(\!\sim\!1.00)}$/80.2$_{\scriptscriptstyle(\!\sim\!1.00)}$ & 80.9$_{\scriptscriptstyle(\!\sim\!0.59)}$/81.5$_{\scriptscriptstyle(\!\sim\!0.79)}$ \\
Qwen3-VL-8B & 83.2/82.2 & \textcolor{GapHighRed}{76.3$_{\scriptscriptstyle(\!<\!0.01)}$/73.2$_{\scriptscriptstyle(\!<\!0.01)}$} & 87.0$_{\scriptscriptstyle(\!\sim\!0.02)}$/84.8$_{\scriptscriptstyle(\!\sim\!0.37)}$ & \textcolor{GapLowBlue}{80.8$_{\scriptscriptstyle(\!\sim\!0.18)}$/82.7$_{\scriptscriptstyle(\!\sim\!1.00)}$} & 84.9$_{\scriptscriptstyle(\!\sim\!0.36)}$/84.1$_{\scriptscriptstyle(\!\sim\!0.69)}$ \\
Qwen3-VL-32B & \textcolor{GapLowBlue}{99.4/99.8} & \textcolor{GapHighRed}{98.3$_{\scriptscriptstyle(\!\sim\!0.12)}$/98.0$_{\scriptscriptstyle(\!<\!0.01)}$} & 100.0$_{\scriptscriptstyle(\!\sim\!0.12)}$/100.0$_{\scriptscriptstyle(\!\sim\!0.88)}$ & 99.6$_{\scriptscriptstyle(\!\sim\!1.00)}$/99.4$_{\scriptscriptstyle(\!\sim\!0.88)}$ & 99.4$_{\scriptscriptstyle(\!\sim\!1.00)}$/99.3$_{\scriptscriptstyle(\!\sim\!0.75)}$ \\
Molmo2-4B & 75.1/52.8 & 72.1$_{\scriptscriptstyle(\!\sim\!0.29)}$/59.5$_{\scriptscriptstyle(\!<\!0.01)}$ & \textcolor{GapHighRed}{80.1$_{\scriptscriptstyle(\!<\!0.01)}$/55.2$_{\scriptscriptstyle(\!\sim\!0.33)}$} & \textcolor{GapLowBlue}{77.2$_{\scriptscriptstyle(\!\sim\!0.29)}$/78.8$_{\scriptscriptstyle(\!<\!0.01)}$} & 78.5$_{\scriptscriptstyle(\!\sim\!0.14)}$/71.8$_{\scriptscriptstyle(\!<\!0.01)}$ \\
Molmo2-8B & \textcolor{GapHighRed}{79.8/63.2} & 75.6$_{\scriptscriptstyle(\!\sim\!0.04)}$/68.2$_{\scriptscriptstyle(\!\sim\!0.02)}$ & 76.8$_{\scriptscriptstyle(\!\sim\!0.10)}$/64.4$_{\scriptscriptstyle(\!\sim\!1.00)}$ & \textcolor{GapLowBlue}{73.8$_{\scriptscriptstyle(\!<\!0.01)}$/73.2$_{\scriptscriptstyle(\!<\!0.01)}$} & 71.3$_{\scriptscriptstyle(\!<\!0.01)}$/56.6$_{\scriptscriptstyle(\!<\!0.01)}$ \\
\midrule
\textbf{Average} & 85.6/76.4 & 79.2/72.5 & 80.7/68.9 & 84.2/84.8 & 84.0/80.8 \\
\bottomrule
\end{tabular}%
}
\end{table*}

\begin{table*}[t]
\centering
\caption{VLM-Fix CoT results on Reversi.}
\label{tab:vlm_fix_cot_reversi}
\resizebox{\textwidth}{!}{%
\begin{tabular}{lccccc}
\toprule
Model & Base & Glyph & Checkerboard & Alias & SemAlias \\
 & Std/Inv (Acc) & Std/Inv (Acc) & Std/Inv (Acc) & Std/Inv (Acc) & Std/Inv (Acc) \\
\midrule
GPT-4.1 & \textcolor{GapHighRed}{98.0/96.0} & -- & -- & \textcolor{GapLowBlue}{96.7$_{\scriptscriptstyle(\!\sim\!1.00)}$/96.7$_{\scriptscriptstyle(\!\sim\!1.00)}$} & 97.7$_{\scriptscriptstyle(\!\sim\!1.00)}$/97.3$_{\scriptscriptstyle(\!\sim\!1.00)}$ \\
GPT-5.2 & \textcolor{GapLowBlue}{100.0/100.0} & -- & -- & 100.0$_{\scriptscriptstyle(\!\sim\!1.00)}$/99.7$_{\scriptscriptstyle(\!\sim\!1.00)}$ & \textcolor{GapHighRed}{100.0$_{\scriptscriptstyle(\!\sim\!1.00)}$/99.0$_{\scriptscriptstyle(\!\sim\!1.00)}$} \\
Sonnet-4.0 & \textcolor{GapLowBlue}{99.3/99.3} & -- & -- & 99.7$_{\scriptscriptstyle(\!\sim\!1.00)}$/99.7$_{\scriptscriptstyle(\!\sim\!1.00)}$ & \textcolor{GapHighRed}{100.0$_{\scriptscriptstyle(\!\sim\!1.00)}$/99.3$_{\scriptscriptstyle(\!\sim\!1.00)}$} \\
Sonnet-4.5 & 100.0/100.0 & -- & -- & 100.0$_{\scriptscriptstyle(\!\sim\!1.00)}$/100.0$_{\scriptscriptstyle(\!\sim\!1.00)}$ & 100.0$_{\scriptscriptstyle(\!\sim\!1.00)}$/100.0$_{\scriptscriptstyle(\!\sim\!1.00)}$ \\
Qwen2.5-VL-3B & \textcolor{GapHighRed}{70.7/33.8} & 67.6$_{\scriptscriptstyle(\!\sim\!0.25)}$/36.9$_{\scriptscriptstyle(\!\sim\!0.32)}$ & 64.1$_{\scriptscriptstyle(\!<\!0.01)}$/31.5$_{\scriptscriptstyle(\!\sim\!0.32)}$ & \textcolor{GapLowBlue}{72.3$_{\scriptscriptstyle(\!\sim\!0.54)}$/53.7$_{\scriptscriptstyle(\!<\!0.01)}$} & 76.8$_{\scriptscriptstyle(\!<\!0.01)}$/41.6$_{\scriptscriptstyle(\!<\!0.01)}$ \\
Qwen2.5-VL-7B & \textcolor{GapHighRed}{85.8/20.7} & 87.8$_{\scriptscriptstyle(\!\sim\!0.36)}$/29.6$_{\scriptscriptstyle(\!<\!0.01)}$ & 76.2$_{\scriptscriptstyle(\!<\!0.01)}$/29.9$_{\scriptscriptstyle(\!<\!0.01)}$ & \textcolor{GapLowBlue}{86.0$_{\scriptscriptstyle(\!\sim\!1.00)}$/84.4$_{\scriptscriptstyle(\!<\!0.01)}$} & 85.6$_{\scriptscriptstyle(\!\sim\!1.00)}$/29.4$_{\scriptscriptstyle(\!<\!0.01)}$ \\
InternVL3.5-4B & 94.0/72.8 & 93.9$_{\scriptscriptstyle(\!\sim\!1.00)}$/73.5$_{\scriptscriptstyle(\!\sim\!1.00)}$ & 95.2$_{\scriptscriptstyle(\!\sim\!0.75)}$/79.6$_{\scriptscriptstyle(\!<\!0.01)}$ & \textcolor{GapLowBlue}{89.5$_{\scriptscriptstyle(\!<\!0.01)}$/93.8$_{\scriptscriptstyle(\!<\!0.01)}$} & \textcolor{GapHighRed}{94.2$_{\scriptscriptstyle(\!\sim\!1.00)}$/68.0$_{\scriptscriptstyle(\!<\!0.01)}$} \\
InternVL3.5-8B & \textcolor{GapHighRed}{94.1/71.5} & 95.7$_{\scriptscriptstyle(\!\sim\!0.44)}$/85.7$_{\scriptscriptstyle(\!<\!0.01)}$ & 93.1$_{\scriptscriptstyle(\!\sim\!0.61)}$/71.8$_{\scriptscriptstyle(\!\sim\!1.00)}$ & \textcolor{GapLowBlue}{89.7$_{\scriptscriptstyle(\!<\!0.01)}$/92.2$_{\scriptscriptstyle(\!<\!0.01)}$} & 92.8$_{\scriptscriptstyle(\!\sim\!0.53)}$/82.3$_{\scriptscriptstyle(\!<\!0.01)}$ \\
InternVL3.5-14B & 76.2/78.5 & \textcolor{GapHighRed}{85.1$_{\scriptscriptstyle(\!<\!0.01)}$/77.9$_{\scriptscriptstyle(\!\sim\!1.00)}$} & 73.5$_{\scriptscriptstyle(\!\sim\!0.07)}$/76.5$_{\scriptscriptstyle(\!\sim\!1.00)}$ & \textcolor{GapLowBlue}{81.1$_{\scriptscriptstyle(\!<\!0.01)}$/88.5$_{\scriptscriptstyle(\!<\!0.01)}$} & 81.8$_{\scriptscriptstyle(\!<\!0.01)}$/78.6$_{\scriptscriptstyle(\!\sim\!1.00)}$ \\
Qwen3-VL-4B & 93.5/85.7 & \textcolor{GapHighRed}{95.2$_{\scriptscriptstyle(\!\sim\!0.62)}$/78.8$_{\scriptscriptstyle(\!<\!0.01)}$} & 94.4$_{\scriptscriptstyle(\!\sim\!1.00)}$/87.3$_{\scriptscriptstyle(\!\sim\!0.35)}$ & \textcolor{GapLowBlue}{93.8$_{\scriptscriptstyle(\!\sim\!1.00)}$/93.0$_{\scriptscriptstyle(\!<\!0.01)}$} & 93.1$_{\scriptscriptstyle(\!\sim\!1.00)}$/90.7$_{\scriptscriptstyle(\!<\!0.01)}$ \\
Qwen3-VL-8B & 98.0/92.3 & \textcolor{GapHighRed}{98.3$_{\scriptscriptstyle(\!\sim\!1.00)}$/90.2$_{\scriptscriptstyle(\!\sim\!0.07)}$} & 97.8$_{\scriptscriptstyle(\!\sim\!1.00)}$/97.7$_{\scriptscriptstyle(\!<\!0.01)}$ & \textcolor{GapLowBlue}{97.7$_{\scriptscriptstyle(\!\sim\!1.00)}$/97.8$_{\scriptscriptstyle(\!<\!0.01)}$} & 98.6$_{\scriptscriptstyle(\!\sim\!1.00)}$/96.5$_{\scriptscriptstyle(\!<\!0.01)}$ \\
Qwen3-VL-32B & 100.0/99.8 & \textcolor{GapLowBlue}{100.0$_{\scriptscriptstyle(\!\sim\!1.00)}$/99.9$_{\scriptscriptstyle(\!\sim\!1.00)}$} & 100.0$_{\scriptscriptstyle(\!\sim\!1.00)}$/99.8$_{\scriptscriptstyle(\!\sim\!1.00)}$ & 99.4$_{\scriptscriptstyle(\!\sim\!0.12)}$/99.1$_{\scriptscriptstyle(\!\sim\!0.07)}$ & \textcolor{GapHighRed}{99.8$_{\scriptscriptstyle(\!\sim\!1.00)}$/97.2$_{\scriptscriptstyle(\!<\!0.01)}$} \\
Molmo2-4B & 85.5/45.2 & \textcolor{GapHighRed}{90.8$_{\scriptscriptstyle(\!<\!0.01)}$/21.8$_{\scriptscriptstyle(\!<\!0.01)}$} & 85.2$_{\scriptscriptstyle(\!\sim\!1.00)}$/37.6$_{\scriptscriptstyle(\!<\!0.01)}$ & \textcolor{GapLowBlue}{80.9$_{\scriptscriptstyle(\!<\!0.01)}$/79.3$_{\scriptscriptstyle(\!<\!0.01)}$} & 79.7$_{\scriptscriptstyle(\!<\!0.01)}$/50.3$_{\scriptscriptstyle(\!\sim\!0.02)}$ \\
Molmo2-8B & 92.0/42.2 & \textcolor{GapHighRed}{93.6$_{\scriptscriptstyle(\!\sim\!0.81)}$/16.3$_{\scriptscriptstyle(\!<\!0.01)}$} & 90.4$_{\scriptscriptstyle(\!\sim\!0.81)}$/36.3$_{\scriptscriptstyle(\!<\!0.01)}$ & \textcolor{GapLowBlue}{91.8$_{\scriptscriptstyle(\!\sim\!1.00)}$/90.3$_{\scriptscriptstyle(\!<\!0.01)}$} & 92.4$_{\scriptscriptstyle(\!\sim\!1.00)}$/47.6$_{\scriptscriptstyle(\!<\!0.01)}$ \\
\midrule
\textbf{Average} & 91.9/74.1 & 90.8/61.1 & 87.0/64.8 & 91.3/90.6 & 92.3/77.0 \\
\bottomrule
\end{tabular}%
}
\end{table*}

\begin{table*}[t]
\centering
\caption{VLM-Fix CoT results on Connect Four.}
\label{tab:vlm_fix_cot_connect4}
\resizebox{\textwidth}{!}{%
\begin{tabular}{lccccc}
\toprule
Model & Base & Glyph & Checkerboard & Alias & SemAlias \\
 & Std/Inv (Acc) & Std/Inv (Acc) & Std/Inv (Acc) & Std/Inv (Acc) & Std/Inv (Acc) \\
\midrule
GPT-4.1 & 98.7/99.0 & -- & -- & \textcolor{GapLowBlue}{98.3$_{\scriptscriptstyle(\!\sim\!1.00)}$/99.0$_{\scriptscriptstyle(\!\sim\!1.00)}$} & \textcolor{GapHighRed}{99.3$_{\scriptscriptstyle(\!\sim\!1.00)}$/94.7$_{\scriptscriptstyle(\!\sim\!0.02)}$} \\
GPT-5.2 & \textcolor{GapLowBlue}{99.0/99.0} & -- & -- & \textcolor{GapHighRed}{100.0$_{\scriptscriptstyle(\!\sim\!1.00)}$/99.0$_{\scriptscriptstyle(\!\sim\!1.00)}$} & 100.0$_{\scriptscriptstyle(\!\sim\!1.00)}$/99.3$_{\scriptscriptstyle(\!\sim\!1.00)}$ \\
Sonnet-4.0 & \textcolor{GapLowBlue}{99.7/99.7} & -- & -- & 100.0$_{\scriptscriptstyle(\!\sim\!1.00)}$/100.0$_{\scriptscriptstyle(\!\sim\!1.00)}$ & \textcolor{GapHighRed}{100.0$_{\scriptscriptstyle(\!\sim\!1.00)}$/99.7$_{\scriptscriptstyle(\!\sim\!1.00)}$} \\
Sonnet-4.5 & \textcolor{GapHighRed}{100.0/98.7} & -- & -- & \textcolor{GapLowBlue}{100.0$_{\scriptscriptstyle(\!\sim\!1.00)}$/100.0$_{\scriptscriptstyle(\!\sim\!0.50)}$} & 100.0$_{\scriptscriptstyle(\!\sim\!1.00)}$/99.0$_{\scriptscriptstyle(\!\sim\!1.00)}$ \\
Qwen2.5-VL-3B & 43.5/42.0 & \textcolor{GapHighRed}{55.9$_{\scriptscriptstyle(\!<\!0.01)}$/46.0$_{\scriptscriptstyle(\!\sim\!0.10)}$} & 40.0$_{\scriptscriptstyle(\!\sim\!0.13)}$/33.5$_{\scriptscriptstyle(\!<\!0.01)}$ & \textcolor{GapLowBlue}{48.2$_{\scriptscriptstyle(\!\sim\!0.06)}$/49.2$_{\scriptscriptstyle(\!<\!0.01)}$} & 49.8$_{\scriptscriptstyle(\!<\!0.01)}$/47.8$_{\scriptscriptstyle(\!\sim\!0.01)}$ \\
Qwen2.5-VL-7B & \textcolor{GapHighRed}{69.0/36.6} & 67.9$_{\scriptscriptstyle(\!\sim\!1.00)}$/44.8$_{\scriptscriptstyle(\!<\!0.01)}$ & 69.7$_{\scriptscriptstyle(\!\sim\!1.00)}$/37.4$_{\scriptscriptstyle(\!\sim\!1.00)}$ & \textcolor{GapLowBlue}{67.3$_{\scriptscriptstyle(\!\sim\!1.00)}$/67.2$_{\scriptscriptstyle(\!<\!0.01)}$} & 63.2$_{\scriptscriptstyle(\!<\!0.01)}$/60.3$_{\scriptscriptstyle(\!<\!0.01)}$ \\
InternVL3.5-4B & 77.8/66.2 & 83.9$_{\scriptscriptstyle(\!<\!0.01)}$/83.0$_{\scriptscriptstyle(\!<\!0.01)}$ & \textcolor{GapHighRed}{77.2$_{\scriptscriptstyle(\!\sim\!1.00)}$/64.7$_{\scriptscriptstyle(\!\sim\!0.64)}$} & \textcolor{GapLowBlue}{79.8$_{\scriptscriptstyle(\!\sim\!1.00)}$/80.2$_{\scriptscriptstyle(\!<\!0.01)}$} & 79.4$_{\scriptscriptstyle(\!\sim\!1.00)}$/73.5$_{\scriptscriptstyle(\!<\!0.01)}$ \\
InternVL3.5-8B & \textcolor{GapLowBlue}{88.2/87.2} & 89.7$_{\scriptscriptstyle(\!\sim\!1.00)}$/86.8$_{\scriptscriptstyle(\!\sim\!1.00)}$ & 89.2$_{\scriptscriptstyle(\!\sim\!1.00)}$/87.2$_{\scriptscriptstyle(\!\sim\!1.00)}$ & 88.5$_{\scriptscriptstyle(\!\sim\!1.00)}$/85.2$_{\scriptscriptstyle(\!\sim\!0.97)}$ & \textcolor{GapHighRed}{88.4$_{\scriptscriptstyle(\!\sim\!1.00)}$/81.1$_{\scriptscriptstyle(\!<\!0.01)}$} \\
InternVL3.5-14B & 82.9/77.4 & \textcolor{GapLowBlue}{84.2$_{\scriptscriptstyle(\!\sim\!1.00)}$/83.8$_{\scriptscriptstyle(\!<\!0.01)}$} & 80.9$_{\scriptscriptstyle(\!\sim\!0.79)}$/75.6$_{\scriptscriptstyle(\!\sim\!0.31)}$ & 84.1$_{\scriptscriptstyle(\!\sim\!1.00)}$/82.3$_{\scriptscriptstyle(\!<\!0.01)}$ & \textcolor{GapHighRed}{82.0$_{\scriptscriptstyle(\!\sim\!1.00)}$/71.2$_{\scriptscriptstyle(\!<\!0.01)}$} \\
Qwen3-VL-4B & 91.6/93.4 & 87.3$_{\scriptscriptstyle(\!<\!0.01)}$/86.5$_{\scriptscriptstyle(\!<\!0.01)}$ & \textcolor{GapLowBlue}{92.8$_{\scriptscriptstyle(\!\sim\!0.54)}$/94.9$_{\scriptscriptstyle(\!\sim\!0.74)}$} & \textcolor{GapHighRed}{93.7$_{\scriptscriptstyle(\!\sim\!0.23)}$/92.3$_{\scriptscriptstyle(\!\sim\!0.74)}$} & 94.6$_{\scriptscriptstyle(\!\sim\!0.02)}$/94.8$_{\scriptscriptstyle(\!\sim\!0.74)}$ \\
Qwen3-VL-8B & \textcolor{GapLowBlue}{86.3/88.8} & 62.7$_{\scriptscriptstyle(\!<\!0.01)}$/62.2$_{\scriptscriptstyle(\!<\!0.01)}$ & 99.1$_{\scriptscriptstyle(\!<\!0.01)}$/99.2$_{\scriptscriptstyle(\!<\!0.01)}$ & 89.2$_{\scriptscriptstyle(\!\sim\!0.03)}$/89.2$_{\scriptscriptstyle(\!\sim\!1.00)}$ & \textcolor{GapHighRed}{90.7$_{\scriptscriptstyle(\!<\!0.01)}$/88.8$_{\scriptscriptstyle(\!\sim\!1.00)}$} \\
Qwen3-VL-32B & \textcolor{GapLowBlue}{99.6/99.7} & 99.2$_{\scriptscriptstyle(\!\sim\!1.00)}$/99.2$_{\scriptscriptstyle(\!\sim\!1.00)}$ & 99.9$_{\scriptscriptstyle(\!\sim\!1.00)}$/99.9$_{\scriptscriptstyle(\!\sim\!1.00)}$ & \textcolor{GapHighRed}{99.7$_{\scriptscriptstyle(\!\sim\!1.00)}$/99.4$_{\scriptscriptstyle(\!\sim\!1.00)}$} & 99.8$_{\scriptscriptstyle(\!\sim\!1.00)}$/99.8$_{\scriptscriptstyle(\!\sim\!1.00)}$ \\
Molmo2-4B & \textcolor{GapLowBlue}{56.2/56.7} & \textcolor{GapHighRed}{68.4$_{\scriptscriptstyle(\!<\!0.01)}$/62.6$_{\scriptscriptstyle(\!\sim\!0.01)}$} & 57.5$_{\scriptscriptstyle(\!\sim\!0.95)}$/54.8$_{\scriptscriptstyle(\!\sim\!0.62)}$ & 66.3$_{\scriptscriptstyle(\!<\!0.01)}$/65.6$_{\scriptscriptstyle(\!<\!0.01)}$ & 66.8$_{\scriptscriptstyle(\!<\!0.01)}$/61.3$_{\scriptscriptstyle(\!\sim\!0.05)}$ \\
Molmo2-8B & 73.3/64.2 & 72.6$_{\scriptscriptstyle(\!\sim\!1.00)}$/66.4$_{\scriptscriptstyle(\!\sim\!1.00)}$ & 71.8$_{\scriptscriptstyle(\!\sim\!1.00)}$/64.8$_{\scriptscriptstyle(\!\sim\!1.00)}$ & \textcolor{GapLowBlue}{66.8$_{\scriptscriptstyle(\!<\!0.01)}$/65.7$_{\scriptscriptstyle(\!\sim\!1.00)}$} & \textcolor{GapHighRed}{67.8$_{\scriptscriptstyle(\!\sim\!0.01)}$/52.4$_{\scriptscriptstyle(\!<\!0.01)}$} \\
\midrule
\textbf{Average} & 83.3/79.2 & 77.2/72.1 & 77.8/71.2 & 84.4/83.9 & 84.4/80.3 \\
\bottomrule
\end{tabular}%
}
\end{table*}

\begin{table*}[t]
\centering
\caption{VLM-Fix CoT results on Dots \& Boxes.}
\label{tab:vlm_fix_cot_dots_boxes}
\resizebox{\textwidth}{!}{%
\begin{tabular}{lccccc}
\toprule
Model & Base & Glyph & Checkerboard & Alias & SemAlias \\
 & Std/Inv (Acc) & Std/Inv (Acc) & Std/Inv (Acc) & Std/Inv (Acc) & Std/Inv (Acc) \\
\midrule
GPT-4.1 & 85.7/81.7 & -- & -- & \textcolor{GapLowBlue}{88.0$_{\scriptscriptstyle(\!\sim\!0.76)}$/85.0$_{\scriptscriptstyle(\!\sim\!0.59)}$} & \textcolor{GapHighRed}{90.7$_{\scriptscriptstyle(\!\sim\!0.13)}$/86.0$_{\scriptscriptstyle(\!\sim\!0.44)}$} \\
GPT-5.2 & \textcolor{GapHighRed}{100.0/100.0} & -- & -- & \textcolor{GapLowBlue}{99.7$_{\scriptscriptstyle(\!\sim\!1.00)}$/100.0$_{\scriptscriptstyle(\!\sim\!1.00)}$} & 100.0$_{\scriptscriptstyle(\!\sim\!1.00)}$/100.0$_{\scriptscriptstyle(\!\sim\!1.00)}$ \\
Sonnet-4.0 & \textcolor{GapHighRed}{100.0/99.7} & -- & -- & \textcolor{GapLowBlue}{99.7$_{\scriptscriptstyle(\!\sim\!1.00)}$/99.7$_{\scriptscriptstyle(\!\sim\!1.00)}$} & 99.7$_{\scriptscriptstyle(\!\sim\!1.00)}$/99.7$_{\scriptscriptstyle(\!\sim\!1.00)}$ \\
Sonnet-4.5 & 100.0/100.0 & -- & -- & 100.0$_{\scriptscriptstyle(\!\sim\!1.00)}$/100.0$_{\scriptscriptstyle(\!\sim\!1.00)}$ & 100.0$_{\scriptscriptstyle(\!\sim\!1.00)}$/100.0$_{\scriptscriptstyle(\!\sim\!1.00)}$ \\
Qwen2.5-VL-3B & \textcolor{GapLowBlue}{52.1/49.0} & 52.5$_{\scriptscriptstyle(\!\sim\!1.00)}$/42.0$_{\scriptscriptstyle(\!\sim\!0.02)}$ & 50.6$_{\scriptscriptstyle(\!\sim\!1.00)}$/45.8$_{\scriptscriptstyle(\!\sim\!0.23)}$ & \textcolor{GapHighRed}{57.8$_{\scriptscriptstyle(\!\sim\!0.02)}$/45.5$_{\scriptscriptstyle(\!\sim\!0.26)}$} & 59.8$_{\scriptscriptstyle(\!<\!0.01)}$/47.8$_{\scriptscriptstyle(\!\sim\!1.00)}$ \\
Qwen2.5-VL-7B & 63.1/38.5 & \textcolor{GapHighRed}{65.2$_{\scriptscriptstyle(\!\sim\!0.56)}$/38.2$_{\scriptscriptstyle(\!\sim\!1.00)}$} & 65.8$_{\scriptscriptstyle(\!\sim\!0.45)}$/40.7$_{\scriptscriptstyle(\!\sim\!0.88)}$ & \textcolor{GapLowBlue}{69.0$_{\scriptscriptstyle(\!<\!0.01)}$/65.1$_{\scriptscriptstyle(\!<\!0.01)}$} & 68.4$_{\scriptscriptstyle(\!\sim\!0.01)}$/45.2$_{\scriptscriptstyle(\!<\!0.01)}$ \\
InternVL3.5-4B & 89.2/74.2 & 86.9$_{\scriptscriptstyle(\!\sim\!0.42)}$/67.9$_{\scriptscriptstyle(\!<\!0.01)}$ & 93.4$_{\scriptscriptstyle(\!<\!0.01)}$/84.2$_{\scriptscriptstyle(\!<\!0.01)}$ & \textcolor{GapLowBlue}{89.6$_{\scriptscriptstyle(\!\sim\!1.00)}$/86.0$_{\scriptscriptstyle(\!<\!0.01)}$} & \textcolor{GapHighRed}{90.0$_{\scriptscriptstyle(\!\sim\!1.00)}$/66.4$_{\scriptscriptstyle(\!<\!0.01)}$} \\
InternVL3.5-8B & 79.1/81.2 & \textcolor{GapLowBlue}{75.9$_{\scriptscriptstyle(\!\sim\!0.09)}$/81.0$_{\scriptscriptstyle(\!\sim\!1.00)}$} & 83.6$_{\scriptscriptstyle(\!<\!0.01)}$/80.4$_{\scriptscriptstyle(\!\sim\!1.00)}$ & 88.5$_{\scriptscriptstyle(\!<\!0.01)}$/84.8$_{\scriptscriptstyle(\!\sim\!0.09)}$ & \textcolor{GapHighRed}{91.5$_{\scriptscriptstyle(\!<\!0.01)}$/74.2$_{\scriptscriptstyle(\!<\!0.01)}$} \\
InternVL3.5-14B & \textcolor{GapHighRed}{64.9/54.6} & 60.3$_{\scriptscriptstyle(\!\sim\!0.08)}$/59.5$_{\scriptscriptstyle(\!\sim\!0.08)}$ & 67.8$_{\scriptscriptstyle(\!\sim\!0.24)}$/60.2$_{\scriptscriptstyle(\!\sim\!0.04)}$ & \textcolor{GapLowBlue}{52.6$_{\scriptscriptstyle(\!<\!0.01)}$/52.4$_{\scriptscriptstyle(\!\sim\!1.00)}$} & 57.6$_{\scriptscriptstyle(\!<\!0.01)}$/55.1$_{\scriptscriptstyle(\!\sim\!1.00)}$ \\
Qwen3-VL-4B & 59.7/58.3 & \textcolor{GapHighRed}{72.2$_{\scriptscriptstyle(\!<\!0.01)}$/68.8$_{\scriptscriptstyle(\!<\!0.01)}$} & 89.1$_{\scriptscriptstyle(\!<\!0.01)}$/89.0$_{\scriptscriptstyle(\!<\!0.01)}$ & \textcolor{GapLowBlue}{56.2$_{\scriptscriptstyle(\!\sim\!0.28)}$/61.8$_{\scriptscriptstyle(\!\sim\!0.32)}$} & 58.8$_{\scriptscriptstyle(\!\sim\!1.00)}$/61.3$_{\scriptscriptstyle(\!\sim\!0.32)}$ \\
Qwen3-VL-8B & 59.1/57.5 & 53.7$_{\scriptscriptstyle(\!\sim\!0.05)}$/56.8$_{\scriptscriptstyle(\!\sim\!1.00)}$ & \textcolor{GapHighRed}{71.0$_{\scriptscriptstyle(\!<\!0.01)}$/68.8$_{\scriptscriptstyle(\!<\!0.01)}$} & \textcolor{GapLowBlue}{57.8$_{\scriptscriptstyle(\!\sim\!1.00)}$/62.2$_{\scriptscriptstyle(\!\sim\!0.06)}$} & 63.0$_{\scriptscriptstyle(\!\sim\!0.16)}$/65.2$_{\scriptscriptstyle(\!<\!0.01)}$ \\
Qwen3-VL-32B & 93.5/93.6 & 90.8$_{\scriptscriptstyle(\!\sim\!0.05)}$/90.5$_{\scriptscriptstyle(\!\sim\!0.02)}$ & 97.9$_{\scriptscriptstyle(\!<\!0.01)}$/98.2$_{\scriptscriptstyle(\!<\!0.01)}$ & \textcolor{GapLowBlue}{93.0$_{\scriptscriptstyle(\!\sim\!1.00)}$/95.7$_{\scriptscriptstyle(\!\sim\!0.04)}$} & \textcolor{GapHighRed}{96.5$_{\scriptscriptstyle(\!<\!0.01)}$/90.4$_{\scriptscriptstyle(\!\sim\!0.02)}$} \\
Molmo2-4B & \textcolor{GapHighRed}{83.9/13.0} & 79.0$_{\scriptscriptstyle(\!<\!0.01)}$/19.0$_{\scriptscriptstyle(\!<\!0.01)}$ & 82.3$_{\scriptscriptstyle(\!\sim\!1.00)}$/18.8$_{\scriptscriptstyle(\!<\!0.01)}$ & \textcolor{GapLowBlue}{82.7$_{\scriptscriptstyle(\!\sim\!1.00)}$/66.2$_{\scriptscriptstyle(\!<\!0.01)}$} & 85.0$_{\scriptscriptstyle(\!\sim\!1.00)}$/48.8$_{\scriptscriptstyle(\!<\!0.01)}$ \\
Molmo2-8B & \textcolor{GapHighRed}{79.0/30.3} & 78.2$_{\scriptscriptstyle(\!\sim\!1.00)}$/33.1$_{\scriptscriptstyle(\!\sim\!0.23)}$ & 67.0$_{\scriptscriptstyle(\!<\!0.01)}$/41.6$_{\scriptscriptstyle(\!<\!0.01)}$ & \textcolor{GapLowBlue}{76.9$_{\scriptscriptstyle(\!\sim\!1.00)}$/77.8$_{\scriptscriptstyle(\!<\!0.01)}$} & 76.8$_{\scriptscriptstyle(\!\sim\!1.00)}$/53.8$_{\scriptscriptstyle(\!<\!0.01)}$ \\
\midrule
\textbf{Average} & 79.2/66.5 & 71.5/55.7 & 76.8/62.8 & 79.4/77.3 & 81.3/71.0 \\
\bottomrule
\end{tabular}%
}
\end{table*}

\subsection{Results with Descriptive prompting}

Under descriptive prompting (Table~\ref{tab:vlm_fix_desc_direct}), performance is generally strong and more balanced between standard and inverse settings than in rule-style prompting, with an overall mean of 70.99\%/68.79\%. GPT-5.2 is the strongest model overall (88.83\%/85.25\%), followed by GPT-4.1 (84.67\%/82.58\%); among open models, Qwen3-VL-32B is best (78.90\%/78.90\%). Across games, Dots and Boxes is the easiest on average (81.29\%/80.66\%), while Connect Four remains the most challenging (59.15\%/56.04\%).

\begin{table*}[t]
\centering
\caption{VLM-Fix Direct Results under Descriptive Prompting (Canonical Rendering). Each cell reports \textbf{Standard/Inverse accuracy (\%)} for the corresponding game. The \textbf{Average} column is the per-model mean across the four games.}
\label{tab:vlm_fix_desc_direct}
\resizebox{0.75\textwidth}{!}{%
\begin{tabular}{lccccc}
\toprule
Model & Tic-Tac-Toe & Reversi & Connect Four & Dots and Boxes & Average \\
\midrule
GPT-4.1 & 85.0/87.0 & 91.0/92.0 & 72.0/60.3 & 90.7/91.0 & 84.7/82.6 \\
GPT-5.2 & 91.7/91.7 & 94.3/94.0 & 73.0/67.0 & 96.3/88.3 & 88.8/85.2 \\
Sonnet-4.0 & 70.0/58.7 & 70.3/50.7 & 64.3/53.7 & 95.7/96.0 & 75.1/64.8 \\
Sonnet-4.5 & 61.3/54.7 & 88.0/52.0 & 71.0/56.0 & 99.7/98.3 & 80.0/65.2 \\
Qwen2.5-VL-3B & 52.3/52.3 & 72.8/72.9 & 49.7/49.6 & 48.7/48.8 & 55.9/55.9 \\
Qwen2.5-VL-7B & 62.2/62.3 & 77.2/77.2 & 55.9/55.8 & 85.8/85.7 & 70.2/70.2 \\
InternVL3.5-4B & 54.8/54.8 & 70.8/70.8 & 50.2/50.1 & 79.9/79.9 & 63.9/63.9 \\
InternVL3.5-8B & 72.0/72.0 & 62.4/62.5 & 55.0/55.1 & 89.8/89.8 & 69.8/69.8 \\
InternVL3.5-14B & 69.2/69.3 & 56.0/56.0 & 58.8/58.7 & 66.4/66.5 & 62.6/62.6 \\
Qwen3-VL-4B & 53.6/53.7 & 79.8/79.8 & 51.2/51.2 & 58.6/58.5 & 60.8/60.8 \\
Qwen3-VL-8B & 54.1/54.1 & 72.0/72.0 & 53.7/53.7 & 81.8/81.8 & 65.4/65.4 \\
Qwen3-VL-32B & 73.2/73.2 & 87.8/87.8 & 62.8/62.8 & 91.8/91.8 & 78.9/78.9 \\
Molmo2-4B & 54.8/54.8 & 80.1/80.1 & 54.8/54.8 & 71.6/71.6 & 65.3/65.3 \\
Molmo2-8B & 62.7/62.7 & 89.6/89.6 & 55.8/55.8 & 81.4/81.4 & 72.4/72.4 \\
\midrule
\textbf{Average} & 65.5/64.4 & 78.0/74.1 & 59.1/56.0 & 81.3/80.7 & 71.0/68.8 \\
\bottomrule
\end{tabular}%
}
\end{table*}

\FloatBarrier
\subsection{Text-only Input Results}

\paragraph{Prompt format for text-only inputs.}
Each text-only query concatenates an instruction prompt and a text board with a blank-line separator. The templates below follow the same content as \texttt{vlm\_fix\_text\_only\_prompts.tex}, using ASCII board rendering for compatibility in this paper.

\noindent\fbox{%
\begin{minipage}{0.97\linewidth}
\ttfamily\footnotesize
<PROMPT>\\[0.6em]
Board:\\
<ASCII\_BOARD>
\end{minipage}%
}

\vspace{0.6em}
\noindent\fbox{%
\begin{minipage}{0.97\linewidth}
\ttfamily\footnotesize
Answer with only <label1> or <label2>. Do not add any other text.
\end{minipage}%
}

\paragraph{Inverse-rule direct prompt templates.}
\begin{PromptBlock}[Tic-Tac-Toe (3x3)]
You are given a 3x3 grid for a two-player game. Players are X and O. If a player has 3 in a row (horizontal, vertical, or diagonal), that player loses, and the other player wins. The game has ended. Who is the winner? Answer with only X or O. Do not add any other text.

Board:
 X | O | X 
---+---+---
 O | X | O 
---+---+---
 X |   | O 
\end{PromptBlock}

\vspace{0.6em}
\begin{PromptBlock}[Reversi (5x5)]
You are given a 5x5 grid for a two-player game. Players are Black and White. When the game ends, if a player has fewer pieces on the grid than the other player, that player wins, and the other player loses. The game has ended. Who is the winner? Answer with only Black or White. Do not add any other text.

Board:
+---+---+---+---+---+
| B | W | B | W | B |
+---+---+---+---+---+
| W | B | W | B | W |
+---+---+---+---+---+
| B | W | B | W | B |
+---+---+---+---+---+
| W | B | W | B | W |
+---+---+---+---+---+
| B | W | B | W | B |
+---+---+---+---+---+
\end{PromptBlock}

\vspace{0.6em}
\begin{PromptBlock}[Connect Four (4x4)]
You are given a 4x4 vertical grid for a two-player game. Players are Red and Yellow. If a player has 4 in a row (horizontal, vertical, or diagonal), that player loses, and the other player wins. The game has ended. Who is the winner? Answer with only Red or Yellow. Do not add any other text.

Board:
 R | Y | R | Y 
---+---+---+---
 Y | R | Y | R 
---+---+---+---
 R | Y | R | Y 
---+---+---+---
 Y | R | Y | R 
\end{PromptBlock}

\vspace{0.6em}
\begin{PromptBlock}[Dots and Boxes (6x6)]
You are given a 6x6 dot grid for a two-player game. Players are A and B. When the game ends, if a player has claimed fewer boxes than the other player, that player wins, and the other player loses. The game has ended. Who is the winner? Answer with only A or B. Do not add any other text.

Board:
o---o---o---o---o---o---o
| A | B | A | B | A | B |
o---o---o---o---o---o---o
| B | A | B | A | B | A |
o---o---o---o---o---o---o
| A | B | A | B | A | B |
o---o---o---o---o---o---o
| B | A | B | A | B | A |
o---o---o---o---o---o---o
| A | B | A | B | A | B |
o---o---o---o---o---o---o
| B | A | B | A | B | A |
o---o---o---o---o---o---o
\end{PromptBlock}

\FloatBarrier

In the text-only setting (Table~\ref{tab:vlm_fix_text_only_direct}), performance is lower than image-based results with CoT or descriptive prompting, and it shows a clear standard--inverse asymmetry (overall 69.22\%/58.62\%). Dots and Boxes has the highest standard accuracy (71.20\%), while Connect Four is lower but more balanced (64.92\%/57.07\%). GPT-5.2 remains the strongest model (94.83\%/90.42\%). Most open models show larger inverse degradation.

\begin{table*}[t]
\centering
\caption{VLM-Fix text-only direct results (canonical text board and standard prompt). API models are aligned to the same reduced 300-state winner-only seed subset used in their direct API runs. Each cell reports \textbf{Standard/Inverse accuracy (\%)} for the corresponding game. The \textbf{Average} column is the per-model mean across the four games.}
\label{tab:vlm_fix_text_only_direct}
\resizebox{\textwidth}{!}{%
\begin{tabular}{lccccc}
\toprule
Model & Tic-Tac-Toe & Reversi & Connect Four & Dots and Boxes & Average \\
\midrule
GPT-4.1 & 96.3/57.7 & 90.0/91.3 & 88.3/40.0 & 93.0/95.7 & 91.9/71.2 \\
GPT-5.2 & 98.3/93.7 & 95.3/95.7 & 87.7/82.0 & 97.7/90.3 & 94.8/90.4 \\
Sonnet-4.0 & 86.0/71.7 & 93.0/93.0 & 82.3/80.3 & 100.0/98.3 & 90.3/85.8 \\
Sonnet-4.5 & 77.7/67.3 & 99.0/95.0 & 66.7/71.0 & 100.0/99.7 & 85.8/83.2 \\
Qwen2.5-VL-3B & 51.5/50.0 & 50.3/50.0 & 49.8/50.2 & 57.2/42.5 & 52.2/48.2 \\
Qwen2.5-VL-7B & 63.0/45.5 & 70.3/36.7 & 70.2/49.8 & 70.5/34.2 & 68.5/41.5 \\
InternVL3.5-4B & 50.2/49.8 & 51.5/48.8 & 54.8/46.2 & 57.8/46.8 & 53.6/47.9 \\
InternVL3.5-8B & 55.0/52.8 & 73.2/52.2 & 61.2/58.3 & 54.5/50.5 & 61.0/53.5 \\
InternVL3.5-14B & 72.0/58.7 & 59.7/59.7 & 53.3/52.5 & 47.2/51.0 & 58.0/55.5 \\
Qwen3-VL-4B & 58.8/52.7 & 50.7/54.0 & 54.3/55.3 & 52.8/38.3 & 54.2/50.1 \\
Qwen3-VL-8B & 65.3/53.2 & 57.2/36.8 & 55.8/49.2 & 57.5/40.7 & 59.0/45.0 \\
Qwen3-VL-32B & 73.0/61.2 & 67.2/58.2 & 65.7/58.5 & 65.5/58.3 & 67.8/59.0 \\
Molmo2-4B & 62.7/48.2 & 57.5/26.2 & 54.2/50.5 & 66.5/41.0 & 60.2/41.5 \\
Molmo2-8B & 67.7/56.5 & 78.3/38.8 & 64.5/55.2 & 76.7/41.3 & 71.8/48.0 \\
\midrule
\textbf{Average} & 69.8/58.5 & 70.9/59.7 & 64.9/57.1 & 71.2/59.2 & 69.2/58.6 \\
\bottomrule
\end{tabular}%
}
\end{table*}

\subsection{Input-Order Marginals for Open Models}

Tables~\ref{tab:vlm_fix_direct_order_marginals_open} and~\ref{tab:vlm_fix_cot_order_marginals_open} summarize image-first versus text-first marginals for the 10 open-weight models, averaging over the four VLM-Fix games. In the direct setting, text-first ordering tends to amplify the standard--inverse gap for the base, glyph, and checkerboard conditions, whereas alias prompting and descriptive prompting remain much more balanced. Under CoT prompting, the same qualitative pattern persists, but the order effect is generally smaller and alias prompting remains the most balanced condition overall.

\begin{table*}[t]
\centering
\caption{VLM-Fix order marginals for direct prompting, averaged over the four games and reported for the 10 open-weight models only. Each cell reports standard/inverse accuracy (\%) for a fixed condition under image-first versus text-first input order.}
\label{tab:vlm_fix_direct_order_marginals_open}
\resizebox{\textwidth}{!}{%
\begin{tabular}{l|cc|cc|cc|cc|cc|cc}
\toprule
Model & \multicolumn{2}{c}{Base} & \multicolumn{2}{c}{Glyph} & \multicolumn{2}{c}{Checkerboard} & \multicolumn{2}{c}{Alias} & \multicolumn{2}{c}{SemAlias} & \multicolumn{2}{c}{Descriptive} \\
 & Img-first & Text-first & Img-first & Text-first & Img-first & Text-first & Img-first & Text-first & Img-first & Text-first & Img-first & Text-first \\
\midrule
Qwen2.5-VL-3B & 53.5/49.3 & 53.8/48.1 & 53.6/47.6 & 53.0/47.1 & 54.1/49.2 & 54.2/46.6 & 55.0/54.2 & 50.9/50.2 & 53.2/52.2 & 51.3/48.2 & 56.1/56.2 & 55.6/55.5 \\
Qwen2.5-VL-7B & 61.2/44.2 & 65.0/39.3 & 59.6/46.2 & 68.7/44.7 & 61.1/43.3 & 60.8/44.2 & 60.8/59.4 & 60.0/59.1 & 60.4/44.5 & 57.0/44.9 & 72.2/72.2 & 68.3/68.2 \\
InternVL3.5-4B & 54.5/42.8 & 68.9/37.3 & 56.7/44.8 & 66.1/48.2 & 52.5/46.5 & 61.1/44.1 & 56.0/56.6 & 66.0/52.4 & 56.7/43.2 & 64.2/45.3 & 58.5/58.5 & 69.4/69.3 \\
InternVL3.5-8B & 61.8/52.8 & 70.2/51.0 & 63.4/50.9 & 70.1/54.2 & 60.7/51.5 & 68.2/50.8 & 59.2/53.8 & 65.4/58.8 & 64.2/49.4 & 66.5/46.2 & 67.5/67.5 & 72.1/72.1 \\
InternVL3.5-14B & 56.4/48.1 & 61.3/55.1 & 61.3/53.2 & 70.3/58.5 & 56.7/48.8 & 61.5/51.8 & 54.0/58.8 & 61.6/67.4 & 54.4/51.6 & 62.0/53.5 & 59.4/59.5 & 65.8/65.8 \\
Qwen3-VL-4B & 56.4/49.0 & 62.1/44.4 & 53.2/52.4 & 58.6/46.3 & 55.2/49.6 & 59.1/45.7 & 55.0/54.0 & 55.6/51.8 & 56.8/46.3 & 53.8/44.9 & 61.2/61.2 & 60.4/60.4 \\
Qwen3-VL-8B & 64.0/49.2 & 63.5/40.0 & 60.9/48.5 & 68.0/41.6 & 62.7/50.6 & 58.2/40.9 & 58.8/54.2 & 59.9/56.0 & 61.8/46.9 & 62.0/39.1 & 66.1/66.1 & 64.7/64.6 \\
Qwen3-VL-32B & 69.8/58.2 & 71.0/57.1 & 63.4/56.0 & 69.2/59.5 & 65.5/58.5 & 69.9/57.6 & 67.6/68.8 & 69.7/78.2 & 68.5/50.1 & 69.0/56.9 & 79.3/79.3 & 78.5/78.5 \\
Molmo2-4B & 57.4/32.2 & 69.2/36.0 & 57.7/39.7 & 66.5/38.3 & 60.3/35.5 & 69.5/40.1 & 51.2/61.0 & 52.3/52.2 & 58.5/41.4 & 59.2/46.1 & 65.5/65.5 & 65.2/65.2 \\
Molmo2-8B & 66.5/44.2 & 76.5/39.9 & 65.4/44.5 & 75.0/47.9 & 65.5/46.9 & 75.0/42.6 & 62.7/57.0 & 72.1/61.5 & 66.2/45.3 & 73.3/49.7 & 73.5/73.5 & 71.3/71.3 \\
\midrule
\textbf{Average} & 60.2/47.0 & 66.2/44.8 & 59.5/48.4 & 66.5/48.6 & 59.4/48.0 & 63.8/46.4 & 58.0/57.8 & 61.3/58.8 & 60.1/47.1 & 61.9/47.5 & 65.9/66.0 & 67.1/67.1 \\
\bottomrule
\end{tabular}%
}
\end{table*}

\begin{table*}[t]
\centering
\caption{VLM-Fix order marginals for CoT prompting, averaged over the four games and reported for the 10 open-weight models only. Each cell reports standard/inverse accuracy (\%) for a fixed condition under image-first versus text-first input order.}
\label{tab:vlm_fix_cot_order_marginals_open}
\resizebox{\textwidth}{!}{%
\begin{tabular}{l|cc|cc|cc|cc|cc}
\toprule
Model & \multicolumn{2}{c}{Base} & \multicolumn{2}{c}{Glyph} & \multicolumn{2}{c}{Checkerboard} & \multicolumn{2}{c}{Alias} & \multicolumn{2}{c}{SemAlias} \\
 & Img-first & Text-first & Img-first & Text-first & Img-first & Text-first & Img-first & Text-first & Img-first & Text-first \\
\midrule
Qwen2.5-VL-3B & 57.2/44.3 & 55.8/41.9 & 58.5/43.9 & 58.1/43.4 & 56.5/42.4 & 50.5/36.4 & 59.6/52.0 & 59.2/51.0 & 63.0/48.4 & 59.6/47.3 \\
Qwen2.5-VL-7B & 72.2/30.2 & 72.4/34.1 & 71.7/36.6 & 73.8/42.2 & 70.2/33.3 & 71.1/37.5 & 75.0/71.0 & 70.6/72.0 & 71.0/50.4 & 72.0/49.4 \\
InternVL3.5-4B & 88.0/65.8 & 85.0/70.1 & 88.6/77.2 & 88.0/76.8 & 87.8/73.0 & 86.6/72.6 & 85.2/84.9 & 84.9/86.2 & 87.7/71.1 & 83.5/71.2 \\
InternVL3.5-8B & 88.8/83.5 & 87.4/79.0 & 89.2/86.4 & 88.0/85.3 & 89.5/82.4 & 88.4/78.5 & 88.2/85.4 & 89.7/90.5 & 87.4/78.5 & 91.8/82.0 \\
InternVL3.5-14B & 78.6/71.8 & 76.7/74.9 & 80.4/77.1 & 78.2/75.4 & 77.1/75.6 & 76.4/72.7 & 75.2/75.1 & 73.6/78.2 & 75.8/72.1 & 73.6/64.8 \\
Qwen3-VL-4B & 80.0/77.9 & 81.8/80.6 & 83.9/74.7 & 80.2/79.2 & 87.6/87.4 & 88.4/85.7 & 79.5/80.3 & 82.1/83.3 & 80.3/79.8 & 83.4/84.3 \\
Qwen3-VL-8B & 82.5/78.3 & 80.8/82.1 & 73.1/69.2 & 72.4/71.9 & 93.6/91.4 & 83.8/83.9 & 81.0/83.8 & 81.7/82.1 & 84.9/84.0 & 83.7/83.3 \\
Qwen3-VL-32B & 99.5/99.4 & 96.7/97.0 & 97.8/97.7 & 96.3/96.1 & 100.0/99.8 & 99.0/99.2 & 98.8/99.0 & 97.0/97.8 & 99.5/97.2 & 98.3/96.1 \\
Molmo2-4B & 74.7/45.1 & 75.7/38.7 & 77.9/42.7 & 77.2/38.8 & 74.5/43.6 & 78.0/39.6 & 77.4/73.5 & 76.2/71.5 & 77.0/56.9 & 78.0/59.2 \\
Molmo2-8B & 81.6/48.2 & 80.5/51.7 & 79.4/44.8 & 80.6/47.2 & 77.6/49.6 & 75.4/54.0 & 75.3/76.9 & 79.4/76.7 & 76.8/60.9 & 77.4/44.3 \\
\midrule
\textbf{Average} & 80.3/64.5 & 79.3/65.0 & 80.0/65.0 & 79.3/65.6 & 81.4/67.8 & 79.8/66.0 & 79.5/78.2 & 79.4/78.9 & 80.3/69.9 & 80.1/68.2 \\
\bottomrule
\end{tabular}%
}
\end{table*}

\subsection{VLMBias Results}
\label{app:vlmbias-results}

Table~\ref{tab:vlms_are_biased_acc_bias_full} reports the full model-wise averages across the four \textit{VLMBias} subsets under the four input configurations.

\begin{table*}[t]
\centering
\caption{VLMBias results averaged across the four subsets (Animals, Flags, Game Boards, Logos). Accuracy and Bias are shown side-by-side for four input configurations. Cells report percentages; for non-baseline columns, subscripts show Holm-adjusted paired McNemar $p$-values versus Base (within the same metric). Within each model row, \textcolor{GapLowBlue}{blue} marks the best value and \textcolor{GapHighRed}{red} marks the worst value.}
\label{tab:vlms_are_biased_acc_bias_full}
\resizebox{\textwidth}{!}{%
\begin{tabular}{l|cccc|cccc}
\toprule
 & \multicolumn{4}{c|}{Accuracy $\uparrow$} & \multicolumn{4}{c}{Bias $\downarrow$} \\
Model & Base & Flip & Alias & Flip+Alias & Base & Flip & Alias & Flip+Alias \\
\midrule
GPT-4.1 & \textcolor{GapHighRed}{9.9} & 10.5$_{\scriptscriptstyle(\!\sim\!1.00)}$ & \textcolor{GapLowBlue}{12.0$_{\scriptscriptstyle(\!\sim\!0.09)}$} & 10.7$_{\scriptscriptstyle(\!\sim\!1.00)}$ & \textcolor{GapHighRed}{81.0} & \textcolor{GapLowBlue}{71.7$_{\scriptscriptstyle(\!<\!0.01)}$} & 73.0$_{\scriptscriptstyle(\!<\!0.01)}$ & 76.1$_{\scriptscriptstyle(\!<\!0.01)}$ \\
GPT-5.2 & \textcolor{GapHighRed}{20.2} & 21.0$_{\scriptscriptstyle(\!\sim\!1.00)}$ & 26.2$_{\scriptscriptstyle(\!<\!0.01)}$ & \textcolor{GapLowBlue}{32.0$_{\scriptscriptstyle(\!<\!0.01)}$} & \textcolor{GapHighRed}{76.8} & 70.8$_{\scriptscriptstyle(\!<\!0.01)}$ & 68.7$_{\scriptscriptstyle(\!<\!0.01)}$ & \textcolor{GapLowBlue}{54.4$_{\scriptscriptstyle(\!<\!0.01)}$} \\
Sonnet-4.0 & \textcolor{GapHighRed}{6.2} & 6.5$_{\scriptscriptstyle(\!\sim\!1.00)}$ & 7.2$_{\scriptscriptstyle(\!\sim\!0.35)}$ & \textcolor{GapLowBlue}{9.8$_{\scriptscriptstyle(\!<\!0.01)}$} & \textcolor{GapHighRed}{89.3} & 84.0$_{\scriptscriptstyle(\!<\!0.01)}$ & 87.5$_{\scriptscriptstyle(\!\sim\!0.08)}$ & \textcolor{GapLowBlue}{79.5$_{\scriptscriptstyle(\!<\!0.01)}$} \\
Sonnet-4.5 & \textcolor{GapHighRed}{6.9} & 8.6$_{\scriptscriptstyle(\!\sim\!0.07)}$ & 9.3$_{\scriptscriptstyle(\!<\!0.01)}$ & \textcolor{GapLowBlue}{17.3$_{\scriptscriptstyle(\!<\!0.01)}$} & \textcolor{GapHighRed}{92.3} & 87.2$_{\scriptscriptstyle(\!<\!0.01)}$ & 85.9$_{\scriptscriptstyle(\!<\!0.01)}$ & \textcolor{GapLowBlue}{77.0$_{\scriptscriptstyle(\!<\!0.01)}$} \\
Qwen2.5-VL-3B & \textcolor{GapHighRed}{7.4} & 7.4$_{\scriptscriptstyle(\!\sim\!1.00)}$ & 8.3$_{\scriptscriptstyle(\!\sim\!0.42)}$ & \textcolor{GapLowBlue}{10.7$_{\scriptscriptstyle(\!<\!0.01)}$} & \textcolor{GapHighRed}{66.4} & 63.6$_{\scriptscriptstyle(\!<\!0.01)}$ & 59.5$_{\scriptscriptstyle(\!<\!0.01)}$ & \textcolor{GapLowBlue}{53.8$_{\scriptscriptstyle(\!<\!0.01)}$} \\
Qwen2.5-VL-7B & 9.5 & \textcolor{GapHighRed}{8.8$_{\scriptscriptstyle(\!\sim\!0.76)}$} & 15.9$_{\scriptscriptstyle(\!<\!0.01)}$ & \textcolor{GapLowBlue}{18.7$_{\scriptscriptstyle(\!<\!0.01)}$} & \textcolor{GapHighRed}{75.5} & 72.4$_{\scriptscriptstyle(\!<\!0.01)}$ & 69.8$_{\scriptscriptstyle(\!<\!0.01)}$ & \textcolor{GapLowBlue}{62.7$_{\scriptscriptstyle(\!<\!0.01)}$} \\
InternVL3.5-4B & \textcolor{GapHighRed}{6.6} & 11.1$_{\scriptscriptstyle(\!<\!0.01)}$ & 9.4$_{\scriptscriptstyle(\!<\!0.01)}$ & \textcolor{GapLowBlue}{14.0$_{\scriptscriptstyle(\!<\!0.01)}$} & \textcolor{GapHighRed}{68.9} & 61.3$_{\scriptscriptstyle(\!<\!0.01)}$ & 55.0$_{\scriptscriptstyle(\!<\!0.01)}$ & \textcolor{GapLowBlue}{30.7$_{\scriptscriptstyle(\!<\!0.01)}$} \\
InternVL3.5-8B & \textcolor{GapHighRed}{7.2} & 9.1$_{\scriptscriptstyle(\!\sim\!0.04)}$ & 11.9$_{\scriptscriptstyle(\!<\!0.01)}$ & \textcolor{GapLowBlue}{27.9$_{\scriptscriptstyle(\!<\!0.01)}$} & \textcolor{GapHighRed}{77.1} & 65.9$_{\scriptscriptstyle(\!<\!0.01)}$ & 67.5$_{\scriptscriptstyle(\!<\!0.01)}$ & \textcolor{GapLowBlue}{40.7$_{\scriptscriptstyle(\!<\!0.01)}$} \\
InternVL3.5-14B & \textcolor{GapHighRed}{13.0} & 14.7$_{\scriptscriptstyle(\!\sim\!0.07)}$ & 15.8$_{\scriptscriptstyle(\!<\!0.01)}$ & \textcolor{GapLowBlue}{31.7$_{\scriptscriptstyle(\!<\!0.01)}$} & \textcolor{GapHighRed}{65.6} & 61.5$_{\scriptscriptstyle(\!<\!0.01)}$ & 56.4$_{\scriptscriptstyle(\!<\!0.01)}$ & \textcolor{GapLowBlue}{40.8$_{\scriptscriptstyle(\!<\!0.01)}$} \\
Qwen3-VL-4B & \textcolor{GapHighRed}{8.2} & 10.4$_{\scriptscriptstyle(\!<\!0.01)}$ & 10.8$_{\scriptscriptstyle(\!<\!0.01)}$ & \textcolor{GapLowBlue}{16.7$_{\scriptscriptstyle(\!<\!0.01)}$} & \textcolor{GapHighRed}{83.5} & 77.1$_{\scriptscriptstyle(\!<\!0.01)}$ & 74.9$_{\scriptscriptstyle(\!<\!0.01)}$ & \textcolor{GapLowBlue}{65.2$_{\scriptscriptstyle(\!<\!0.01)}$} \\
Qwen3-VL-8B & \textcolor{GapHighRed}{5.6} & 8.6$_{\scriptscriptstyle(\!<\!0.01)}$ & 7.6$_{\scriptscriptstyle(\!<\!0.01)}$ & \textcolor{GapLowBlue}{14.2$_{\scriptscriptstyle(\!<\!0.01)}$} & \textcolor{GapHighRed}{86.5} & 81.8$_{\scriptscriptstyle(\!<\!0.01)}$ & 79.2$_{\scriptscriptstyle(\!<\!0.01)}$ & \textcolor{GapLowBlue}{72.9$_{\scriptscriptstyle(\!<\!0.01)}$} \\
Qwen3-VL-32B & \textcolor{GapHighRed}{8.3} & 11.0$_{\scriptscriptstyle(\!<\!0.01)}$ & 12.3$_{\scriptscriptstyle(\!<\!0.01)}$ & \textcolor{GapLowBlue}{16.8$_{\scriptscriptstyle(\!<\!0.01)}$} & \textcolor{GapHighRed}{88.0} & 83.6$_{\scriptscriptstyle(\!<\!0.01)}$ & 84.2$_{\scriptscriptstyle(\!<\!0.01)}$ & \textcolor{GapLowBlue}{77.6$_{\scriptscriptstyle(\!<\!0.01)}$} \\
Molmo2-4B & \textcolor{GapHighRed}{25.2} & 29.9$_{\scriptscriptstyle(\!<\!0.01)}$ & 34.6$_{\scriptscriptstyle(\!<\!0.01)}$ & \textcolor{GapLowBlue}{36.4$_{\scriptscriptstyle(\!<\!0.01)}$} & \textcolor{GapHighRed}{66.2} & 57.3$_{\scriptscriptstyle(\!<\!0.01)}$ & 54.7$_{\scriptscriptstyle(\!<\!0.01)}$ & \textcolor{GapLowBlue}{44.7$_{\scriptscriptstyle(\!<\!0.01)}$} \\
Molmo2-8B & \textcolor{GapHighRed}{27.8} & 27.9$_{\scriptscriptstyle(\!\sim\!1.00)}$ & 28.3$_{\scriptscriptstyle(\!\sim\!1.00)}$ & \textcolor{GapLowBlue}{32.7$_{\scriptscriptstyle(\!<\!0.01)}$} & 56.5 & 54.9$_{\scriptscriptstyle(\!\sim\!0.63)}$ & \textcolor{GapHighRed}{57.0$_{\scriptscriptstyle(\!\sim\!1.00)}$} & \textcolor{GapLowBlue}{48.7$_{\scriptscriptstyle(\!<\!0.01)}$} \\
\midrule
\textbf{Average} & 11.6 & 13.3 & 15.0 & 20.7 & 76.7 & 70.9 & 69.5 & 58.9 \\
\bottomrule
\end{tabular}%
}
\end{table*}

Task-wise accuracy in \textit{VLMBias} (Tables~\ref{tab:vlms_are_biased_acc_bias_animals}, \ref{tab:vlms_are_biased_acc_bias_flags}, \ref{tab:vlms_are_biased_acc_bias_game_boards}, and \ref{tab:vlms_are_biased_acc_bias_logos}) is strongly subset-dependent and generally improves with alias-based prompting. Averaged across models, Flags shows the highest accuracy overall (25.89, 24.23, 30.57, and 29.02 across the four configurations), while Animals is the most difficult under base settings (3.62/5.70) but rises markedly with aliasing, especially Flip+Alias (22.17). Game Boards and Logos remain in a mid-low range, improving from 11.39 to 16.16 and from 13.85 to 16.65/15.72, respectively, indicating that prompt and image perturbations can help accuracy but do not fully close the gap across subsets.

\providecolor{GapHighRed}{HTML}{B35757}
\providecolor{GapLowBlue}{HTML}{2E5FA7}
\begin{table*}[t]
\centering
\caption{VLMBias results for the Animals subset (same metrics and annotations as the VLMBias tables below).}
\label{tab:vlms_are_biased_acc_bias_animals}
\resizebox{\textwidth}{!}{%
\begin{tabular}{l|cccc|cccc}
\toprule
 & \multicolumn{4}{c|}{Accuracy $\uparrow$} & \multicolumn{4}{c}{Bias $\downarrow$} \\
Model & Base & Flip & Alias & Flip+Alias & Base & Flip & Alias & Flip+Alias \\
\midrule
GPT-4.1 & 9.9 & \textcolor{GapHighRed}{2.0$_{\scriptscriptstyle(\!<\!0.01)}$} & \textcolor{GapLowBlue}{14.1$_{\scriptscriptstyle(\!\sim\!0.07)}$} & 5.9$_{\scriptscriptstyle(\!\sim\!0.07)}$ & 72.5 & 65.2$_{\scriptscriptstyle(\!<\!0.01)}$ & \textcolor{GapLowBlue}{58.6$_{\scriptscriptstyle(\!<\!0.01)}$} & \textcolor{GapHighRed}{76.6$_{\scriptscriptstyle(\!\sim\!0.17)}$} \\
GPT-5.2 & \textcolor{GapHighRed}{4.6} & 9.2$_{\scriptscriptstyle(\!<\!0.01)}$ & 17.0$_{\scriptscriptstyle(\!<\!0.01)}$ & \textcolor{GapLowBlue}{39.9$_{\scriptscriptstyle(\!<\!0.01)}$} & \textcolor{GapHighRed}{94.1} & 89.6$_{\scriptscriptstyle(\!<\!0.01)}$ & 79.5$_{\scriptscriptstyle(\!<\!0.01)}$ & \textcolor{GapLowBlue}{53.3$_{\scriptscriptstyle(\!<\!0.01)}$} \\
Sonnet-4.0 & \textcolor{GapHighRed}{0.0} & 0.2$_{\scriptscriptstyle(\!\sim\!1.00)}$ & 0.0$_{\scriptscriptstyle(\!\sim\!1.00)}$ & \textcolor{GapLowBlue}{3.7$_{\scriptscriptstyle(\!<\!0.01)}$} & \textcolor{GapHighRed}{99.8} & 97.4$_{\scriptscriptstyle(\!<\!0.01)}$ & 98.5$_{\scriptscriptstyle(\!\sim\!0.03)}$ & \textcolor{GapLowBlue}{92.3$_{\scriptscriptstyle(\!<\!0.01)}$} \\
Sonnet-4.5 & \textcolor{GapHighRed}{0.0} & 0.5$_{\scriptscriptstyle(\!\sim\!0.50)}$ & 2.0$_{\scriptscriptstyle(\!<\!0.01)}$ & \textcolor{GapLowBlue}{17.8$_{\scriptscriptstyle(\!<\!0.01)}$} & \textcolor{GapHighRed}{100.0} & 97.6$_{\scriptscriptstyle(\!<\!0.01)}$ & 95.1$_{\scriptscriptstyle(\!<\!0.01)}$ & \textcolor{GapLowBlue}{79.5$_{\scriptscriptstyle(\!<\!0.01)}$} \\
Qwen2.5-VL-3B & \textcolor{GapHighRed}{0.0} & 0.2$_{\scriptscriptstyle(\!\sim\!1.00)}$ & 0.0$_{\scriptscriptstyle(\!\sim\!1.00)}$ & \textcolor{GapLowBlue}{1.5$_{\scriptscriptstyle(\!\sim\!0.05)}$} & 96.7 & \textcolor{GapHighRed}{97.3$_{\scriptscriptstyle(\!\sim\!1.00)}$} & 96.7$_{\scriptscriptstyle(\!\sim\!1.00)}$ & \textcolor{GapLowBlue}{90.5$_{\scriptscriptstyle(\!<\!0.01)}$} \\
Qwen2.5-VL-7B & \textcolor{GapHighRed}{0.0} & 4.0$_{\scriptscriptstyle(\!<\!0.01)}$ & 10.3$_{\scriptscriptstyle(\!<\!0.01)}$ & \textcolor{GapLowBlue}{20.5$_{\scriptscriptstyle(\!<\!0.01)}$} & \textcolor{GapHighRed}{97.6} & 93.2$_{\scriptscriptstyle(\!<\!0.01)}$ & 86.3$_{\scriptscriptstyle(\!<\!0.01)}$ & \textcolor{GapLowBlue}{75.1$_{\scriptscriptstyle(\!<\!0.01)}$} \\
InternVL3.5-4B & \textcolor{GapHighRed}{0.0} & 4.2$_{\scriptscriptstyle(\!<\!0.01)}$ & 0.0$_{\scriptscriptstyle(\!\sim\!1.00)}$ & \textcolor{GapLowBlue}{15.0$_{\scriptscriptstyle(\!<\!0.01)}$} & \textcolor{GapHighRed}{98.4} & 86.8$_{\scriptscriptstyle(\!<\!0.01)}$ & 72.5$_{\scriptscriptstyle(\!<\!0.01)}$ & \textcolor{GapLowBlue}{22.0$_{\scriptscriptstyle(\!<\!0.01)}$} \\
InternVL3.5-8B & \textcolor{GapHighRed}{0.0} & 1.5$_{\scriptscriptstyle(\!\sim\!0.03)}$ & 1.1$_{\scriptscriptstyle(\!\sim\!0.06)}$ & \textcolor{GapLowBlue}{41.2$_{\scriptscriptstyle(\!<\!0.01)}$} & \textcolor{GapHighRed}{99.3} & 92.9$_{\scriptscriptstyle(\!<\!0.01)}$ & 90.7$_{\scriptscriptstyle(\!<\!0.01)}$ & \textcolor{GapLowBlue}{50.0$_{\scriptscriptstyle(\!<\!0.01)}$} \\
InternVL3.5-14B & \textcolor{GapHighRed}{0.0} & 2.6$_{\scriptscriptstyle(\!<\!0.01)}$ & 1.8$_{\scriptscriptstyle(\!<\!0.01)}$ & \textcolor{GapLowBlue}{43.2$_{\scriptscriptstyle(\!<\!0.01)}$} & \textcolor{GapHighRed}{99.1} & 94.7$_{\scriptscriptstyle(\!<\!0.01)}$ & 82.4$_{\scriptscriptstyle(\!<\!0.01)}$ & \textcolor{GapLowBlue}{49.8$_{\scriptscriptstyle(\!<\!0.01)}$} \\
Qwen3-VL-4B & \textcolor{GapHighRed}{0.0} & 2.0$_{\scriptscriptstyle(\!<\!0.01)}$ & 3.7$_{\scriptscriptstyle(\!<\!0.01)}$ & \textcolor{GapLowBlue}{17.6$_{\scriptscriptstyle(\!<\!0.01)}$} & \textcolor{GapHighRed}{99.8} & 94.0$_{\scriptscriptstyle(\!<\!0.01)}$ & 93.6$_{\scriptscriptstyle(\!<\!0.01)}$ & \textcolor{GapLowBlue}{76.2$_{\scriptscriptstyle(\!<\!0.01)}$} \\
Qwen3-VL-8B & \textcolor{GapHighRed}{0.5} & 6.4$_{\scriptscriptstyle(\!<\!0.01)}$ & 3.7$_{\scriptscriptstyle(\!<\!0.01)}$ & \textcolor{GapLowBlue}{17.2$_{\scriptscriptstyle(\!<\!0.01)}$} & \textcolor{GapHighRed}{99.5} & 93.6$_{\scriptscriptstyle(\!<\!0.01)}$ & 90.8$_{\scriptscriptstyle(\!<\!0.01)}$ & \textcolor{GapLowBlue}{82.4$_{\scriptscriptstyle(\!<\!0.01)}$} \\
Qwen3-VL-32B & \textcolor{GapHighRed}{2.7} & 6.4$_{\scriptscriptstyle(\!<\!0.01)}$ & 10.3$_{\scriptscriptstyle(\!<\!0.01)}$ & \textcolor{GapLowBlue}{16.3$_{\scriptscriptstyle(\!<\!0.01)}$} & \textcolor{GapHighRed}{96.9} & 93.6$_{\scriptscriptstyle(\!<\!0.01)}$ & 89.4$_{\scriptscriptstyle(\!<\!0.01)}$ & \textcolor{GapLowBlue}{83.7$_{\scriptscriptstyle(\!<\!0.01)}$} \\
Molmo2-4B & \textcolor{GapHighRed}{12.1} & 18.7$_{\scriptscriptstyle(\!<\!0.01)}$ & 27.8$_{\scriptscriptstyle(\!<\!0.01)}$ & \textcolor{GapLowBlue}{33.0$_{\scriptscriptstyle(\!<\!0.01)}$} & \textcolor{GapHighRed}{83.3} & 74.0$_{\scriptscriptstyle(\!<\!0.01)}$ & 69.6$_{\scriptscriptstyle(\!<\!0.01)}$ & \textcolor{GapLowBlue}{55.3$_{\scriptscriptstyle(\!<\!0.01)}$} \\
Molmo2-8B & \textcolor{GapHighRed}{20.9} & 22.0$_{\scriptscriptstyle(\!\sim\!1.00)}$ & 27.1$_{\scriptscriptstyle(\!<\!0.01)}$ & \textcolor{GapLowBlue}{37.7$_{\scriptscriptstyle(\!<\!0.01)}$} & 68.3 & \textcolor{GapHighRed}{75.5$_{\scriptscriptstyle(\!<\!0.01)}$} & 70.0$_{\scriptscriptstyle(\!\sim\!0.68)}$ & \textcolor{GapLowBlue}{59.0$_{\scriptscriptstyle(\!<\!0.01)}$} \\
\midrule
\textbf{Average} & 3.6 & 5.7 & 8.5 & 22.2 & 93.2 & 88.9 & 83.8 & 67.5 \\
\bottomrule
\end{tabular}%
}
\end{table*}

\providecolor{GapHighRed}{HTML}{B35757}
\providecolor{GapLowBlue}{HTML}{2E5FA7}
\begin{table*}[t]
\centering
\caption{VLMBias results for the Flags subset (same format as Table~\ref{tab:vlms_are_biased_acc_bias_animals}).}
\label{tab:vlms_are_biased_acc_bias_flags}
\resizebox{\textwidth}{!}{%
\begin{tabular}{l|cccc|cccc}
\toprule
 & \multicolumn{4}{c|}{Accuracy $\uparrow$} & \multicolumn{4}{c}{Bias $\downarrow$} \\
Model & Base & Flip & Alias & Flip+Alias & Base & Flip & Alias & Flip+Alias \\
\midrule
GPT-4.1 & \textcolor{GapLowBlue}{13.3} & 9.2$_{\scriptscriptstyle(\!\sim\!0.03)}$ & 13.3$_{\scriptscriptstyle(\!\sim\!1.00)}$ & \textcolor{GapHighRed}{8.3$_{\scriptscriptstyle(\!\sim\!0.03)}$} & \textcolor{GapLowBlue}{83.8} & 90.0$_{\scriptscriptstyle(\!<\!0.01)}$ & 85.0$_{\scriptscriptstyle(\!\sim\!0.75)}$ & \textcolor{GapHighRed}{91.7$_{\scriptscriptstyle(\!<\!0.01)}$} \\
GPT-5.2 & \textcolor{GapHighRed}{22.9} & 24.6$_{\scriptscriptstyle(\!\sim\!1.00)}$ & 26.7$_{\scriptscriptstyle(\!\sim\!0.20)}$ & \textcolor{GapLowBlue}{31.2$_{\scriptscriptstyle(\!\sim\!0.04)}$} & \textcolor{GapHighRed}{73.8} & 64.6$_{\scriptscriptstyle(\!\sim\!0.02)}$ & 68.8$_{\scriptscriptstyle(\!\sim\!0.03)}$ & \textcolor{GapLowBlue}{52.5$_{\scriptscriptstyle(\!<\!0.01)}$} \\
Sonnet-4.0 & 25.0 & \textcolor{GapHighRed}{22.9$_{\scriptscriptstyle(\!\sim\!1.00)}$} & \textcolor{GapLowBlue}{30.0$_{\scriptscriptstyle(\!\sim\!0.21)}$} & 28.3$_{\scriptscriptstyle(\!\sim\!1.00)}$ & \textcolor{GapHighRed}{74.2} & 74.2$_{\scriptscriptstyle(\!\sim\!1.00)}$ & 70.0$_{\scriptscriptstyle(\!\sim\!0.39)}$ & \textcolor{GapLowBlue}{67.1$_{\scriptscriptstyle(\!\sim\!0.07)}$} \\
Sonnet-4.5 & 24.2 & \textcolor{GapHighRed}{20.0$_{\scriptscriptstyle(\!\sim\!0.98)}$} & \textcolor{GapLowBlue}{27.1$_{\scriptscriptstyle(\!\sim\!0.98)}$} & 26.2$_{\scriptscriptstyle(\!\sim\!1.00)}$ & 73.8 & \textcolor{GapHighRed}{74.2$_{\scriptscriptstyle(\!\sim\!1.00)}$} & 69.6$_{\scriptscriptstyle(\!\sim\!0.21)}$ & \textcolor{GapLowBlue}{66.7$_{\scriptscriptstyle(\!\sim\!0.16)}$} \\
Qwen2.5-VL-3B & 24.2 & \textcolor{GapHighRed}{22.9$_{\scriptscriptstyle(\!\sim\!1.00)}$} & 24.2$_{\scriptscriptstyle(\!\sim\!1.00)}$ & \textcolor{GapLowBlue}{30.8$_{\scriptscriptstyle(\!\sim\!0.04)}$} & \textcolor{GapHighRed}{35.4} & 33.3$_{\scriptscriptstyle(\!\sim\!1.00)}$ & 32.5$_{\scriptscriptstyle(\!\sim\!0.84)}$ & \textcolor{GapLowBlue}{29.2$_{\scriptscriptstyle(\!\sim\!0.24)}$} \\
Qwen2.5-VL-7B & 27.9 & \textcolor{GapHighRed}{24.6$_{\scriptscriptstyle(\!\sim\!0.86)}$} & \textcolor{GapLowBlue}{31.7$_{\scriptscriptstyle(\!\sim\!0.73)}$} & 25.8$_{\scriptscriptstyle(\!\sim\!1.00)}$ & 45.0 & \textcolor{GapHighRed}{46.7$_{\scriptscriptstyle(\!\sim\!1.00)}$} & \textcolor{GapLowBlue}{41.7$_{\scriptscriptstyle(\!\sim\!1.00)}$} & 42.5$_{\scriptscriptstyle(\!\sim\!1.00)}$ \\
InternVL3.5-4B & \textcolor{GapHighRed}{23.8} & 26.2$_{\scriptscriptstyle(\!\sim\!1.00)}$ & \textcolor{GapLowBlue}{29.2$_{\scriptscriptstyle(\!<\!0.01)}$} & 26.7$_{\scriptscriptstyle(\!\sim\!1.00)}$ & 33.8 & \textcolor{GapHighRed}{41.2$_{\scriptscriptstyle(\!\sim\!0.12)}$} & \textcolor{GapLowBlue}{30.0$_{\scriptscriptstyle(\!\sim\!0.60)}$} & 35.0$_{\scriptscriptstyle(\!\sim\!1.00)}$ \\
InternVL3.5-8B & \textcolor{GapHighRed}{27.1} & 27.9$_{\scriptscriptstyle(\!\sim\!1.00)}$ & 34.6$_{\scriptscriptstyle(\!\sim\!0.06)}$ & \textcolor{GapLowBlue}{37.1$_{\scriptscriptstyle(\!\sim\!0.06)}$} & \textcolor{GapHighRed}{47.5} & 43.8$_{\scriptscriptstyle(\!\sim\!0.65)}$ & 42.5$_{\scriptscriptstyle(\!\sim\!0.65)}$ & \textcolor{GapLowBlue}{30.0$_{\scriptscriptstyle(\!<\!0.01)}$} \\
InternVL3.5-14B & 35.0 & \textcolor{GapHighRed}{33.3$_{\scriptscriptstyle(\!\sim\!1.00)}$} & \textcolor{GapLowBlue}{40.0$_{\scriptscriptstyle(\!\sim\!0.58)}$} & 35.0$_{\scriptscriptstyle(\!\sim\!1.00)}$ & 33.8 & \textcolor{GapHighRed}{36.2$_{\scriptscriptstyle(\!\sim\!1.00)}$} & \textcolor{GapLowBlue}{30.8$_{\scriptscriptstyle(\!\sim\!1.00)}$} & 31.7$_{\scriptscriptstyle(\!\sim\!1.00)}$ \\
Qwen3-VL-4B & 19.2 & \textcolor{GapHighRed}{16.2$_{\scriptscriptstyle(\!\sim\!0.55)}$} & \textcolor{GapLowBlue}{21.7$_{\scriptscriptstyle(\!\sim\!0.58)}$} & 16.7$_{\scriptscriptstyle(\!\sim\!0.58)}$ & 61.3 & \textcolor{GapHighRed}{65.4$_{\scriptscriptstyle(\!\sim\!0.52)}$} & 61.3$_{\scriptscriptstyle(\!\sim\!1.00)}$ & \textcolor{GapLowBlue}{57.5$_{\scriptscriptstyle(\!\sim\!0.84)}$} \\
Qwen3-VL-8B & 17.9 & \textcolor{GapHighRed}{15.8$_{\scriptscriptstyle(\!\sim\!0.53)}$} & \textcolor{GapLowBlue}{23.3$_{\scriptscriptstyle(\!\sim\!0.01)}$} & 23.3$_{\scriptscriptstyle(\!\sim\!0.04)}$ & \textcolor{GapHighRed}{65.8} & 65.0$_{\scriptscriptstyle(\!\sim\!1.00)}$ & 65.0$_{\scriptscriptstyle(\!\sim\!1.00)}$ & \textcolor{GapLowBlue}{62.5$_{\scriptscriptstyle(\!\sim\!0.91)}$} \\
Qwen3-VL-32B & \textcolor{GapHighRed}{20.0} & 22.1$_{\scriptscriptstyle(\!\sim\!0.85)}$ & 28.3$_{\scriptscriptstyle(\!<\!0.01)}$ & \textcolor{GapLowBlue}{30.0$_{\scriptscriptstyle(\!<\!0.01)}$} & \textcolor{GapHighRed}{73.3} & 68.3$_{\scriptscriptstyle(\!\sim\!0.12)}$ & 67.5$_{\scriptscriptstyle(\!<\!0.01)}$ & \textcolor{GapLowBlue}{62.5$_{\scriptscriptstyle(\!<\!0.01)}$} \\
Molmo2-4B & 40.8 & \textcolor{GapHighRed}{38.8$_{\scriptscriptstyle(\!\sim\!1.00)}$} & \textcolor{GapLowBlue}{51.7$_{\scriptscriptstyle(\!<\!0.01)}$} & 51.7$_{\scriptscriptstyle(\!<\!0.01)}$ & \textcolor{GapHighRed}{37.1} & 34.6$_{\scriptscriptstyle(\!\sim\!0.65)}$ & \textcolor{GapLowBlue}{27.5$_{\scriptscriptstyle(\!<\!0.01)}$} & 27.5$_{\scriptscriptstyle(\!<\!0.01)}$ \\
Molmo2-8B & 41.2 & \textcolor{GapHighRed}{34.6$_{\scriptscriptstyle(\!\sim\!0.37)}$} & \textcolor{GapLowBlue}{46.2$_{\scriptscriptstyle(\!\sim\!0.35)}$} & 35.0$_{\scriptscriptstyle(\!\sim\!0.37)}$ & 33.8 & 37.9$_{\scriptscriptstyle(\!\sim\!0.62)}$ & \textcolor{GapLowBlue}{32.5$_{\scriptscriptstyle(\!\sim\!1.00)}$} & \textcolor{GapHighRed}{40.8$_{\scriptscriptstyle(\!\sim\!0.16)}$} \\
\midrule
\textbf{Average} & 25.9 & 24.2 & 30.6 & 29.0 & 55.1 & 55.4 & 51.8 & 49.8 \\
\bottomrule
\end{tabular}%
}
\end{table*}

\providecolor{GapHighRed}{HTML}{B35757}
\providecolor{GapLowBlue}{HTML}{2E5FA7}
\begin{table*}[t]
\centering
\caption{VLMBias results for the Game Boards subset (same format as Table~\ref{tab:vlms_are_biased_acc_bias_animals}).}
\label{tab:vlms_are_biased_acc_bias_game_boards}
\resizebox{\textwidth}{!}{%
\begin{tabular}{l|cccc|cccc}
\toprule
 & \multicolumn{4}{c|}{Accuracy $\uparrow$} & \multicolumn{4}{c}{Bias $\downarrow$} \\
Model & Base & Flip & Alias & Flip+Alias & Base & Flip & Alias & Flip+Alias \\
\midrule
GPT-4.1 & 0.0 & 0.0$_{\scriptscriptstyle(\!\sim\!1.00)}$ & 0.0$_{\scriptscriptstyle(\!\sim\!1.00)}$ & 0.0$_{\scriptscriptstyle(\!\sim\!1.00)}$ & \textcolor{GapHighRed}{97.6} & \textcolor{GapLowBlue}{86.3$_{\scriptscriptstyle(\!<\!0.01)}$} & 97.6$_{\scriptscriptstyle(\!\sim\!1.00)}$ & 88.1$_{\scriptscriptstyle(\!<\!0.01)}$ \\
GPT-5.2 & 35.7 & \textcolor{GapLowBlue}{41.1$_{\scriptscriptstyle(\!\sim\!0.73)}$} & 39.9$_{\scriptscriptstyle(\!\sim\!1.00)}$ & \textcolor{GapHighRed}{35.1$_{\scriptscriptstyle(\!\sim\!1.00)}$} & \textcolor{GapHighRed}{56.5} & 49.4$_{\scriptscriptstyle(\!\sim\!0.20)}$ & \textcolor{GapLowBlue}{47.6$_{\scriptscriptstyle(\!\sim\!0.08)}$} & 50.0$_{\scriptscriptstyle(\!\sim\!0.20)}$ \\
Sonnet-4.0 & \textcolor{GapHighRed}{14.9} & 16.7$_{\scriptscriptstyle(\!\sim\!1.00)}$ & 16.1$_{\scriptscriptstyle(\!\sim\!1.00)}$ & \textcolor{GapLowBlue}{24.4$_{\scriptscriptstyle(\!\sim\!0.08)}$} & \textcolor{GapHighRed}{75.6} & 74.4$_{\scriptscriptstyle(\!\sim\!1.00)}$ & 74.4$_{\scriptscriptstyle(\!\sim\!1.00)}$ & \textcolor{GapLowBlue}{61.3$_{\scriptscriptstyle(\!<\!0.01)}$} \\
Sonnet-4.5 & \textcolor{GapHighRed}{20.8} & 28.0$_{\scriptscriptstyle(\!\sim\!0.13)}$ & 28.6$_{\scriptscriptstyle(\!\sim\!0.06)}$ & \textcolor{GapLowBlue}{30.4$_{\scriptscriptstyle(\!\sim\!0.10)}$} & \textcolor{GapHighRed}{78.6} & 69.0$_{\scriptscriptstyle(\!\sim\!0.04)}$ & 70.2$_{\scriptscriptstyle(\!\sim\!0.04)}$ & \textcolor{GapLowBlue}{65.5$_{\scriptscriptstyle(\!\sim\!0.02)}$} \\
Qwen2.5-VL-3B & \textcolor{GapHighRed}{6.5} & 8.9$_{\scriptscriptstyle(\!\sim\!0.91)}$ & 17.9$_{\scriptscriptstyle(\!<\!0.01)}$ & \textcolor{GapLowBlue}{22.6$_{\scriptscriptstyle(\!<\!0.01)}$} & \textcolor{GapHighRed}{50.0} & 48.2$_{\scriptscriptstyle(\!\sim\!1.00)}$ & 26.2$_{\scriptscriptstyle(\!<\!0.01)}$ & \textcolor{GapLowBlue}{19.0$_{\scriptscriptstyle(\!<\!0.01)}$} \\
Qwen2.5-VL-7B & 12.5 & \textcolor{GapHighRed}{10.7$_{\scriptscriptstyle(\!\sim\!1.00)}$} & 13.1$_{\scriptscriptstyle(\!\sim\!1.00)}$ & \textcolor{GapLowBlue}{14.3$_{\scriptscriptstyle(\!\sim\!1.00)}$} & \textcolor{GapHighRed}{83.3} & 81.0$_{\scriptscriptstyle(\!\sim\!0.85)}$ & 78.6$_{\scriptscriptstyle(\!\sim\!0.61)}$ & \textcolor{GapLowBlue}{75.0$_{\scriptscriptstyle(\!\sim\!0.17)}$} \\
InternVL3.5-4B & 7.1 & \textcolor{GapLowBlue}{22.6$_{\scriptscriptstyle(\!<\!0.01)}$} & \textcolor{GapHighRed}{4.8$_{\scriptscriptstyle(\!\sim\!1.00)}$} & 13.1$_{\scriptscriptstyle(\!\sim\!0.49)}$ & \textcolor{GapHighRed}{35.1} & 20.8$_{\scriptscriptstyle(\!<\!0.01)}$ & 28.6$_{\scriptscriptstyle(\!\sim\!0.01)}$ & \textcolor{GapLowBlue}{13.1$_{\scriptscriptstyle(\!<\!0.01)}$} \\
InternVL3.5-8B & \textcolor{GapHighRed}{14.3} & 15.5$_{\scriptscriptstyle(\!\sim\!1.00)}$ & \textcolor{GapLowBlue}{22.6$_{\scriptscriptstyle(\!\sim\!0.15)}$} & 16.7$_{\scriptscriptstyle(\!\sim\!1.00)}$ & \textcolor{GapHighRed}{32.7} & 19.6$_{\scriptscriptstyle(\!<\!0.01)}$ & 14.9$_{\scriptscriptstyle(\!<\!0.01)}$ & \textcolor{GapLowBlue}{6.5$_{\scriptscriptstyle(\!<\!0.01)}$} \\
InternVL3.5-14B & \textcolor{GapHighRed}{18.5} & \textcolor{GapLowBlue}{23.2$_{\scriptscriptstyle(\!\sim\!1.00)}$} & 20.2$_{\scriptscriptstyle(\!\sim\!1.00)}$ & 20.2$_{\scriptscriptstyle(\!\sim\!1.00)}$ & \textcolor{GapHighRed}{48.8} & 36.9$_{\scriptscriptstyle(\!<\!0.01)}$ & 45.2$_{\scriptscriptstyle(\!\sim\!0.57)}$ & \textcolor{GapLowBlue}{32.1$_{\scriptscriptstyle(\!<\!0.01)}$} \\
Qwen3-VL-4B & \textcolor{GapHighRed}{4.2} & 4.2$_{\scriptscriptstyle(\!\sim\!1.00)}$ & \textcolor{GapLowBlue}{9.5$_{\scriptscriptstyle(\!\sim\!0.47)}$} & 9.5$_{\scriptscriptstyle(\!\sim\!0.47)}$ & \textcolor{GapHighRed}{82.1} & 78.6$_{\scriptscriptstyle(\!\sim\!0.57)}$ & \textcolor{GapLowBlue}{57.1$_{\scriptscriptstyle(\!<\!0.01)}$} & 60.7$_{\scriptscriptstyle(\!<\!0.01)}$ \\
Qwen3-VL-8B & \textcolor{GapLowBlue}{9.5} & \textcolor{GapHighRed}{3.6$_{\scriptscriptstyle(\!\sim\!0.04)}$} & 4.8$_{\scriptscriptstyle(\!\sim\!0.54)}$ & 6.0$_{\scriptscriptstyle(\!\sim\!0.61)}$ & 82.1 & \textcolor{GapHighRed}{83.3$_{\scriptscriptstyle(\!\sim\!1.00)}$} & 78.6$_{\scriptscriptstyle(\!\sim\!1.00)}$ & \textcolor{GapLowBlue}{76.2$_{\scriptscriptstyle(\!\sim\!0.31)}$} \\
Qwen3-VL-32B & 8.9 & 10.7$_{\scriptscriptstyle(\!\sim\!1.00)}$ & \textcolor{GapHighRed}{7.1$_{\scriptscriptstyle(\!\sim\!1.00)}$} & \textcolor{GapLowBlue}{14.9$_{\scriptscriptstyle(\!\sim\!0.66)}$} & 73.8 & 72.6$_{\scriptscriptstyle(\!\sim\!1.00)}$ & \textcolor{GapHighRed}{81.0$_{\scriptscriptstyle(\!\sim\!0.32)}$} & \textcolor{GapLowBlue}{62.5$_{\scriptscriptstyle(\!\sim\!0.19)}$} \\
Molmo2-4B & \textcolor{GapHighRed}{6.5} & 7.1$_{\scriptscriptstyle(\!\sim\!1.00)}$ & 10.7$_{\scriptscriptstyle(\!\sim\!1.00)}$ & \textcolor{GapLowBlue}{16.7$_{\scriptscriptstyle(\!\sim\!0.06)}$} & \textcolor{GapHighRed}{79.2} & 78.6$_{\scriptscriptstyle(\!\sim\!1.00)}$ & 58.3$_{\scriptscriptstyle(\!<\!0.01)}$ & \textcolor{GapLowBlue}{54.8$_{\scriptscriptstyle(\!<\!0.01)}$} \\
Molmo2-8B & \textcolor{GapHighRed}{0.0} & 0.0$_{\scriptscriptstyle(\!\sim\!1.00)}$ & \textcolor{GapLowBlue}{2.4$_{\scriptscriptstyle(\!\sim\!0.75)}$} & 2.4$_{\scriptscriptstyle(\!\sim\!0.75)}$ & \textcolor{GapHighRed}{57.1} & 56.0$_{\scriptscriptstyle(\!\sim\!1.00)}$ & 53.6$_{\scriptscriptstyle(\!\sim\!0.12)}$ & \textcolor{GapLowBlue}{51.2$_{\scriptscriptstyle(\!\sim\!0.01)}$} \\
\midrule
\textbf{Average} & 11.4 & 13.7 & 14.1 & 16.2 & 66.6 & 61.1 & 58.0 & 51.1 \\
\bottomrule
\end{tabular}%
}
\end{table*}

\providecolor{GapHighRed}{HTML}{B35757}
\providecolor{GapLowBlue}{HTML}{2E5FA7}
\begin{table*}[t]
\centering
\caption{VLMBias results for the Logos subset (same format as Table~\ref{tab:vlms_are_biased_acc_bias_animals}).}
\label{tab:vlms_are_biased_acc_bias_logos}
\resizebox{\textwidth}{!}{%
\begin{tabular}{l|cccc|cccc}
\toprule
 & \multicolumn{4}{c|}{Accuracy $\uparrow$} & \multicolumn{4}{c}{Bias $\downarrow$} \\
Model & Base & Flip & Alias & Flip+Alias & Base & Flip & Alias & Flip+Alias \\
\midrule
GPT-4.1 & \textcolor{GapHighRed}{12.1} & \textcolor{GapLowBlue}{26.8$_{\scriptscriptstyle(\!<\!0.01)}$} & 13.3$_{\scriptscriptstyle(\!\sim\!0.92)}$ & 22.7$_{\scriptscriptstyle(\!<\!0.01)}$ & \textcolor{GapHighRed}{83.8} & 63.8$_{\scriptscriptstyle(\!<\!0.01)}$ & 74.9$_{\scriptscriptstyle(\!<\!0.01)}$ & \textcolor{GapLowBlue}{61.6$_{\scriptscriptstyle(\!<\!0.01)}$} \\
GPT-5.2 & \textcolor{GapLowBlue}{33.1} & 26.3$_{\scriptscriptstyle(\!\sim\!0.03)}$ & 32.6$_{\scriptscriptstyle(\!\sim\!1.00)}$ & \textcolor{GapHighRed}{20.8$_{\scriptscriptstyle(\!<\!0.01)}$} & \textcolor{GapHighRed}{63.8} & \textcolor{GapLowBlue}{58.2$_{\scriptscriptstyle(\!\sim\!0.03)}$} & 63.0$_{\scriptscriptstyle(\!\sim\!1.00)}$ & 58.7$_{\scriptscriptstyle(\!\sim\!0.04)}$ \\
Sonnet-4.0 & \textcolor{GapHighRed}{0.0} & \textcolor{GapLowBlue}{1.2$_{\scriptscriptstyle(\!\sim\!0.38)}$} & 0.0$_{\scriptscriptstyle(\!\sim\!1.00)}$ & 1.2$_{\scriptscriptstyle(\!\sim\!0.38)}$ & \textcolor{GapHighRed}{89.6} & \textcolor{GapLowBlue}{75.8$_{\scriptscriptstyle(\!<\!0.01)}$} & 88.4$_{\scriptscriptstyle(\!\sim\!1.00)}$ & 77.1$_{\scriptscriptstyle(\!<\!0.01)}$ \\
Sonnet-4.5 & \textcolor{GapHighRed}{0.5} & 4.6$_{\scriptscriptstyle(\!<\!0.01)}$ & 0.7$_{\scriptscriptstyle(\!\sim\!1.00)}$ & \textcolor{GapLowBlue}{6.0$_{\scriptscriptstyle(\!<\!0.01)}$} & \textcolor{GapHighRed}{98.3} & 88.4$_{\scriptscriptstyle(\!<\!0.01)}$ & 89.6$_{\scriptscriptstyle(\!<\!0.01)}$ & \textcolor{GapLowBlue}{84.3$_{\scriptscriptstyle(\!<\!0.01)}$} \\
Qwen2.5-VL-3B & \textcolor{GapLowBlue}{7.7} & 7.2$_{\scriptscriptstyle(\!\sim\!1.00)}$ & \textcolor{GapHighRed}{6.3$_{\scriptscriptstyle(\!\sim\!1.00)}$} & 6.3$_{\scriptscriptstyle(\!\sim\!1.00)}$ & \textcolor{GapHighRed}{51.0} & 43.0$_{\scriptscriptstyle(\!<\!0.01)}$ & 39.6$_{\scriptscriptstyle(\!<\!0.01)}$ & \textcolor{GapLowBlue}{33.8$_{\scriptscriptstyle(\!<\!0.01)}$} \\
Qwen2.5-VL-7B & 10.1 & \textcolor{GapHighRed}{5.3$_{\scriptscriptstyle(\!<\!0.01)}$} & \textcolor{GapLowBlue}{15.5$_{\scriptscriptstyle(\!<\!0.01)}$} & 14.0$_{\scriptscriptstyle(\!<\!0.01)}$ & \textcolor{GapHighRed}{60.9} & 56.5$_{\scriptscriptstyle(\!<\!0.01)}$ & 60.9$_{\scriptscriptstyle(\!\sim\!1.00)}$ & \textcolor{GapLowBlue}{53.1$_{\scriptscriptstyle(\!<\!0.01)}$} \\
InternVL3.5-4B & \textcolor{GapHighRed}{5.1} & 6.8$_{\scriptscriptstyle(\!\sim\!1.00)}$ & \textcolor{GapLowBlue}{12.1$_{\scriptscriptstyle(\!<\!0.01)}$} & 5.8$_{\scriptscriptstyle(\!\sim\!1.00)}$ & \textcolor{GapHighRed}{64.3} & 55.8$_{\scriptscriptstyle(\!<\!0.01)}$ & 57.0$_{\scriptscriptstyle(\!<\!0.01)}$ & \textcolor{GapLowBlue}{46.9$_{\scriptscriptstyle(\!<\!0.01)}$} \\
InternVL3.5-8B & \textcolor{GapHighRed}{2.4} & 5.8$_{\scriptscriptstyle(\!\sim\!0.03)}$ & 8.7$_{\scriptscriptstyle(\!<\!0.01)}$ & \textcolor{GapLowBlue}{9.4$_{\scriptscriptstyle(\!<\!0.01)}$} & \textcolor{GapHighRed}{83.1} & 62.1$_{\scriptscriptstyle(\!<\!0.01)}$ & 72.9$_{\scriptscriptstyle(\!<\!0.01)}$ & \textcolor{GapLowBlue}{48.6$_{\scriptscriptstyle(\!<\!0.01)}$} \\
InternVL3.5-14B & \textcolor{GapHighRed}{15.2} & 16.4$_{\scriptscriptstyle(\!\sim\!0.85)}$ & 18.4$_{\scriptscriptstyle(\!\sim\!0.02)}$ & \textcolor{GapLowBlue}{19.3$_{\scriptscriptstyle(\!<\!0.01)}$} & \textcolor{GapHighRed}{46.9} & 42.3$_{\scriptscriptstyle(\!<\!0.01)}$ & 41.5$_{\scriptscriptstyle(\!<\!0.01)}$ & \textcolor{GapLowBlue}{37.7$_{\scriptscriptstyle(\!<\!0.01)}$} \\
Qwen3-VL-4B & \textcolor{GapHighRed}{14.3} & \textcolor{GapLowBlue}{20.5$_{\scriptscriptstyle(\!<\!0.01)}$} & 14.5$_{\scriptscriptstyle(\!\sim\!1.00)}$ & 18.4$_{\scriptscriptstyle(\!\sim\!0.15)}$ & \textcolor{GapHighRed}{75.4} & 61.1$_{\scriptscriptstyle(\!<\!0.01)}$ & 65.2$_{\scriptscriptstyle(\!<\!0.01)}$ & \textcolor{GapLowBlue}{57.0$_{\scriptscriptstyle(\!<\!0.01)}$} \\
Qwen3-VL-8B & \textcolor{GapHighRed}{3.6} & \textcolor{GapLowBlue}{9.2$_{\scriptscriptstyle(\!<\!0.01)}$} & 4.8$_{\scriptscriptstyle(\!\sim\!0.45)}$ & 8.2$_{\scriptscriptstyle(\!<\!0.01)}$ & \textcolor{GapHighRed}{83.1} & 75.4$_{\scriptscriptstyle(\!<\!0.01)}$ & 72.5$_{\scriptscriptstyle(\!<\!0.01)}$ & \textcolor{GapLowBlue}{65.0$_{\scriptscriptstyle(\!<\!0.01)}$} \\
Qwen3-VL-32B & 8.5 & \textcolor{GapLowBlue}{10.9$_{\scriptscriptstyle(\!\sim\!0.66)}$} & \textcolor{GapHighRed}{7.7$_{\scriptscriptstyle(\!\sim\!0.75)}$} & 10.6$_{\scriptscriptstyle(\!\sim\!0.66)}$ & \textcolor{GapHighRed}{90.6} & \textcolor{GapLowBlue}{83.6$_{\scriptscriptstyle(\!<\!0.01)}$} & 88.4$_{\scriptscriptstyle(\!\sim\!0.04)}$ & 84.5$_{\scriptscriptstyle(\!<\!0.01)}$ \\
Molmo2-4B & 41.1 & \textcolor{GapLowBlue}{48.8$_{\scriptscriptstyle(\!<\!0.01)}$} & 43.5$_{\scriptscriptstyle(\!\sim\!0.35)}$ & \textcolor{GapHighRed}{40.1$_{\scriptscriptstyle(\!\sim\!1.00)}$} & \textcolor{GapHighRed}{55.3} & 39.9$_{\scriptscriptstyle(\!<\!0.01)}$ & 49.3$_{\scriptscriptstyle(\!<\!0.01)}$ & \textcolor{GapLowBlue}{36.7$_{\scriptscriptstyle(\!<\!0.01)}$} \\
Molmo2-8B & 40.3 & \textcolor{GapLowBlue}{43.2$_{\scriptscriptstyle(\!\sim\!1.00)}$} & \textcolor{GapHighRed}{30.0$_{\scriptscriptstyle(\!<\!0.01)}$} & 37.2$_{\scriptscriptstyle(\!\sim\!1.00)}$ & 53.9 & \textcolor{GapLowBlue}{37.2$_{\scriptscriptstyle(\!<\!0.01)}$} & \textcolor{GapHighRed}{55.6$_{\scriptscriptstyle(\!\sim\!0.24)}$} & 38.6$_{\scriptscriptstyle(\!<\!0.01)}$ \\
\midrule
\textbf{Average} & 13.9 & 16.6 & 14.9 & 15.7 & 71.4 & 60.2 & 65.6 & 56.0 \\
\bottomrule
\end{tabular}%
}
\end{table*}

\section{Post-Training}
\label{app:posttrain-additional-models}

\subsection{VLM-Fix Post-Training Details}
\label{sec:appendix_vlmfix_posttrain_details}

\subsubsection{Game/state generation details}
We build balanced terminal-state pools for all four games. Reversi and Dots and Boxes each use \(1024\) generated terminal states. Connect Four also uses \(1024\) states, with balanced win-pattern categories (\(341\) horizontal, \(341\) vertical, \(171\) main diagonal, \(171\) anti-diagonal). For Tic-Tac-Toe, we start from \(920\) exclusive single-line terminal-winner states and add \(104\) sampled top-up slots to reach \(1024\) total state slots.

\subsubsection{Exact split sizes and composition}
\begin{itemize}
  \item \textbf{D1 train (8192):}
  \(4\) games \(\times\) \(1024\) states/game \(\times\) \(2\) targets (winner/loser), standard rule only.
  \item \textbf{D2 train (8192):}
  \(4\) games \(\times\) \(1024\) states/game \(\times\) \(2\) targets, inverse rule only.
  \item \textbf{D3 train (8192):}
  \(2\) games (Tic-Tac-Toe, Reversi) \(\times\) \(2\) rules \(\times\) \(1024\) states/game \(\times\) \(2\) targets.
  \item \textbf{D1 test (2400):}
  canonical evaluation file for D1-trained models: all \(4\) games, inverse rule only, \(300\) benchmark states/game \(\times\) \(2\) targets.
  \item \textbf{D2 test (2400):}
  canonical evaluation file for D2-trained models: all \(4\) games, standard rule only, \(300\) benchmark states/game \(\times\) \(2\) targets.
  \item \textbf{D3 test (2400):}
  Connect Four + Dots and Boxes, both rules, \(300\) states/game/rule \(\times\) \(2\) targets.
\end{itemize}

For the main-text D1/D2 comparison in Figure~\ref{fig:vlm_fix_posttrain_d123_qwen}, we additionally evaluate each trained model on the complementary held-out rule slice drawn from the same benchmark-derived pools. Concretely, D1-trained models are evaluated on both the canonical D1 inverse slice and the held-out standard slice (the D2 test file), while D2-trained models are evaluated on both the canonical D2 standard slice and the held-out inverse slice (the D1 test file). The figure groups these paired evaluations by training split to make same-rule versus cross-rule transfer explicit.

\subsubsection{SFT hyperparameters}
We train Qwen2.5-VL-3B and Qwen2.5-VL-7B with QLoRA SFT (NF4 4-bit quantization, LoRA rank \(8\), all-module targeting) for one epoch, using a global batch size of \(128\), learning rate \(1\times10^{-4}\), cosine decay with warmup ratio \(0.03\), and bf16 precision.

\subsubsection{RLVR hyperparameters}
RLVR uses GRPO with a binary exact-match reward and KL regularization (\(\beta = 10^{-2}\)). We use one epoch, rollout multiplicity \(n=5\), global batch size \(128\), and AdamW in bf16 with learning rate \(1\times10^{-6}\) and weight decay \(10^{-2}\).

\subsubsection{Evaluation protocol}
We evaluate Base, SFT-merged, and RLVR-merged checkpoints on the canonical D1/D2/D3 test files and report exact-match accuracy from extracted final labels. For the main-text D1/D2 comparison, we additionally evaluate on the complementary held-out rule slices to make same-rule versus cross-rule transfer explicit under matched input formatting.

\subsection{Synthetic Leg-Count Transfer Details}
\label{sec:appendix_synthlegs_posttrain_details}

\subsubsection{Dataset construction}
\begin{figure*}[t]
\centering
\setlength{\tabcolsep}{2pt}
\renewcommand{\arraystretch}{1.0}
\begin{tabular}{cccc}
\includegraphics[width=0.18\linewidth]{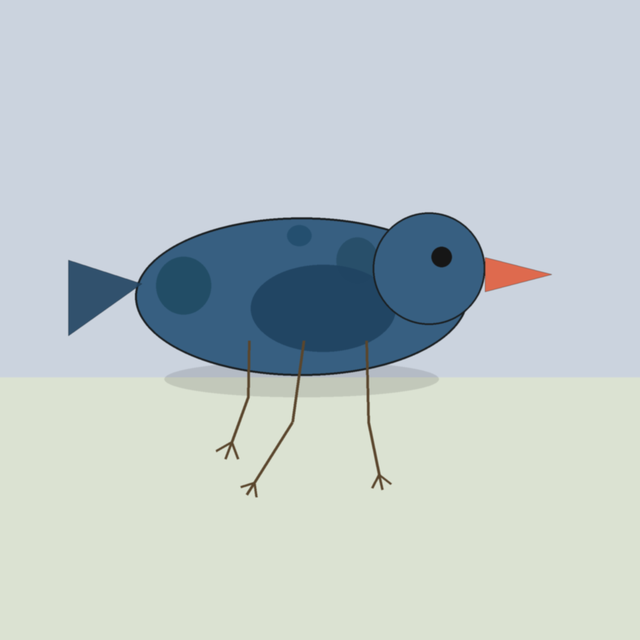} &
\includegraphics[width=0.18\linewidth]{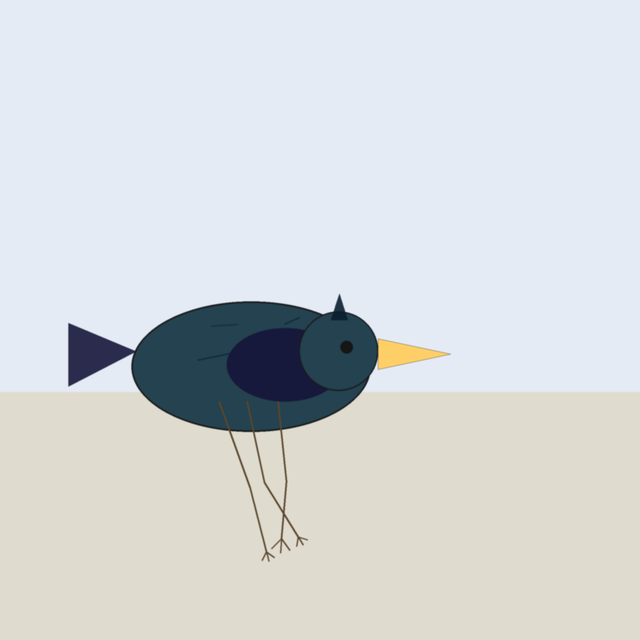} &
\includegraphics[width=0.18\linewidth]{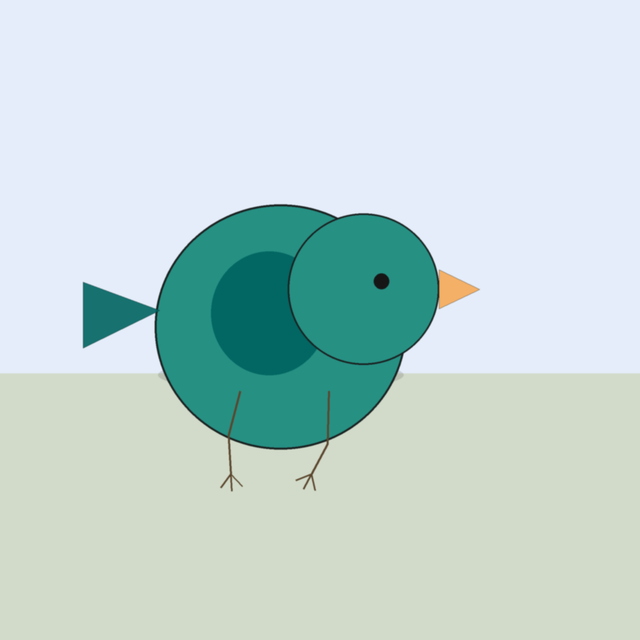} &
\includegraphics[width=0.18\linewidth]{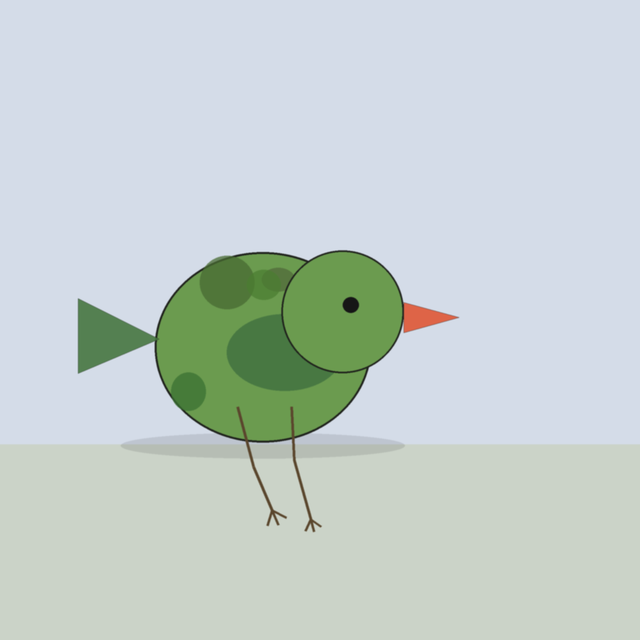} \\
{\tiny Duck} & {\tiny Heron} & {\tiny Owl} & {\tiny Passerine} \\
\includegraphics[width=0.18\linewidth]{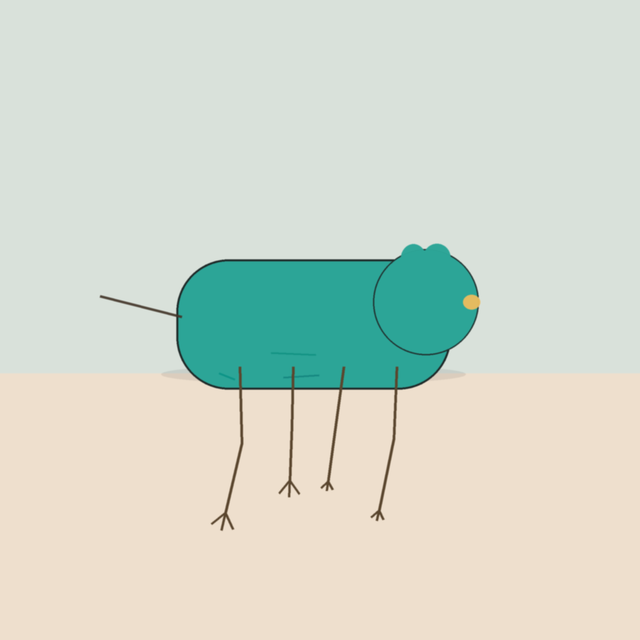} &
\includegraphics[width=0.18\linewidth]{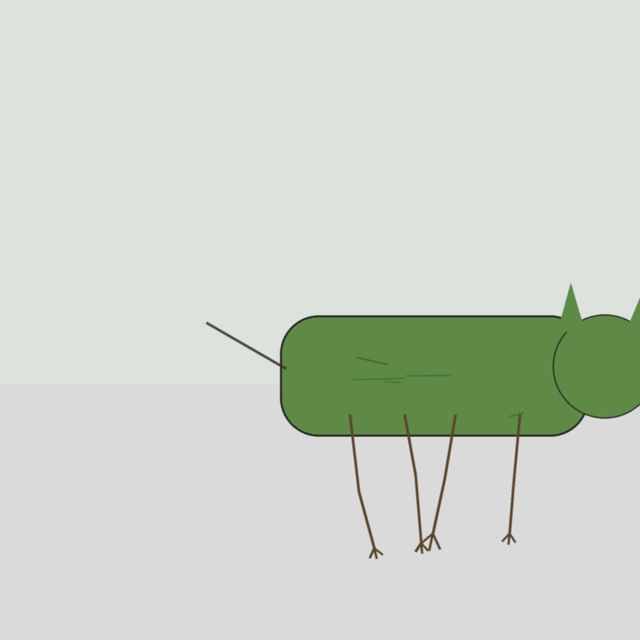} &
\includegraphics[width=0.18\linewidth]{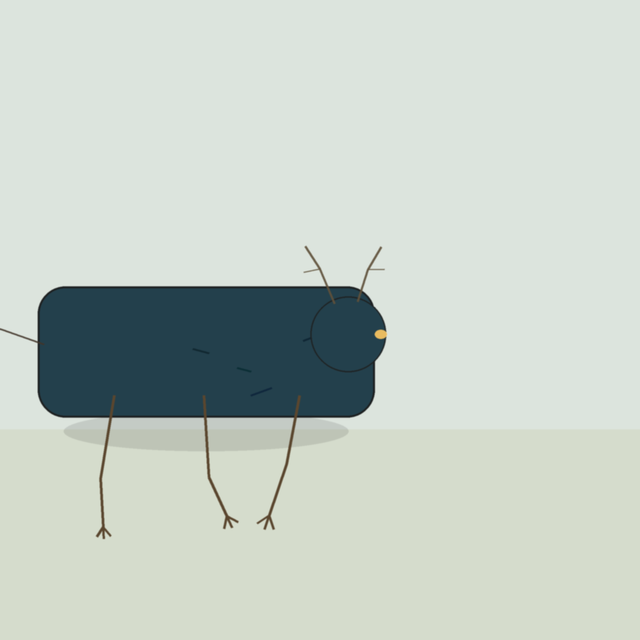} &
\includegraphics[width=0.18\linewidth]{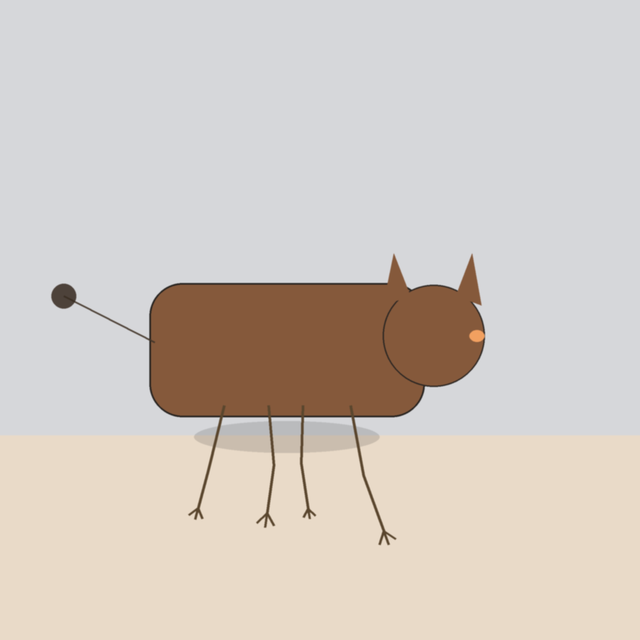} \\
{\tiny Boar} & {\tiny Canine} & {\tiny Deer} & {\tiny Feline}
\end{tabular}
\caption{Additional synthetic glyph examples from the procedurally rendered leg-counting training set. Top: bird categories. Bottom: quadruped-style animal categories.}
\label{fig:appendix_synthleg_gallery}
\end{figure*}

We build a synthetic counting dataset (\texttt{synth-legs-train-8192}) with \(8192\) image--text pairs, balanced 50/50 between bird glyphs and quadruped-style animal glyphs. Bird leg counts are sampled from \(\{1,2,3\}\), and quadruped counts from \(\{3,4,5\}\). We use procedural variation in pose, shape, color, and background to reduce template-specific overfitting. Representative synthetic glyph categories are shown in Figure~\ref{fig:appendix_synthleg_gallery}.

The training instruction is shared across all rows:
\begin{quote}
\small
Count the number of legs in this animal glyph image. Answer with only a number in curly brackets, e.g., \{4\}. Do not add any other text.
\end{quote}

\subsubsection{Post-training protocols}
We post-train Qwen2.5-VL-3B and Qwen2.5-VL-7B with both SFT and RLVR using the same core settings as above. SFT uses QLoRA (NF4, LoRA rank \(8\), one epoch, global batch \(128\), learning rate \(1\times10^{-4}\), bf16). RLVR uses GRPO with binary exact-match reward and KL regularization (\(\beta = 10^{-2}\)), one epoch, rollout \(n=5\), global batch \(128\), and AdamW with learning rate \(1\times10^{-6}\). We evaluate Base/SFT/RLVR with the same answer extractor and held-out test suites.

\subsection{Additional Model Results}

\subsubsection{VLM-Fix transfer}
\begin{figure*}[!t]
\centering
\includegraphics[width=0.49\linewidth,height=0.13\textheight,keepaspectratio]{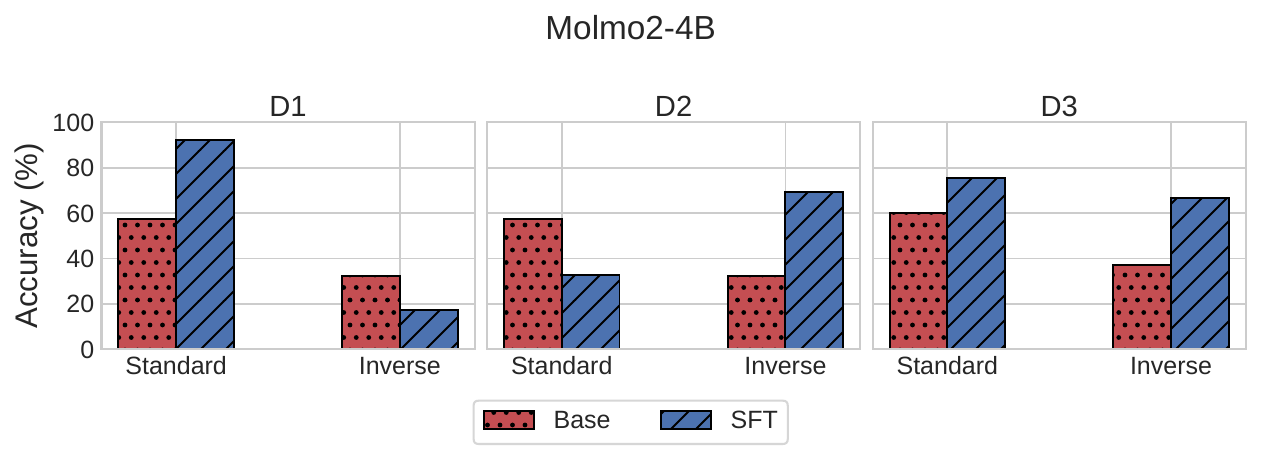}\hfill
\includegraphics[width=0.49\linewidth,height=0.13\textheight,keepaspectratio]{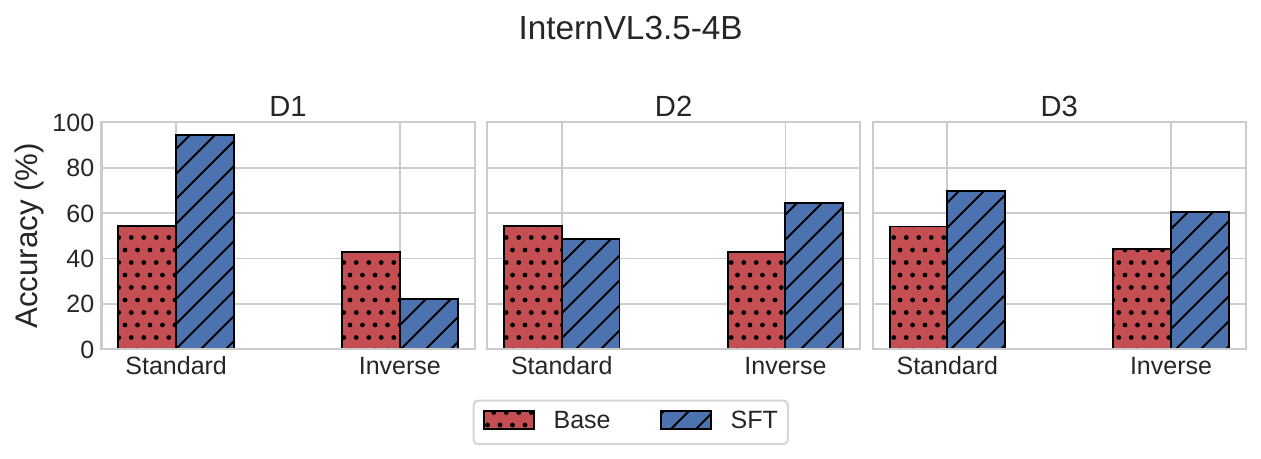}
\caption{Post-training performance on the three VLM-Fix transfer splits (D1--D3) for additional models with available post-training results: Molmo2-4B (left) and InternVL3.5-4B (right). For D1 and D2, we report both held-out standard-rule and inverse-rule evaluation under the original image and base prompt; for D3, we report held-out standard and inverse evaluation on Connect Four and Dots and Boxes.}
\label{fig:appendix_posttrain_additional_d123_abs}
\end{figure*}

Additional post-training transfer results for Molmo2-4B and InternVL3.5-4B on D1--D3 are shown in Figure~\ref{fig:appendix_posttrain_additional_d123_abs}.

\subsubsection{VLMBias Animals transfer}
\begin{figure*}[!t]
\centering
\includegraphics[width=0.49\linewidth]{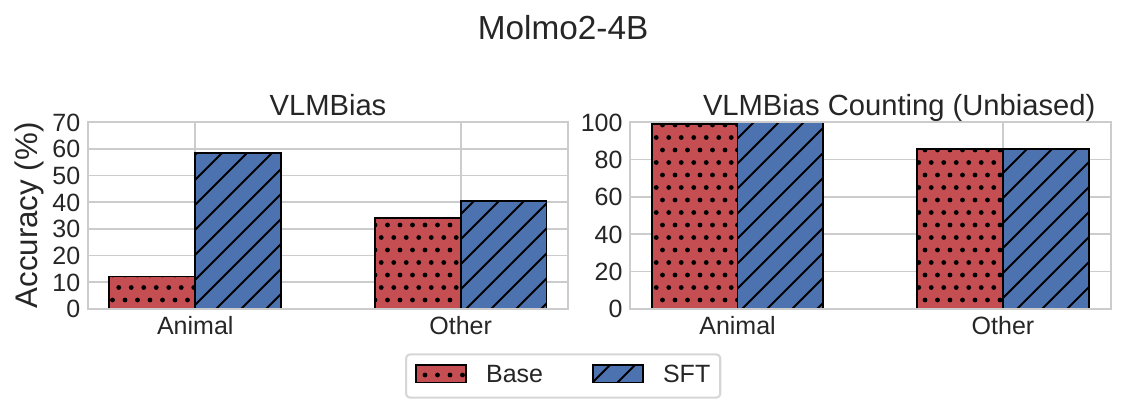}\hfill
\includegraphics[width=0.49\linewidth]{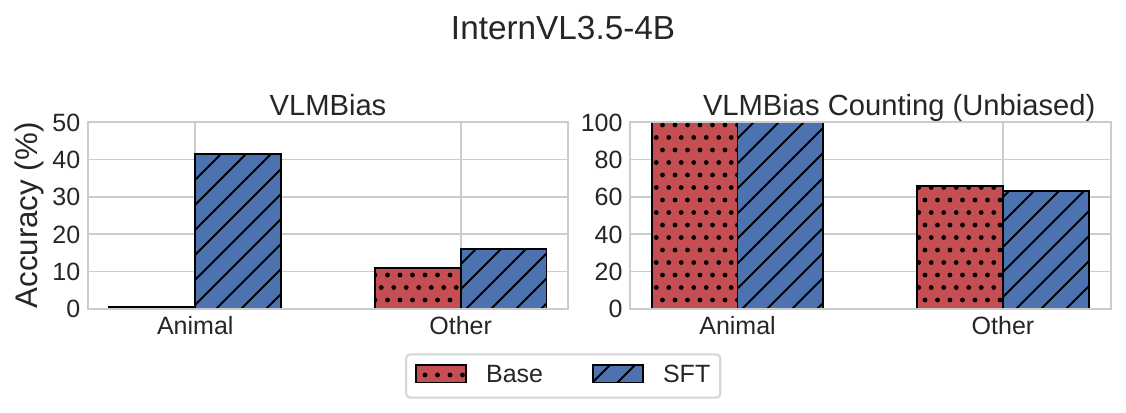}
\caption{Post-training transfer from the synthetic leg-counting dataset for additional models, comparing Base and SFT for Molmo2-4B (left) and InternVL3.5-4B (right). Each panel reports accuracy on \textit{VLMBias} (Animals vs Other) and \textit{VLMBias Counting (Unbiased)} (Animals vs Other).}
\label{fig:appendix_posttrain_additional_vlmsbias_transfer}
\end{figure*}

Additional synthetic-leg-count transfer results for Molmo2-4B and InternVL3.5-4B are shown in Figure~\ref{fig:appendix_posttrain_additional_vlmsbias_transfer}.

\noindent\textbf{Evaluation-coverage note.}
InternVL3.5-4B and Molmo2-4B use full D1/D2/D3 evaluation files (\(n=2400\) each) in this repository snapshot.

\section{Activation Steering}
\label{app:activation_steering_details}

\subsection{VLM-Fix}
\label{app:steering_details_vlmfix}
For each VLM-Fix game, we use a fixed pool of 100 underlying board states and include all paired variants (standard/inverse rule, winner/loser query, image-first/text-first order). Train/test splitting is done at the \emph{state level} (70/30) and repeated over three random splits to prevent leakage across paired variants.

At each of the final 12 decoder layers, we train a two-stage linear router on the training split (rule classifier, then rule-conditional answer classifier), compute donor centroids by rule--answer bucket (with fallback to all examples in a bucket when donor-correct examples are unavailable), and apply a single query-token patch with intervention scale $\alpha=1.0$. Reported curves show mean and standard deviation across the three splits.

\begin{figure*}[!t]
\centering
\includegraphics[width=0.99\linewidth,height=0.24\textheight,keepaspectratio]{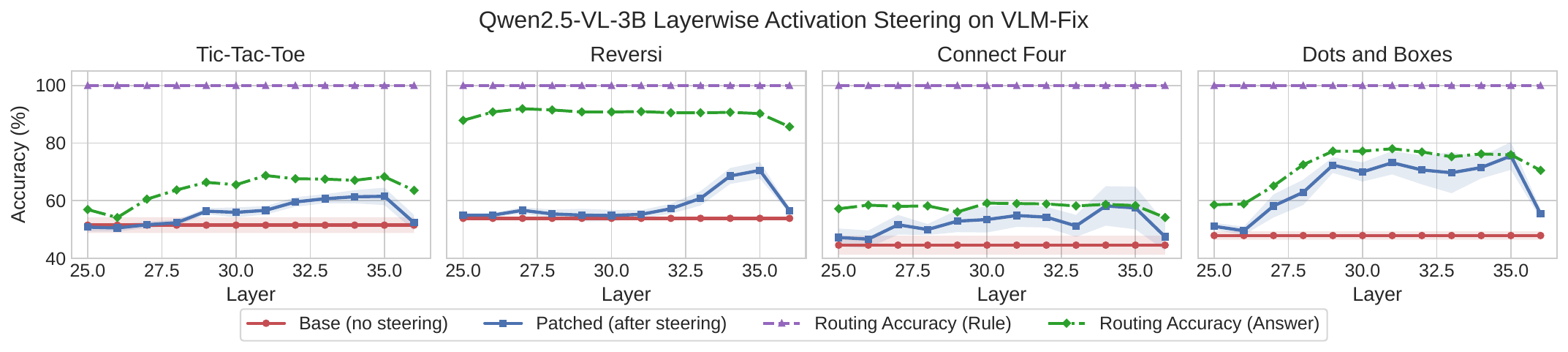}
\caption{Qwen2.5-VL-3B layerwise activation steering on VLM-Fix across Tic-Tac-Toe, Reversi, Connect Four, and Dots and Boxes.}
\label{fig:appendix_mech_vlmfix_qwen25vl3b_4games}
\end{figure*}

\begin{figure*}[!t]
\centering
\includegraphics[width=0.99\linewidth,height=0.24\textheight,keepaspectratio]{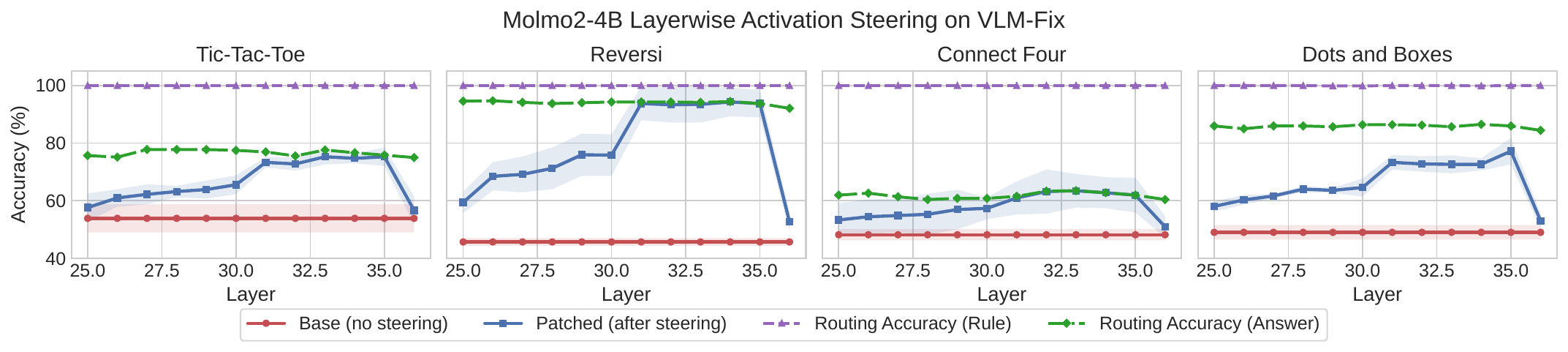}
\caption{Molmo2-4B layerwise activation steering on VLM-Fix across Tic-Tac-Toe, Reversi, Connect Four, and Dots and Boxes.}
\label{fig:appendix_mech_vlmfix_molmo2_4b_4games}
\end{figure*}

\begin{figure*}[!t]
\centering
\includegraphics[width=0.99\linewidth,height=0.24\textheight,keepaspectratio]{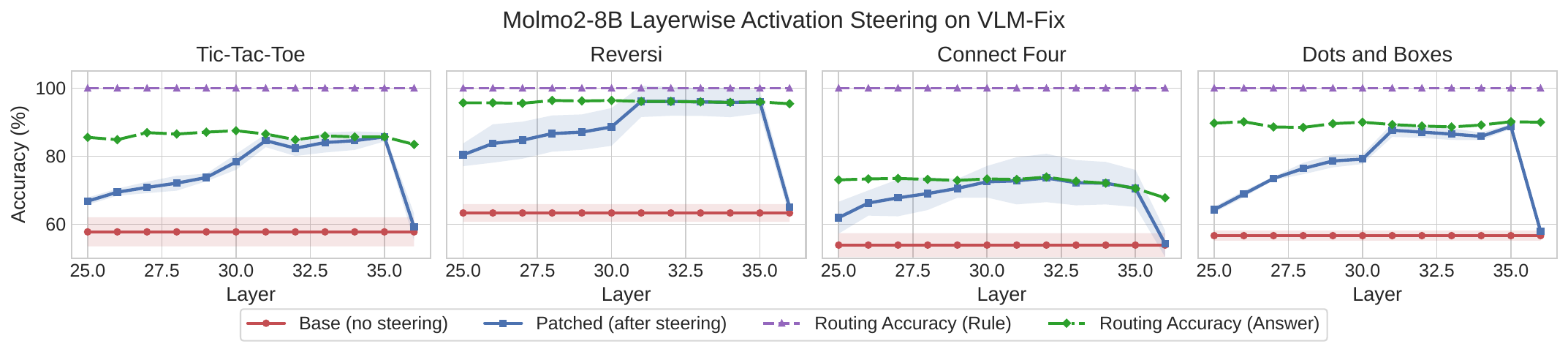}
\caption{Molmo2-8B layerwise activation steering on VLM-Fix across Tic-Tac-Toe, Reversi, Connect Four, and Dots and Boxes.}
\label{fig:appendix_mech_vlmfix_molmo2_8b_4games}
\end{figure*}

\begin{figure*}[!t]
\centering
\includegraphics[width=0.99\linewidth,height=0.24\textheight,keepaspectratio]{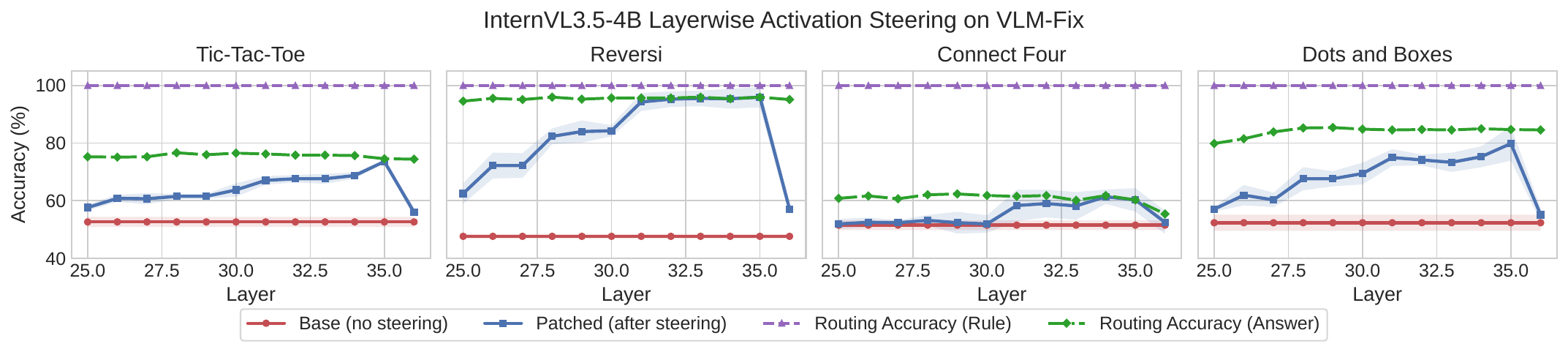}
\caption{InternVL3.5-4B layerwise activation steering on VLM-Fix across Tic-Tac-Toe, Reversi, Connect Four, and Dots and Boxes.}
\label{fig:appendix_mech_vlmfix_internvl35_4b_4games}
\end{figure*}

Additional VLM-Fix steering results for Qwen2.5-VL-3B, Molmo2-4B, Molmo2-8B, and InternVL3.5-4B are shown in Figures~\ref{fig:appendix_mech_vlmfix_qwen25vl3b_4games}, \ref{fig:appendix_mech_vlmfix_molmo2_4b_4games}, \ref{fig:appendix_mech_vlmfix_molmo2_8b_4games}, and \ref{fig:appendix_mech_vlmfix_internvl35_4b_4games}.

\subsection{VLMBias Animals}
\label{app:steering_details_vlmbias}
\paragraph{Additional details not shown in the main text.}
The VLMBias steering analysis is run in a controlled 2-leg vs 3-leg regime. We deduplicate by underlying image identity before splitting, then use stratified 70/30 train/test splits across ten random seeds so class balance is preserved.

For each of the final 12 decoder layers, we train a linear router on base-model activations, compute class centroids from SFT-donor activations (with fallback when donor-correct examples are sparse), and patch once at the query token with $\alpha=1.0$. Layerwise results report base, donor, patched, and routing accuracy as mean and standard deviation over seeds.

\end{document}